%% file: neurips_2026.tex
\documentclass{article}

\usepackage[preprint]{neurips_2026}

\usepackage[utf8]{inputenc}
\usepackage[T1]{fontenc}
\usepackage{microtype}
\usepackage{graphicx}
\usepackage{subcaption}
\usepackage{booktabs}
\usepackage{hyperref}
\usepackage{url}
\usepackage{xcolor}
\usepackage{pifont}
\usepackage{enumitem}
\usepackage{amsmath}
\usepackage{amssymb}
\usepackage{amsfonts}
\usepackage{makecell}
\usepackage{wrapfig}
  \usepackage{fontawesome5}
  \definecolor{gmailred}{HTML}{EA4335}
  \definecolor{googleblue}{HTML}{4285F4}
  \definecolor{linkedinblue}{HTML}{0A66C2}
  \definecolor{whatsappgreen}{HTML}{25D366}  
\usepackage{tikz}
\usetikzlibrary{positioning, arrows.meta, calc, fit, backgrounds}   

\usepackage{mathtools}
\usepackage{amsthm}
\usepackage[capitalize,noabbrev]{cleveref}
\usepackage{tabularx}
\usepackage{array}
\usepackage[most]{tcolorbox}
\usepackage{listings}

\newtheorem{finding}{Finding}

\lstdefinestyle{jsonprompt}{
  basicstyle=\ttfamily\footnotesize,
  breaklines=true,
  columns=fullflexible,
  keepspaces=true,
  showstringspaces=false,
  frame=none
}
\newcommand{\modeheader}[1]{%
  {\color{gray}\footnotesize\bfseries #1}\par\vspace{0.2em}%
}

\newcommand{\muse}{\texttt{DynamicMem}}

\tcbset{
  promptbox/.style={
    colback=gray!10,
    colframe=gray!70,
    fonttitle=\bfseries,
    title=Evaluation Prompt,
    boxrule=0.5mm,
    arc=4mm,
    left=4mm,
    right=4mm,
    top=2mm,
    bottom=2mm
  }
}

\theoremstyle{plain}

\theoremstyle{definition}

\theoremstyle{remark}

\theoremstyle{plain}

\title{DynamicMem: A Long-Horizon Memory Benchmark in Real-World Settings}

\author{%
  \normalfont
  Wenya Xie$^{1}$ \quad
  Shengming Zhou$^{1}$ \quad
  Zelin Li$^{1}$ \quad
  Pouya Parsa$^{1}$ \\
  Shuang Zhou$^{1}$ \quad
  Xinheng Ding$^{1}$ \quad
  Chinmay Arvind$^{1}$ \quad
  Guanchu Wang$^{2}$ \\
  Vladimir Braverman$^{3}$ \quad
  Ali Payani$^{4}$ \quad
  Yantao Zheng$^{5}$ \quad
  Zirui Liu$^{1}$\thanks{Corresponding author: \texttt{zrliu@umn.edu}} \\[0.6em]
  $^{1}$University of Minnesota \quad
  $^{2}$University of North Carolina at Charlotte \quad
  $^{3}$Johns Hopkins University \\
  $^{4}$Cisco \quad $^{5}$Adobe
}

\begin{document}

\maketitle

\begin{abstract}
\input{NeurIPS_2026/sections/abstract}
\end{abstract}

\input{NeurIPS_2026/sections/introduction}

\input{NeurIPS_2026/sections/related_work}

\input{NeurIPS_2026/sections/data_construction}

\input{NeurIPS_2026/sections/benchmark}

\input{NeurIPS_2026/sections/experiment}
\input{NeurIPS_2026/sections/error_analysis}

\input{NeurIPS_2026/sections/conclusion}

\clearpage
\bibliographystyle{plainnat}
\bibliography{NeurIPS_2026/example_paper}

\newpage
\input{NeurIPS_2026/sections/appendix}

\makeatletter
\if@preprint\else
  \clearpage
  \newpage
  \input{NeurIPS_2026/checklist}
\fi
\makeatother

\end{document}

%% file: NeurIPS_2026/sections/abstract.tex
LLM agents increasingly act as personal assistants that must remember a user’s profile
over months, including who they are (\emph{attributes}), what they routinely do (\emph{habits}), and what they prefer (\emph{preferences}), and keep it updated as the user’s job, routines, and
preferences drift.
Existing benchmarks evaluate this “memory” ability through short, simplified interactions, but they miss three core properties of real long-horizon user behavior. First, user profile is heterogeneous: attributes, habits, and preferences evolve on different timelines. Second, profile changes are not random but driven by external context, such as seasons and major life events. Third, the evidence for user profile is rarely stated explicitly or concentrated in one place. Instead, it is scattered across many small actions in different applications, requiring a memory system to infer the user’s current profile from distributed behavioral traces.
We introduce \muse, a synthetic benchmark that constructs 15 months of activity for each user, providing the kind of long-term, multi-app data that real users' privacy keeps out of reach. \muse~provides 15-month, user-consistent trajectories averaging 2.2M tokens and 1,772 grounded events per user across 16 applications such as e-commerce, fitness, and social platforms. A user's attributes, habits, and preferences evolve over this period, driven by seasons and life events, and every recorded action reflects the user's profile at that moment. 
In \muse, user profile is not given explicitly. Each attribute, habit, or preference must be inferred from small behavioral signals scattered across different apps.
We evaluate at five quarterly checkpoints to track how each system scales as the history grows. Benchmarking five representative systems exposes problems that a single accuracy score hides: (i) profile reconstruction degrades with history length while service-task accuracy stays flat, despite both drawing on the same memory; (ii) no system manages to both keep facts that stay true and replace facts that change, and the errors cluster on preferences (which users never state outright) and on naming the exact thing a fact refers to (which gym, which work tool); and (iii) over 93\% of failures trace to what the memory retrieves, not to the model that writes the final answer—so the largest room for improvement lies in memory itself. Code is available at \url{https://wenyaxie023.github.io/DynamicMem/}.

%% file: NeurIPS_2026/sections/introduction.tex
\section{Introduction}


LLM agents are increasingly deployed as personal assistants that carry out tasks for users in everyday scenarios, e.g., searching, scheduling, and purchasing. What sets a personal assistant apart from a generic one is its ability to build and maintain a model of the user's profile from behavioral history—who they are, what they routinely do, and what they prefer, and how these change over time.
Maintaining this profile is fundamentally a memory problem, which makes long-term memory a core capability for personal assistants rather than a peripheral one~\citep{hu2025evaluating}. Evaluating this capability, however, is extremely difficult: although a real user's long-term history would be the ideal testbed, privacy keeps such data out of reach, leaving the community without a reliable basis for measuring progress~\citep{bougie2025simusersimulatinguserbehavior}.

Thus, synthetic user data becomes the only available source for building such
benchmarks. However, existing benchmarks fall short on four fronts.
\ding{182} \textit{Lacking trajectory-level, cross-app evidence for user-profile
modeling.} Prior benchmarks render history as multi-session
dialogue~\citep{wu2024longmemeval,maharana2024evaluating}, or, at best, as a list of separate events that are each generated on their own, with nothing linking them. In reality, a single underlying attribute or preference drives a coordinated set of behaviors across apps, where each event is one visible trace of the same cause.
\ding{183} \emph{Treating the user as an all-in-one profile, ignoring that different aspects of a user change in different ways.} A user's profile is heterogeneous: some aspects change abruptly with discrete life events (e.g., starting a new job), while others adapt gradually through repeated behavior (e.g., settling into a weekly grocery routine). Flattening these into one profile
prevents diagnosing which aspect a memory system actually fails on.
\ding{184} \textit{Treating the user's profile as static, or letting it drift without a cause.} Real users' attributes, habits, and preferences shift over time, typically triggered by external conditions such as seasons, major events, or context shifts. Existing benchmarks either freeze profiles at creation time,
reducing memory tasks to retrieval over a fixed snapshot, or allow unmotivated drift with no causal driver linking each change to a cause. \ding{185} \textit{Single-snapshot evaluation hides how performance scales with the time horizon.} Existing benchmarks report accuracy at one fixed history length, offering no controlled view of how methods degrade as trajectories grow. Yet temporal scalability is precisely what long-term memory must deliver.

\begin{figure*}[t]
\centering
\includegraphics[width=1\textwidth]{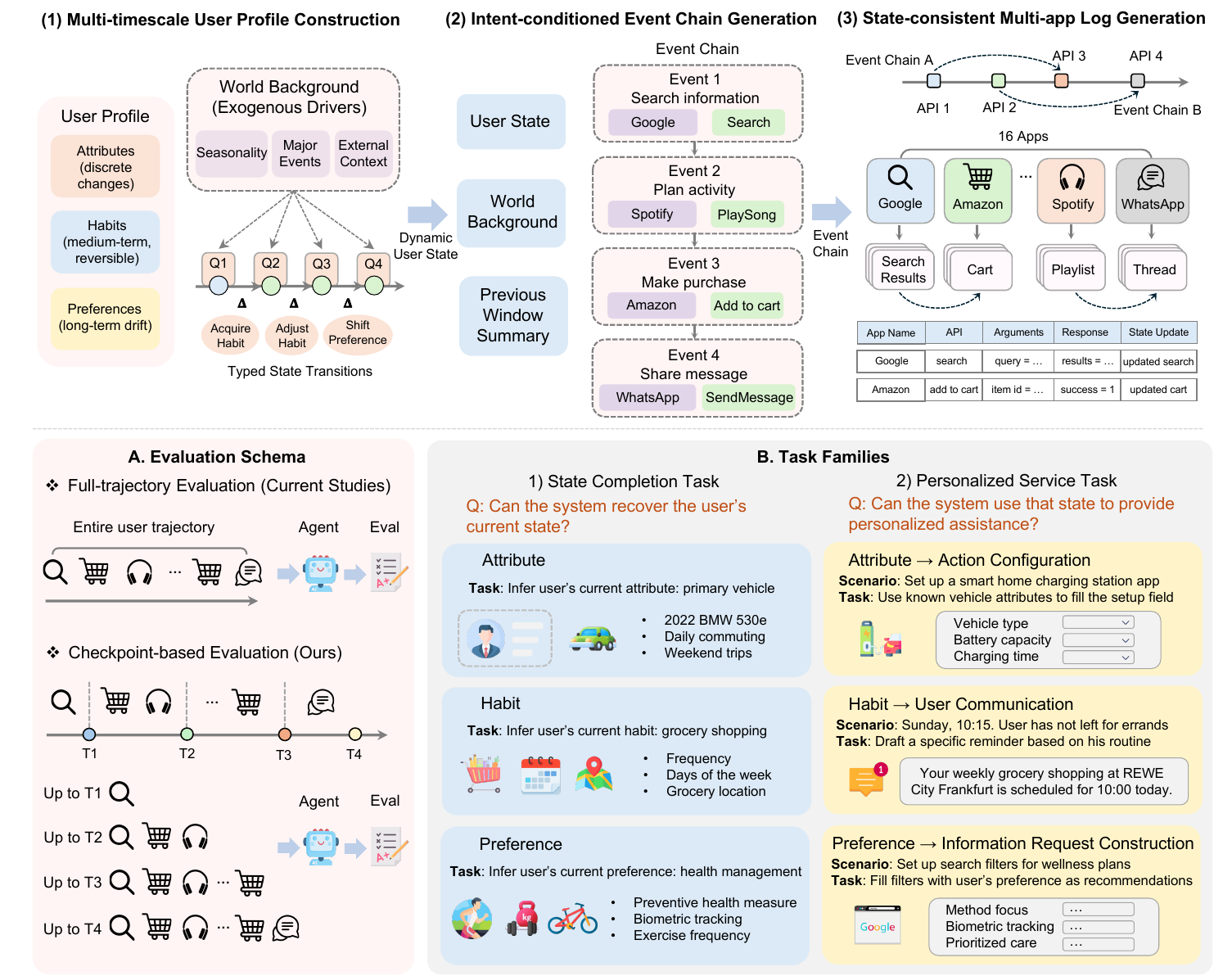}
\caption{\muse~constructs evolving user profiles, generates intent-driven event chains, and grounds them into state-consistent multi-app interaction logs for long-horizon memory evaluation.}
\label{fig1}
\end{figure*}

To fill these gaps, we introduce \muse, a synthetic benchmark whose scalable pipeline produces 15-month, fine-grained user trajectories without concatenating chat histories from other datasets, exceeding two million tokens per user. As shown in Figure~\ref{fig1}, \muse~addresses the four limitations as follows:
\textbf{First, coordinated cross-app behaviors.} As shown in Figure \ref{fig:event_chain_add}, across 16 apps we generate event chains where a single user intent unfolds over months, with each log entry conditioned on the current app state and on earlier steps in the same chain. The apps are \emph{stateful environments}, so entity references stay consistent both within each app and across apps over the whole trajectory.
\textbf{Second, multi-timescale user modeling.} Rather than representing the user as a single flat profile, we decompose it into three categories that change in distinct ways: \emph{attributes} (discrete facts that change with life events), \emph{habits} (medium-term routines that adapt to context), and \emph{preferences} (long-term inclinations that drift gradually). Each category is probed by a dedicated pair of tasks, namely \emph{State Completion} (reconstructing the category) and \emph{Personalized Service} (acting on it).
\textbf{Third, causally grounded, coupled profile evolution.} As shown in Figure~\ref{fig:trajectory}, the user's profile does not drift at random: each change traces to an external cause—a season, a major event, a shift in circumstances—and a single cause often moves several facets together, e.g., a winter fitness push changes the user's gear, routine, and tastes at once.

\textbf{Fourth, checkpoint-based evaluation.} Rather than a single full-history evaluation, we probe memory at five quarterly checkpoints, exposing how performance scales with history length and whether systems correctly overwrite superseded values as the profile evolves.
To our knowledge, \muse~is the most fine-grained long-horizon memory benchmark to date: about 2.2M tokens per user over 15 months, 17,715 grounded events spanning 16 apps and 66 APIs, 1,837 evolving attribute / habit / preference items linked by 383 causally-attributed state transitions, and 4,994 fine-grained scoring points across 3,634 checkpoint-based problems. 
Benchmarking five representative memory systems, we surface drawbacks that a single accuracy number hides. First, as the history grows, State Completion accuracy falls steadily while Personalized Service does not, even though both read the same memory. Second, no system both retains facts that stay true and overwrites facts that change; errors concentrate on implicit preferences and on naming the exact entity behind an attribute (e.g., the wrong work tool). Third, over 93\% of failures trace to what the memory system retrieves rather than to the answer model, so the headroom lies in memory itself.

%% file: NeurIPS_2026/sections/related_work.tex
\vspace{-1em}
\section{Related Work}

Due to the page limits, we briefly discuss two research lines here. The full related work section can be found in Appendix \ref{app:full_related_work}.

  \noindent \textbf{Long-Horizon Memory Benchmarks.}                     
  Prior benchmarks cover only parts of long-horizon personalization.                 
  Conversation-centric                                                               
  datasets~\citep{wu2024longmemeval,maharana2024evaluating,pakhomov2025convomem}
  test multi-session recall but stop at dialogue context, not heterogeneous          
  application traces. Recent memory-agent benchmarks~\citep{tan2025membench,hu2025evaluating, xu2026memgym} broaden evaluation to interactive or agentic memory settings, but they still do not model a user profile as an evolving state grounded in coordinated behavior across applications. Lifelog and timeline                                           
  benchmarks~\citep{tan2023timelineqa,hu2026clonemem}                                 
  strengthen temporally grounded recall, yet treat events as evidence to retrieve    
  rather than as drivers of later state change. Personalization                      
  benchmarks~\citep{jiang2025know,jiang2025personamem,wu2026knowme,huang2026mem} evaluate
  persona or preference consistency, but assume user attributes are fixed facts      
  to be recalled. \muse~instead requires retrieval, causal attribution, and
  dynamic state tracking to be solved jointly over months of interleaved             
  multi-app activity, where the same evidence both answers a query and updates
  the underlying user state.    
                                                                                     
  \noindent \textbf{Memory Systems for Agents.}                                             
  Memory architectures for long-horizon agents fall into three families:             
  retrieval-first methods~\citep{lewis2021retrievalaugmentedgenerationknowledgeintensivenlp,gutiérrez2025ragmemorynonparametriccontinual,rasmussen2025zep}               
  improve evidence access via vector or graph retrieval but rely on heuristic        
  updates; structured external-memory                                                
  systems~\citep{packer2023memgpt,chhikara2025mem0,xu2025mem,wang2025mem,kang2025memoryosaiagent,nan2025nemori}                                                          
  expose explicit operations such as summarization, consolidation, and
  overwrite, improving controllability but leaving evolving-state tracking           
  under sparse evidence open; and learned-policy memory                              
  methods~\citep{wang2023augmenting,wang2024memoryllm,wang2025m+,yan2025memory,yu2025memagent,yu2026agentic}                                                             
  decide what and when to write or update. Recent personalization-oriented memory systems~\citep{tan2025prospect,zhang2025prime} further study reflective or cognitive memory organization for long-term interaction. These systems motivate the need for evaluation beyond aggregate recall: \muse~ diagnoses whether memory designs can retain stable facts, overwrite changed facts, infer implicit preferences, and integrate evidence scattered across applications.

%% file: NeurIPS_2026/sections/data_construction.tex
\section{\muse: Trajectory Synthesis Pipeline}
\label{sec:data_construction}
Here we introduce our scalable trajectory synthesis pipeline that generates long-horizon, fine-grained user trajectories.
A key strength of \muse~is its scalability: for the first time, we can generate more than one year long trajectories with over two million tokens per user without concatenating irrelevant chat histories from other datasets.
As shown in Figure~\ref{fig1}, the pipeline is organized top-down in three 
stages: 
\textit{(1) Multi-timescale user profile construction} produces an evolving 
user profile; 
\textit{(2) Intent-conditioned event chain generation} which turns that profile into 
coherent multi-step intents; and 
\textit{(3) State-consistent multi-app log generation} which grounds each intent 
into concrete app interactions. 
Each stage is conditioned on the output of the previous one, so every app log 
is traceable to the user intent and profile state that produced it. We detail each stage below.

\subsection{User Modeling with Multi-Timescale Dynamics}
\label{sec:user_modeling}

\input{NeurIPS_2026/figures/traj}

\noindent \textbf{Profile schema.}
Each user is represented as a static base profile $b$ plus a dynamic profile that evolves over time.
As listed in Appendix~\ref{app:base_profile}, the static base profile $b$ contains stable characteristics, e.g., demographics, socioeconomic factors, and personality traits, and is fixed throughout the synthesis horizon.
Following SWB literature and OECD Better Life Index \citep{oecdbetterlife}, the dynamic profile spans $K{=}6$ life domains, i.e., \textit{Work \& Education}; \textit{Family \& Close Relationships}; \textit{Social \& Community}; \textit{Health \& Self-care}; \textit{Finances \& Material Living}; \textit{Leisure \& Media} (details are listed in Appendix~\ref{app:life_domains}).

As shown in Figure \ref{fig:trajectory}, the user profile evolves over five quarterly checkpoints across a 15-month horizon.
Within each domain and quarter-long time window, the user profile factorizes into three categories with distinct dynamics:
\begin{itemize}[leftmargin=*, itemsep=1pt, topsep=0pt]
    \item \textbf{Attributes} ($A$): \emph{facts and possessions}---what the user \emph{has} or \emph{is}. They change discretely when life events occur (e.g., getting a new job, acquiring a gym membership, moving to a new city).

\item \textbf{Habits} ($H$): \emph{recurring behavioral routines}---what the user \emph{regularly does}. They adapt to medium-term context and may revert once that context fades (e.g., morning runs picked up for marathon training, dropped after the race).

\item \textbf{Preferences} ($P$): \emph{comparative inclinations}---what the user \emph{tends to choose} and \emph{wants}. They drift gradually over the long term (e.g., a slow shift from fast food toward healthier options).
\end{itemize}
Rather than free text, each item is stored as a structured record suited to its category, enabling fine-grained tracking of what changes between checkpoints. Details can be found in Appendix~\ref{app:representation_of_dynamic_profile}.

\noindent \textbf{Evolution mechanism.}
Between consecutive windows, state evolves through a three-step pipeline: \textit{world background} $\rightarrow$ \textit{typed deltas} $\rightarrow$ \textit{state update}.

We first derive a personalized \textbf{world background} for the upcoming window from $b$, encoding seasonality, major events, and external context.

The same external condition (e.g., a heatwave) affects users differently depending on their location and lifestyle (Appendix~\ref{app:world background}).
Conditioned on the world background and the prior state, we sample item-level \textbf{typed deltas} restricted to category-specific transition types: \textsc{Add}/\textsc{Remove}/\textsc{Modify} for attributes, \textsc{Acquire}/\textsc{Adjust}/\textsc{Drop} for habits, and \textsc{Shift}/\textsc{Refine} for preferences.
Each delta touches at most a few items and must carry an explicit causal reason linking back to the world background, ruling out implausible flips and enabling evaluation stratified by transition type.
A single world-background driver often induces \emph{coupled} deltas across
categories rather than an isolated edit. In one user's winter window, for example, a New-Year effort to relieve back pain simultaneously adds home gym equipment (an attribute), acquires a thrice-weekly basement strength routine (a habit), and refines the user's fitness philosophy toward structured, data-driven workouts (a preference). Because these changes share one cause and each is annotated with it, the trajectory carries correlated, cross-category evidence rather than independent edits.
Applying the delta yields the new state.
Crucially, each window conditions on a summary of the previous window's full trajectory, so drift accumulates coherently across the year rather than resetting at each window boundary.

\textbf{Human Validation.}
In \muse, we initially synthesize ten diverse personas from PersonaHub~\citep{ge2025scalingsyntheticdatacreation}, expand each into a full base profile, and generate domain trajectories over five quarterly windows with cross-domain consistency constraints. Manual validation is prohibitively expensive: verifying consistency across $n$ events requires $O(n^2)$ pairwise checks, which took over 5 hours per user in our pilot study. We instead summarize common failure patterns (e.g., unreverted temporary habits, invalid transitions, intra-/inter-domain conflicts) into seven rubrics, embedded inline as an LLM reviser that detects and repairs violations during synthesis and escalates only low-confidence cases to humans. Details can be found in Appendix~\ref{app:validation-multitimescale-details}.

\input{NeurIPS_2026/figures/event_chain}

\subsection{\mbox{Intent-Conditioned Event Chain Generation}}
Given the user profiles from Section \ref{sec:user_modeling}, we then generate behavioral sequences that turn both inter-window changes and stable within-window context into observable app-level events.
As shown in Figure \ref{fig:event_chain_add}, we define an \emph{event chain} as an ordered sequence of intent-driven events that reflects one state change (or a small group of related changes) within a single domain and time window.
Each event specifies: (1) a fine-grained \emph{intent} capturing the user's motivation, and (2) an \emph{app--API pair} selected from a predefined suite of 16 apps (detailed in Appendix~\ref{app:app_schema}) that will execute the intent.

We generate chains rather than isolated events because one single event may be noisy, whereas a coherent chain provides multi-event evidence that can be reliably attributed to the underlying change.\\
\noindent\textbf{Generation procedure.} 
We generate event chains per domain and per window, conditioning each window on a summary of the previous one so intent carries over across boundaries. Within a window, we enforce the \textbf{health coverage}: every non-empty state delta must be realized by at least one chain. In this way, every change has corresponding evidence. Each chain is then generated by LLMs with the current state, the world background, and the specific change or stable item it should express.
For example, 
Figure \ref{fig:event_chain_add} illustrates how an event chain embeds user intent: three events across Gmail, LinkedIn, and WhatsApp jointly realize a single goal, i.e., obtaining and acting on an EPA VOC certification, triggered by the Q1 regulatory update in Figure \ref{fig:trajectory}. Each event plays a distinct role, namely, trigger, announce, consequence. 

\textbf{Human Validation.}
We manually validate 190 event chains across all users and domains.
Each chain is assessed for (1) \emph{intent coherence}: whether events logically follow from the stated intent, and (2) \emph{state grounding}: whether the chain plausibly reflects the associated state change.
The generated chains achieve 97\% intent coherence and 98\% state grounding accuracy.

\subsection{State-Consistent Multi-app Log Generation}
Given logically linked event chains, we ground each event into concrete multi-app interaction logs so that the trajectory reflects user behaviors.
Each log entry follows a strict schema that includes the app name, API name, request, and the response, as detailed in Appendix~\ref{app:app_schema}.

\noindent \textbf{From event chains to executable API calls.}
For each chain step, we translate the free-form intent into structured API arguments, e.g., converting an information-seeking intent into a concrete search query, filters, and a result size. 
Argument generation is constrained by (i) API signatures and value types; (ii) the current app state, e.g., only previously surfaced item identifiers can be referenced; (iii) the preceding steps within the same chain.
Together these constraints rule out hallucinated identifiers and ensure that every action is grounded in previously observable evidence.\\
\noindent \textbf{Stateful app environments.} Each of the 16 apps is a \emph{stateful environment} rather than a one-off text template: it carries an explicit state that every API call reads and updates, so the trajectory stays internally consistent the way a real app would. This consistency holds at two levels. \emph{Within} an app, we maintain a persistent store of the objects prior responses introduced—search results, carts, wishlists, playlists, message threads—and each step (1) produces schema-compliant arguments and a response, (2) deterministically updates the store (e.g., adds an item to the cart, then empties it at checkout), and (3) conditions later generation on the updated state. \emph{Across} apps, each step is also conditioned on earlier steps in the same chain, so an entity introduced in one app is carried forward and referenced in another—a certification obtained by email is the one later announced on a
professional network. Rather than retrieving from a handcrafted database, we synthesize responses with an LLM conditioned on the user's base profile, the step intent, the event-chain context, and the current app state, yielding both entity consistency (stable identifiers across references) and behavioral consistency
(transitions that match application semantics).

\noindent \textbf{Human Validation.}
We manually verified consistency along two axes: (i) we randomly sampled 5 consecutive interaction steps per app.
We checked argument validity, state transitions, and entity reference consistency; and (ii) 20 complete event chains, checking that later steps are grounded in earlier outputs and app states without hallucinated entities. 
For each chain, we verified that later steps were properly grounded in earlier outputs and application states, and that no hallucinated entities or invalid references appeared. 

%% file: NeurIPS_2026/figures/traj.tex
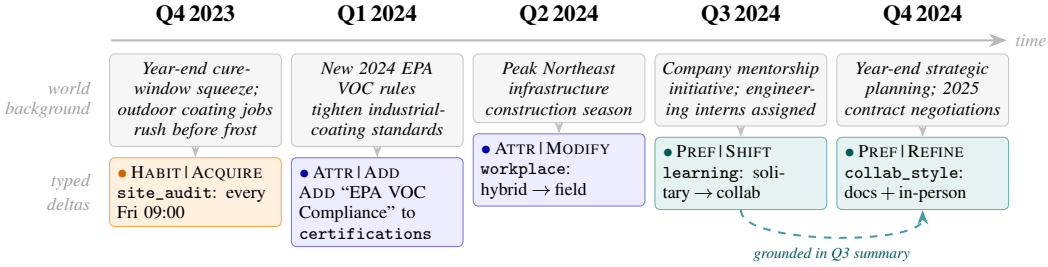
\begin{figure}[h]
  \centering
  \small
  \resizebox{\linewidth}{!}{%
  \begin{tikzpicture}[
    font=\footnotesize,
    qhead/.style={font=\bfseries\small, anchor=center},
    bg/.style={
      draw=gray!45, rounded corners=2pt, fill=gray!7,
      text width=2.15cm, align=center, inner sep=3pt,
      font=\scriptsize\itshape
    },
    delta/.style={
      rounded corners=2pt, draw, text width=2.15cm,
      align=left, inner sep=3pt, font=\scriptsize
    },
    attr/.style ={delta, fill=blue!7,    draw=blue!55},
    habit/.style={delta, fill=orange!12, draw=orange!70},
    pref/.style ={delta, fill=teal!10,   draw=teal!60},
    tag/.style={font=\scriptsize\bfseries},
    axis/.style={-{Stealth[length=2.5mm]}, thick, gray!55},
    ground/.style={-{Stealth[length=2mm]}, thick, teal!75, dashed},
    colsep/.initial=2.5,
  ]

  \foreach \i/\q in {1/Q4\,2023, 2/Q1\,2024, 3/Q2\,2024, 4/Q3\,2024, 5/Q4\,2024}{
    \node[qhead] (h\i) at ({(\i-1)*2.5}, 0) {\q};
  }

  \draw[axis] ($(h1.west)+(-0.5,-0.35)$) -- ($(h5.east)+(0.5,-0.35)$)
    node[right, font=\scriptsize\itshape, gray!70] {time};

  \node[bg, below=8pt of h1.south] (b1)
    {Year-end cure-window squeeze; outdoor coating jobs rush before frost};
  \node[bg, below=8pt of h2.south] (b2)
    {New 2024 EPA VOC rules tighten industrial-coating standards};
  \node[bg, below=8pt of h3.south] (b3)
    {Peak Northeast infrastructure construction season};
  \node[bg, below=8pt of h4.south] (b4)
    {Company mentorship initiative; engineering interns assigned};
  \node[bg, below=8pt of h5.south] (b5)
    {Year-end strategic planning; 2025 contract negotiations};

  \node[font=\scriptsize\itshape, gray!75, anchor=east]
    at ($(b1.west)+(-0.15,0)$) {\shortstack[r]{world\\background}};

  \node[habit, below=4pt of b1] (q1a)
    {\textcolor{orange!80!black}{$\bullet$}\,\textsc{Habit}\,\textbar\,\textsc{Acquire}\\
     \texttt{site\_audit}: every Fri 09:00};

  \node[attr, below=4pt of b2] (q2a)
    {\textcolor{blue!65!black}{$\bullet$}\,\textsc{Attr}\,\textbar\,\textsc{Add}\\
     \textsc{Add} ``EPA VOC Compliance'' to \texttt{certifications}};

  \node[attr, below=4pt of b3] (q3a)
    {\textcolor{blue!65!black}{$\bullet$}\,\textsc{Attr}\,\textbar\,\textsc{Modify}\\
     \texttt{workplace}: hybrid$\,\to\,$field};

  \node[pref, below=4pt of b4] (q4a)
    {\textcolor{teal!70!black}{$\bullet$}\,\textsc{Pref}\,\textbar\,\textsc{Shift}\\
     \texttt{learning}: solitary\,$\to$\,collab};

  \node[pref, below=4pt of b5] (q5a)
    {\textcolor{teal!70!black}{$\bullet$}\,\textsc{Pref}\,\textbar\,\textsc{Refine}\\
     \texttt{collab\_style}: docs\,$+$\,in-person};

  \node[font=\scriptsize\itshape, gray!75, anchor=east]
    at ($(q1a.west)+(-0.15,0)$) {\shortstack[r]{typed\\deltas}};

  \foreach \src/\dst in {b1/q1a, b2/q2a, b3/q3a, b4/q4a, b5/q5a}{
    \draw[-{Stealth[length=1.6mm]}, gray!50, thin] (\src.south) -- (\dst.north);
  }

  \draw[ground]
    (q4a.south) to[out=-90, in=-90, looseness=0.5]
    node[pos=0.5, below=1pt, font=\tiny\itshape, teal!70!black]
      {grounded in Q3 summary}
    (q5a.south);

  \end{tikzpicture}%
  }

  \vspace{-.5em}
  \caption{A snapshot of five quarterly checkpoints capturing how a Pittsburgh-based coating
  consultant's profile evolved across one year from \muse, anchored at the end of 2023.
  Each checkpoint pairs a personalized world background with the item-level typed deltas it
  triggers.
  (\textcolor{blue!65!black}{$\bullet$}\,attribute,
  \textcolor{orange!80!black}{$\bullet$}\,habit,
  \textcolor{teal!70!black}{$\bullet$}\,preference).
  The dashed arrow illustrates cross-window grounding: the Q4 \textsc{Refine} on
  \texttt{collab\_style} is conditioned on the Q3 mentorship summary rather than emerging in
  isolation.}
  \vspace{-1em}
  \label{fig:trajectory}
\end{figure}

%% file: NeurIPS_2026/figures/event_chain.tex
    \begin{wrapfigure}{r}{0.50\textwidth}                           
    \centering                                                                                    
    \small                                                                                        
                                                                  
    \resizebox{\linewidth}{!}{%
    \begin{tikzpicture}[
      font=\scriptsize,
      ev/.style={                                                 
        draw, rounded corners=2pt, inner sep=2pt,                                                 
        minimum width=1.15cm, minimum height=8mm,                                                 
        align=center, fill=blue!4, draw=blue!45                                                   
      },                                                                                          
      arr/.style={-{Stealth[length=1.4mm]}, thin, gray!60},                                       
      appname/.style={font=\scriptsize\bfseries, anchor=north, inner sep=1pt},                    
      role/.style={font=\tiny\itshape, gray!75, anchor=north, inner sep=1pt},                     
      date/.style={font=\tiny, gray!55, anchor=north, inner sep=0pt},                             
    ]                                                                                             
                                                                                                  
    \def\colW{2.3}                                                                                
                                                                  
    \node[ev] at (0*\colW, 0) (a1) {\textcolor{gmailred}{\Large\faEnvelope}};
    \node[ev] at (1*\colW, 0) (a2) {\textcolor{linkedinblue}{\Large\faLinkedin}};                 
    \node[ev] at (2*\colW, 0) (a3) {\textcolor{whatsappgreen}{\Large\faWhatsapp}};                
                                                                                                  
    \draw[arr] (a1.east) -- (a2.west);                            
    \draw[arr] (a2.east) -- (a3.west);                                                            
                                                                  
    \node[appname] at (a1.south) (a1n) {Gmail};
    \node[appname] at (a2.south) (a2n) {LinkedIn};                                                
    \node[appname] at (a3.south) (a3n) {WhatsApp};
                                                                                                  
    \node[role] at (a1n.south) (a1r) {trigger};                                                   
    \node[role] at (a2n.south) (a2r) {announce};                  
    \node[role] at (a3n.south) (a3r) {consequence};                                               
  
    \node[date] at (a1r.south) {Jan 3};                           
    \node[date] at (a2r.south) {Feb 15};                                                          
    \node[date] at (a3r.south) {Mar 4};
                                                                                                  
    \end{tikzpicture}%
    }

    
    \begin{tcolorbox}[                                            
      colback=blue!3, colframe=blue!55,
      boxrule=0.5pt, arc=2pt,
      left=6pt, right=6pt, top=0pt, bottom=4pt,                                                   
      title={\textbf{Zoom-in:} Event Chain ``\texttt{professional\_certifications}'' (Q1\,2024)},
      fonttitle=\scriptsize,                                                                      
      coltitle=white, colbacktitle=blue!55,                       
      fontupper=\scriptsize,                                                                      
    ]                                                             
    \setlength{\parskip}{2pt}\setlength{\parindent}{0pt}                                          
                                                                                                  
    \textcolor{gmailred}{\faEnvelope}\ \textbf{Jan~3 — Gmail.ReadEmail.}
    \emph{Reading a regulatory-update bulletin on \textbf{2024 VOC emission standards}; motivated 
  to \textbf{stay ahead of compliance} to maintain senior-consultant status.}                     
  
    \textcolor{linkedinblue}{\faLinkedin}\ \textbf{Feb~15 — LinkedIn.AddSkill.}                   
    \emph{Updating credentials with the \textbf{newly acquired EPA Compliance Certification}.}
                                                                                                  
    \textcolor{whatsappgreen}{\faWhatsapp}\ \textbf{Mar~4 — WhatsApp.SendMessage.}                
    \emph{Messaging the primary client to confirm the cert is \textbf{secured}, enabling          
  \textbf{finalization of 2024 project specs} without regulatory delay.}                          
    \end{tcolorbox}                                                                                                                       
                                                                  
    \caption{
    Three app-level events across two months collectively reflect the user's intent,
  i.e., acting based on the EPA VOC certification.}                                           
    \label{fig:event_chain_add}                                            
  \end{wrapfigure}
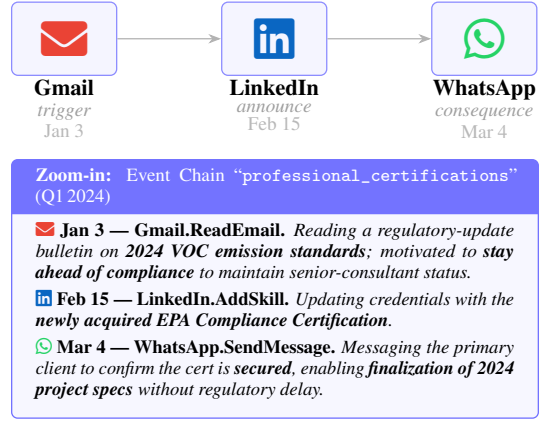 

%% file: NeurIPS_2026/sections/benchmark.tex
\section{\muse~Benchmark}
\muse~turns the trajectories constructed in Section~\ref{sec:data_construction} into evaluation tasks for long-term personalization. For each user trajectory, we define multiple temporal checkpoints. At each checkpoint, a memory system can access only the app logs before that time and is evaluated on the user's \emph{state} at that moment, i.e., the snapshot of their evolving \emph{profile} that is current at that checkpoint. The benchmark then asks two questions: can the system recover the user's current state, and can it use that state to provide personalized assistance?

\subsection{Checkpoint-Based Evaluation and Dataset Statistics}
\label{sec:muse_traj_setup}

\noindent \textbf{Checkpoint based evaluations.} 
Rather than evaluating a system after it has observed the entire trajectory, \muse~evaluates it at multiple points in time. We use the same five quarterly checkpoints introduced in Section~\ref{sec:user_modeling}. At each checkpoint, the system only has access to app logs up to the end of that quarter, and is then asked to perform two tasks, which are described later in Section~\ref{sec:muse_task_design}.

This setting supports two detailed analyses that a single full-history evaluation cannot. First, it exposes how memory performance scales with history length: as checkpoints advancing, accessible evidence grows from 3 months to 15 months, which reveals whether a system benefits from more evidence, saturates early, or degrades when old and new evidence must be reconciled or conflict with each other. Second, it tests whether the system tracks change rather than compressing history into a static profile: since attributes, habits, and preferences evolve over time, the gold answer to queries like ``the user's current workout routine'' generally shifts between checkpoints, so a system is rewarded only when it overwrites superseded values.

\noindent \textbf{Dataset Statistics.} Table \ref{tab:dataset_at_a_glance} summarizes the scale and composition of \muse. We initially synthesize ten diverse personas from PersonaHub~\citep{ge2025scalingsyntheticdatacreation}. In aggregate, these trajectories contain 1,790 event chains and 17,715 app-log events.
An average of over 2M tokens per user, substantially exceeding the per-user context lengths considered in prior long-horizon memory benchmarks. 
The two tasks, i.e., state completion and personalized service, will be introduced in the following subsections.

\input{NeurIPS_2026/tables/data_statistics}

\subsection{Task Families}
\label{sec:muse_task_design}
\muse~contains two checkpoint task families that probe memory at two complementary levels. \textbf{State Completion} tests whether a memory-augmented agent can \emph{recover} the user's profile at a given checkpoint, e.g., describing the user's current workout routine or dietary preference. \textbf{Personalized Service} tests whether the agent can \emph{act on} that state in downstream services, e.g., reminding the user when it is time to perform a habitual activity or pre-filtering options according to their preferences. 
Specifically, as shown in Figure~\ref{fig1}, \textbf{each of the three profile
categories from Section~\ref{sec:user_modeling} (\emph{attributes},
\emph{habits}, and \emph{preferences}) is probed by \emph{both} task
families}: a \emph{state completion} task that tests whether the system
can recover that category of state, and a \emph{personalized service}
task that tests whether it can act on that same state in a given
scenario. 

\noindent \textbf{State Completion}: As shown in Figure \ref{fig:state-completion-cloze}, this task requires the model to fill blank fields in a structured JSON schema at a given checkpoint. The question may ask about a user’s current workout routine, financial constraint, or media preference, and the answer should summarize the details supported by the app logs. This task evaluates whether a memory system can form a complete and checkpoint-specific view of a user profile. These categories require different memory behavior. Attributes are often grounded in explicit facts or discrete events; habits require abstraction over repeated behavior; and preferences are implicit, comparative, and distributed across many choices.

\input{NeurIPS_2026/figures/state_completion}

\noindent  \textbf{Personalized Service:} This task evaluates whether a memory system can turn what it remembers about the user into a service that a generic assistant could not provide. 
Each task pairs a generic, leakage-safe ``scenario'' description (time/place/situation only) with a fixed task instruction.
The scenario deliberately omits the user-state details needed to give a correct 
  answer. 

As shown in Figure \ref{fig1}, personalized service is decomposed into three sub-categories corresponding to the habit, attribute, and preference in the user modeling. 
\textbf{Attributes} tasks require filling entity-specific configurations, e.g., vehicle details for a parking permit. 
\textbf{Habits} tasks require drafting context-specific reminders, e.g., given ``Sunday 9:15, coffee just poured,'' invoking the user's 09:30 budget review at the home office. Success requires jointly recovering routine, frequency, time, day, and location.
\textbf{Preferences} tasks require configuring scenario-appropriate filters, e.g., the role-type filter for a civic-engagement search.  We provide details about the data synthesis pipeline in Appendix~\ref{app:data_synthesis_task_construction}.

  \subsection{Evaluation Protocol and Metrics}                                                                                                                                                                                       
  \label{sec:muse_eval_metrics}                                                                                                                                                                                                      
Both tasks are evaluated using LLM-as-judge method. For each field
to be filled, e.g, each \textcolor{red!70!black}{\textbf{<fill>}} in 
Figure~\ref{fig:state-completion-cloze}, the judge returns two scores: 
$\textsc{Core}\in\{0,1\}$, indicating whether the prediction recovers 
the central fact, and $\textsc{Detail}\in\{0,1,2\}$, which measures how well it preserves the supporting specifics, with $0$, $1$, and $2$ corresponding to incorrect, partially correct, and fully correct details, respectively. The final per-field score 
is $s = 0.8\cdot\textsc{Core} + 0.2\cdot(\textsc{Detail}/2)$.

  \noindent \textbf{State Completion.}
We score every field (each \textcolor{red!70!black}{\textbf{<fill>}} in Figure~\ref{fig:state-completion-cloze}) independently and average into the final score. Take the weekend-walk habit in Figure~\ref{fig:state-completion-cloze}: the reference specifies the walk's frequency, days, start/end time, and location. Suppose the prediction is \emph{``Saturday weekly walk at 06:35 by the river''} against the ground truth of \emph{``Saturday/Sunday weekly walk at 06:30--07:30, lakefront trail''}. Frequency receives $\textsc{Core}{=}1$ (\emph{weekly} matches); start time receives $\textsc{Core}{=}1$, $\textsc{Detail}{=}1$ (06:35 is within five minutes of 06:30 but not exact); days, end time, and location receive $\textsc{Core}{=}0$. All field-level scores are averaged into the final score.

\noindent \textbf{Personalized Service.}
Personalized Service is scored under the same field-level $\textsc{Core}$+$\textsc{Detail}$ rubric as State Completion, applied to fields of the assistant response rather than fields of a state cloze. The full judging prompt is provided in Appendix ~\ref{app:prompt_template}.

%% file: NeurIPS_2026/tables/data_statistics.tex
\begin{wraptable}{r}{0.40\textwidth}
\centering
\footnotesize
\setlength{\tabcolsep}{3pt}
\renewcommand{\arraystretch}{1.05}
\vspace{-1.5em}
\caption{\small{Statistics of \muse}}
\label{tab:dataset_at_a_glance}
\vspace{-0.6em}
\resizebox{\linewidth}{!}{%
\begin{tabular}{@{}lr@{}}
\toprule
\multicolumn{2}{@{}l}{\textit{Trajectory scale}} \\
\addlinespace[1pt]
Users (from PersonaHub) & 10 \\
Time horizon & 15 months \\
Event chains & 1{,}790 \\
App-log events & 17{,}715 \\
Tokens per user & $\sim$2.2\,M \\
\midrule
\multicolumn{2}{@{}l}{\textit{Behavioral coverage}} \\
\addlinespace[1pt]
Life domains & 6 \\
Applications & 16 \\
Distinct APIs & 66 \\
\midrule
\multicolumn{2}{@{}l}{\textit{Evolving user profile}} \\
\addlinespace[1pt]
Attribute instances & 883 \\
Habit instances & 466 \\
Preference instances & 488 \\
\makecell[l]{Fields (e.g., ``\texttt{start\_time}'' \\ \quad in Fig.~\ref{fig:state-completion-cloze})} & 3{,}702 \\
\makecell[l]{State transitions \\ \quad (i.e., typed deltas in Fig.~\ref{fig:trajectory})} & 383 \\
\midrule
\multicolumn{2}{@{}l}{\textit{Evaluation tasks}} \\
\addlinespace[1pt]
State-completion problems & 1{,}824 \\
Personalized-service problems & 1{,}810 \\
\bottomrule
\end{tabular}%
}
\end{wraptable}

%% file: NeurIPS_2026/figures/state_completion.tex
  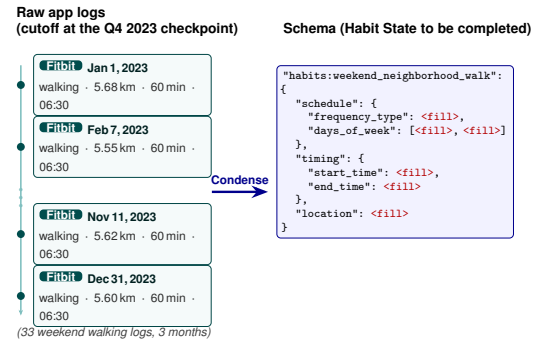
\begin{wrapfigure}{r}{0.5\textwidth}                            
  \setlength{\abovecaptionskip}{3pt}                                                                                                                                                                                                                                                     
  \setlength{\belowcaptionskip}{0pt}                                                                                                                                                                                                                                                     
  \setlength{\intextsep}{4pt}
  \setlength{\columnsep}{8pt}                                                                                                                                                                                                                                                            
  \centering     
  \vspace{-1.5em}
  \resizebox{\linewidth}{!}{%
  \begin{tikzpicture}[
    font=\footnotesize,                                                                                                                                                                                                                                                                  
    card/.style={                                                 
      rounded corners=2pt, draw=teal!55!black, line width=0.4pt,
      fill=teal!4, inner sep=3pt, text width=2.95cm, align=left,                                                                                                                                                                                                                         
    },
    pill/.style={                                                                                                                                                                                                                                                                        
      rounded corners=2pt, fill=teal!60!black, draw=none,                                                                                                                                                                                                                                
      text=white, inner xsep=2.5pt, inner ysep=0.3pt,
      font=\sffamily\bfseries\tiny,                                                                                                                                                                                                                                                      
    },                                                                                                                                                                                                                                                                                   
    tdot/.style={circle, fill=teal!60!black, inner sep=0pt, minimum size=4pt},
    schemabox/.style={                                                                                                                                                                                                                                                                   
      rounded corners=2pt, draw=blue!55!black, fill=blue!5, inner sep=3pt,                                                                                                                                                                                                               
      align=left, font=\ttfamily\tiny, text width=4.0cm,
    },                                                                                                                                                                                                                                                                                   
    header/.style={font=\sffamily\scriptsize\bfseries, anchor=west},
  ]                                                                                                                                                                                                                                                                                      
                                                                  
\node[header, anchor=south west, align=left] at (0, 5.95)
  {Raw app logs\\(cutoff at the Q4 2023 checkpoint)};                                                                                                                                                                                                                                                                                         
  \draw[-{Stealth[length=3pt]}, thick, teal!50] (0.20, 5.55) -- (0.20, 1.10);
  \node[tdot] at (0.20, 5.20) {};                                                                                                                                                                                                                                                        
  \node[tdot] at (0.20, 4.10) {};                                 
  \node[tdot] at (0.20, 2.55) {};                                                                                                                                                                                                                                                        
  \node[tdot] at (0.20, 1.45) {};                                 
  \node[font=\bfseries, text=teal!50] at (0.20, 3.30) {$\vdots$};                                                                                                                                                                                                                        
                                                                                                                                                                                                                                                                                         
  \node[card, anchor=west] at (0.42, 5.20) {%
    \tikz[baseline=-0.9ex]\node[pill]{\,Fitbit\,};\ %
    {\sffamily\bfseries\tiny\color{teal!25!black} Jan\,1,\,2023}\\                                                                                                                                                                                                                       
    {\sffamily\tiny\color{gray!35!black} walking $\cdot$ 5.68\,km $\cdot$ 60\,min $\cdot$ 06:30}%
  };                                                                                                                                                                                                                                                                                     
  \node[card, anchor=west] at (0.42, 4.10) {%
    \tikz[baseline=-0.9ex]\node[pill]{\,Fitbit\,};\ %
    {\sffamily\bfseries\tiny\color{teal!25!black} Feb\,7,\,2023}\\                                                                                                                                                                                                                       
    {\sffamily\tiny\color{gray!35!black} walking $\cdot$ 5.55\,km $\cdot$ 60\,min $\cdot$ 06:30}%
  };                                                                                                                                                                                                                                                                                     
  \node[card, anchor=west] at (0.42, 2.55) {%
    \tikz[baseline=-0.9ex]\node[pill]{\,Fitbit\,};\ %
    {\sffamily\bfseries\tiny\color{teal!25!black} Nov\,11,\,2023}\\                                                                                                                                                                                                                      
    {\sffamily\tiny\color{gray!35!black} walking $\cdot$ 5.62\,km $\cdot$ 60\,min $\cdot$ 06:30}%
  };                                                                                                                                                                                                                                                                                     
  \node[card, anchor=west] at (0.42, 1.45) {%
    \tikz[baseline=-0.9ex]\node[pill]{\,Fitbit\,};\ %
    {\sffamily\bfseries\tiny\color{teal!25!black} Dec\,31,\,2023}\\
    {\sffamily\tiny\color{gray!35!black} walking $\cdot$ 5.60\,km $\cdot$ 60\,min $\cdot$ 06:30}%
  };                                                                                                                                                                                                                                                                                     
                                                                                                                                                                                                                                                                                         
  \node[font=\sffamily\tiny\itshape, text=gray!50!black, anchor=west]
    at (0, 0.80) {(33 weekend walking logs, 3 months)};                                                                                                                                                                                                                                  
  
  \draw[-Stealth, very thick, blue!55!black]                      
    (3.60, 3.30) -- node[above, font=\sffamily\tiny\bfseries, text=blue!55!black]                                                                                                                                                                                                        
    {Condense} (4.60, 3.30);                                                                                                                                                                                                                                                            
                                                                                                                                                                                                                                                                                         
\node[header, anchor=south west] at (4.75, 5.95)
  {Schema (Habit State to be completed)};                                                                                                                                                                                                                                                                                         
  \node[schemabox, anchor=north west] at (4.75, 5.55) {%
  "habits:weekend\_neighborhood\_walk": \{\\                                                                                                                                                                                                                                      
  \hspace*{1em}"schedule": \{\\                                                                                                                                                                                                                                                          
  \hspace*{2em}"frequency\_type": \textcolor{red!70!black}{\textbf{<fill>}},\\                                                                                                                                                                                                           
  \hspace*{2em}"days\_of\_week": [\textcolor{red!70!black}{\textbf{<fill>}}, \textcolor{red!70!black}{\textbf{<fill>}}]\\
  \hspace*{1em}\},\\                                                                                                                                                                                                                                                                     
  \hspace*{1em}"timing": \{\\                                     
  \hspace*{2em}"start\_time": \textcolor{red!70!black}{\textbf{<fill>}},\\                                                                                                                                                                                                               
  \hspace*{2em}"end\_time": \textcolor{red!70!black}{\textbf{<fill>}}\\                                                                                                                                                                                                                  
  \hspace*{1em}\},\\                                                                                                                                                                                                                                                                     
  \hspace*{1em}"location": \textcolor{red!70!black}{\textbf{<fill>}}\\                                                                                                                                                                                                                   
  \}                                                                                                                                                                                                                                                                                     
  };                                                                                                                                                                                                                                                                                     
  
  \end{tikzpicture}%
  }                                                               
  \caption{One example for illustrating the state completion task at the Q4 2023 checkpoint. It requires the model to condense many time-scattered app logs (\textbf{left}) to fill the blank in a fixed JSON schema (\textbf{right}).}
  \vspace{-1em}
  \label{fig:state-completion-cloze}                                                                                                                                                                                                                                                     
  \end{wrapfigure}

%% file: NeurIPS_2026/sections/experiment.tex
\section{Experiment}

\noindent \textbf{Baselines.}
We evaluate five memory systems:
\textbf{Vanilla RAG (RAG)}~\citep{lewis2021retrievalaugmentedgenerationknowledgeintensivenlp},
\textbf{HippoRAG2}~\citep{gutiérrez2025ragmemorynonparametriccontinual}, \textbf{MemoryOS}~\citep{kang2025memoryosaiagent},
\textbf{A-Mem}~\citep{xu2025mem}, \textbf{SimpleMem}~\citep{simplemem2025}.
We additionally report an \textbf{Oracle} upper bound that is given the ground-truth evidence required to complete each task item.
We focus on query-agnostic memory systems that pre-build a persistent per-user memory and then serve many downstream task items efficiently.
For all baselines, we use \texttt{gpt-5-mini} as the answer-generation LLM and \texttt{text-embedding-3-large}  for retrieval. 
Other experiment details are provided in Appendix~\ref{app:exp_setup}. 

\subsection{Results Analysis}

\input{NeurIPS_2026/figures/finding1}
\input{NeurIPS_2026/figures/finding1_verification}
\begin{finding}
State Completion performance decreases as the horizon grows, whereas Personalized Service performance remains stable and even improves slightly by leveraging the user’s stateful trajectory.
\end{finding}

From C1 to C5, State Completion declines consistently across all memory systems. Compared with C1, the C5 score drops by 4.4 points for RAG, 6.6 for HippoRAG2, 8.9 for A-Mem, 14.2 for MemoryOS, and 16.7 for SimpleMem, as shown in Figure~\ref{fig:state_completion}. Surprisingly, Personalized Service shows no such degradation for four of the five systems: the C5 score is higher than the C1 score by 2.8 points for RAG, 4.0 for HippoRAG2, 4.9 for A-Mem, and 1.2 for MemoryOS. SimpleMem is the lone exception, declining by 1.5 points. Since both tasks use the same memory corpus, this difference suggests that the two tasks are affected by horizon growth in different ways.

To locate the source of this divergence, we elicit supporting evidence alongside each answer for RAG, HippoRAG2, and A-Mem, and compute evidence recall as the fraction of gold supporting logs cited. Recall declines over the horizon for both tasks, by 20 to 24 points for State Completion and 16 to 19 points for Personalized Service (Figure~\ref{fig:verification_finding1}a,b). This is surprising: gold retrieval degrades comparably, yet only State Completion's score drops. The resolution lies in our annotation: we label as gold only the first occurrence of each fact, but since the trajectory is stateful, that fact is typically restated later in forms not counted as gold. A drop in gold recall thus overstates information loss, as the content may still be reachable through a restatement. We therefore move beyond exact gold matching to a coarser view of retrieval quality.

To understand this, we use a coarser measure for RAG and A-Mem: the mean top-5 cosine similarity between the query and the retrieved trajectory chunks, which captures how well retrieval aligns with the user's trajectory rather than with the single annotated log. As the horizon grows, this similarity rises for both task types, since a larger corpus contains more related content. The rise is consistently larger for Personalized Service than for State Completion, and the gap between them keeps widening, from +0.065 to +0.087 for RAG and from +0.050 to +0.074 for A-Mem (Figure~\ref{fig:verification_finding1} c,d). So even though the annotated gold log becomes harder to retrieve, the longer trajectory does not hurt Personalized Service retrieval. A larger corpus instead gives its queries better aligned content to draw on, while State Completion queries gain little because they depend on the specific gold log that records the state.

The two tasks therefore respond to a longer horizon in opposite ways. Personalized Service retrieval stays robust because it benefits from the growing corpus, while State Completion retrieval degrades because it relies on one specific log that becomes harder to retrieve as the corpus grows.

\input{NeurIPS_2026/figures/finding2}
\begin{finding}
As the horizon increases, the decline in State Completion is mainly due to Preferences. Habits remain stable, while the Attributes gap appears early and then stays flat.
\end{finding}
To identify what drives the decline in State Completion, we break down the task by state family and report results for Habits, Preferences, and Attributes separately. As shown in Figure~\ref{fig:state_completion_by_category}, the three families behave differently as the horizon grows.

Preferences account for most of the decline. From C1 to C5 its score drops by 11.3 points for RAG, 15.9 for HippoRAG2, 26.5 for A-Mem, 25.2 for MemoryOS, and 23.6 for SimpleMem, and the drop grows steadily over the horizon rather than appearing all at once. We attribute this to the implicit nature of preferences: a user rarely states a preference directly, so it must be inferred from many behavioral logs across the trajectory. As the horizon grows, more logs are added, and the weak implicit signal for a given preference is increasingly blurred by unrelated logs. This makes Preferences the family most exposed to horizon growth, and its trend closely tracks the overall State Completion decline.

The other two families do not show this pattern. Habits stay roughly flat over the horizon, with C5 within 1 to 4 points of C1 for RAG, HippoRAG2, and A-Mem; MemoryOS and SimpleMem are exceptions, both dropping about 12 points. We attribute the stability of the first three to the way habits appear in the trajectory: habitual behaviors recur repeatedly and the corresponding logs are highly similar to one another, so a longer horizon mostly adds redundant rather than competing evidence, leaving retrieval largely unaffected. Attributes are roughly flat over the horizon for four of the five systems. 
Most of the drop happens at C2 and the score then stays at a similar level: RAG falls 8.7 points by C2 and is still only 7.8 points below C1 at C5, and HippoRAG2, A-Mem, and MemoryOS follow the same shape. SimpleMem is the exception, continuing to decline to 20 points below C1 by C5. 
\input{NeurIPS_2026/figures/finding3}
\begin{finding}
Retention and update are separate failure modes. No single architecture wins on both: the same mechanism that lets a system hold a stable fact often prevents it from adopting a changed one, and this trade-off differs by state family.
\end{finding}
A falling State Completion score could mean two different things: the system might be failing to remember a fact it saw long ago, or it might be failing to adopt a fact that has just changed. These are different failure modes that need different fixes, but an aggregate curve cannot distinguish them. 

We therefore split each system's score into two regimes, both anchored at C1: \textbf{retention} probes the first mode by re-asking about a fact that has stayed true since C1, while \textbf{update} probes the second by asking about a fact that has since changed. The retention lookback grows with the horizon, so the retention row reads as a forgetting curve; both regimes start at C2 (Figure~\ref{fig:finding3}).

On habit retention, A-Mem and MemoryOS go in opposite directions. A-Mem's score does not decay with the lookback; it ends 5.7 points above C2 and is higher than any other memory system at C5. We attribute this to how A-Mem stores recurring facts: each new occurrence of a habit appends a linked note that reaffirms the fact, so a longer horizon brings more reaffirmations and makes the C1 fact more retrievable, not less. MemoryOS loses about 11 points on habits over the same range, and its score swings randomly across checkpoints rather than decaying smoothly. Its hierarchical compression summarizes logs batch by batch, and when occurrences of the same habit fall into different batches, the higher-level merge step does not reliably pull them back together. The habit is therefore retained or lost depending on how batches happen to align, which matches the non-monotonic pattern we observe.
SimpleMem loses about 12 points on preference retention, the largest decline of any system. At short lookback it does well: at C2 its preference-retention score is the highest of any other memory system, so the C1 preference is clearly stored. The decline shows up later, when recursive consolidation runs over the rest of the trajectory. We see this directly in one user's case: a C1 preference for "in-depth, self-paced technical white papers and webinars" traced back to a single email log; SimpleMem retrieved that email and answered correctly at C1, C2, and C3, but by C4 the email had been pushed out of the top evidence by a growing cluster of repeated workshop and cooking logs, and at C5 SimpleMem's answer described only "hands-on, experiential, data-driven" learning with no mention of white papers or webinars. The mechanism is exactly what recursive consolidation is designed to do—merge frequent and similar surface behaviors into a higher-level summary—but the side effect is that a one-off, uncorrelated preference statement gets crowded out by the dominant behavioral cluster. Habits and attributes are not exposed to this effect: habits sit at the level of consolidation itself, and attributes are anchored to named entities that the symbolic indices preserve. Preferences are stated once and rarely repeated, so they are precisely what consolidation drowns out.

On update, HippoRAG2 stands out on changed preferences: it averages 68\% across C2--C5 at the change checkpoint, against 56--59\% for the other non-Oracle systems. All systems handle a fresh preference change well at C2 (70--89\%), but as the corpus grows behind the change, A-Mem (89\% $\to$ 39\%) and SimpleMem (89\% $\to$ 47\%) decline sharply, while HippoRAG2 (88\% $\to$ 63\%) and RAG (85\% $\to$ 59\%) both recover to within a few points of each other by C5. We attribute this divergence to how each system stores updates: HippoRAG2 and RAG are effectively append-only (each new statement enters as a separate triple or chunk alongside the prior one), so a refined preference neither erases nor merges with the previous value. A-Mem's note revision merges new statements into the existing note, SimpleMem's recursive consolidation folds related memories together, and MemoryOS's batch summarization integrates new logs into a higher-level summary. All three approaches work at short horizons but blur the boundary between new and old as more content accumulates. A-Mem shows this most dramatically with a 50-point drop, the largest decline of any system.

  \begin{finding}
For Personalized Service, habits stay near zero because current models cannot proactively select which routine a scenario calls for.
  \end{finding}       
  Table~\ref{tab:ps_habit_collapse} isolates the effect. Every system answers habit queries well under State Completion, where the habit is named as a cloze key, but the identical habit scores far lower under Personalized Service—a 43-to-50-point drop for every system (47 on average)—and habits are by far the lowest-scoring family on Personalized Service, roughly an order of magnitude below preferences (5–10\% versus 62–66\%). Because both tasks read the same memory, this is not a memory failure: the Personalized Service prompt provides only a scenario, e.g., "Sunday 9:15, coffee just poured", with no signal that this is the moment for the user's weekly budget review. The model must, on its own, summarize the user's habits, match them against the scenario's temporal and contextual cues, and surface the right routine. Current baselines are bottlenecked at this proactive routine selection step, not at recalling the routine itself, which motivates the per-stage failure analysis in Section~\ref{sec:error-analysis}.
 
\input{NeurIPS_2026/tables/ps_habit_collapse}

%% file: NeurIPS_2026/figures/finding1.tex
\begin{figure*}[t]
\centering
\includegraphics[width=1.0\textwidth]{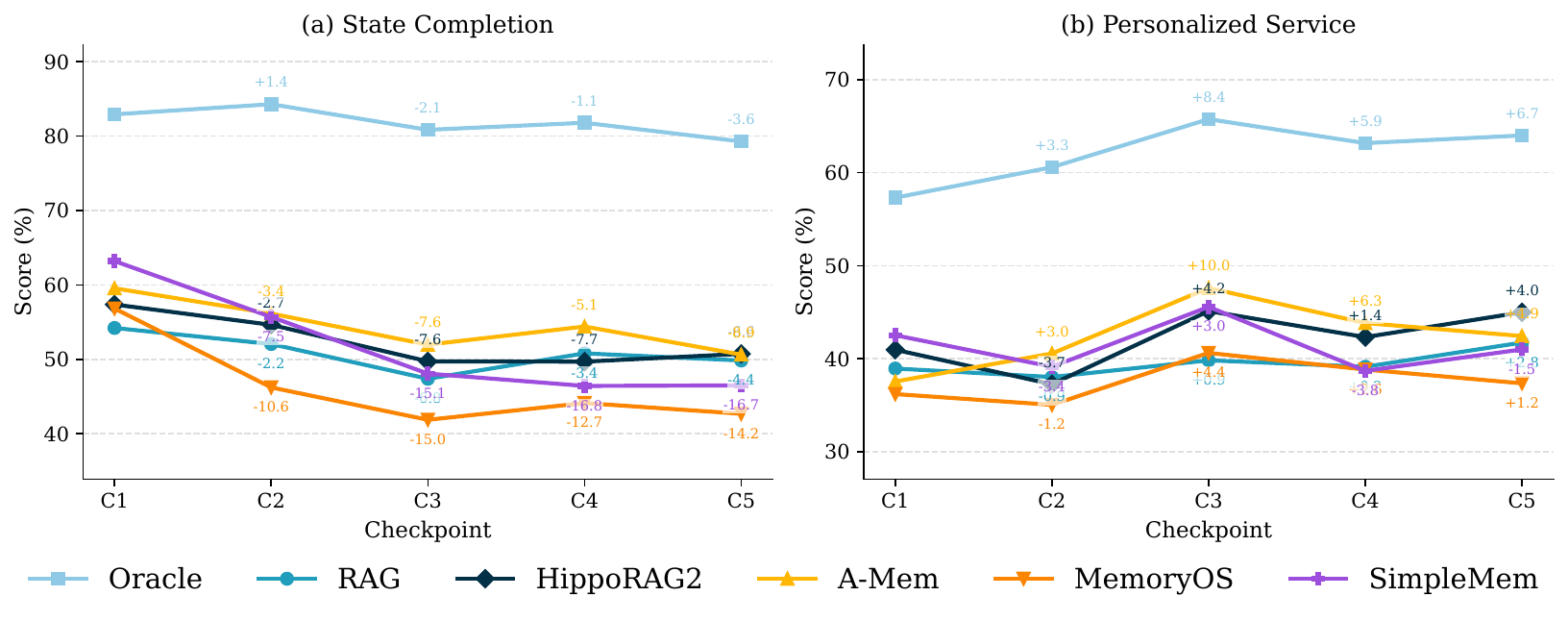}
\vspace{-1.5em}
\caption{(a) State Completion and (b) Personalized Service scores across the five quarterly checkpoints C1–C5, for all memory systems (Oracle, RAG, HippoRAG2, A-Mem, MemoryOS). Markers are annotated with their change relative to C1.}
\label{fig:state_completion}
\end{figure*}

%% file: NeurIPS_2026/figures/finding1_verification.tex
\begin{figure*}[t]
\centering
\includegraphics[width=1.0\textwidth]{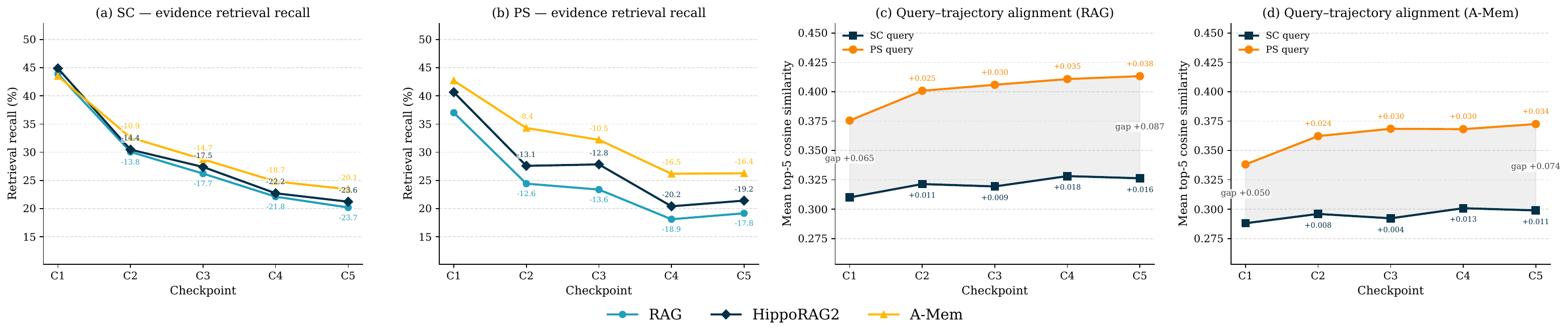}
\vspace{-1.5em}
\caption{(a, b) Evidence retrieval recall across the five quarterly checkpoints C1–C5, for State Completion (a) and Personalized Service (b), shown for the three retrieval-based memory systems (RAG, HippoRAG2, A-Mem). (c, d) Mean top-5 cosine similarity between each retrieval query and its retrieved app-log embeddings, for SC (\emph{state probe}) vs.\ PS (\emph{scenario + action}) queries, using baseline RAG and A-Mem.}
\label{fig:verification_finding1}
\end{figure*}

%% file: NeurIPS_2026/figures/finding2.tex
\begin{figure*}[t]
\centering
\includegraphics[width=1.0\textwidth]{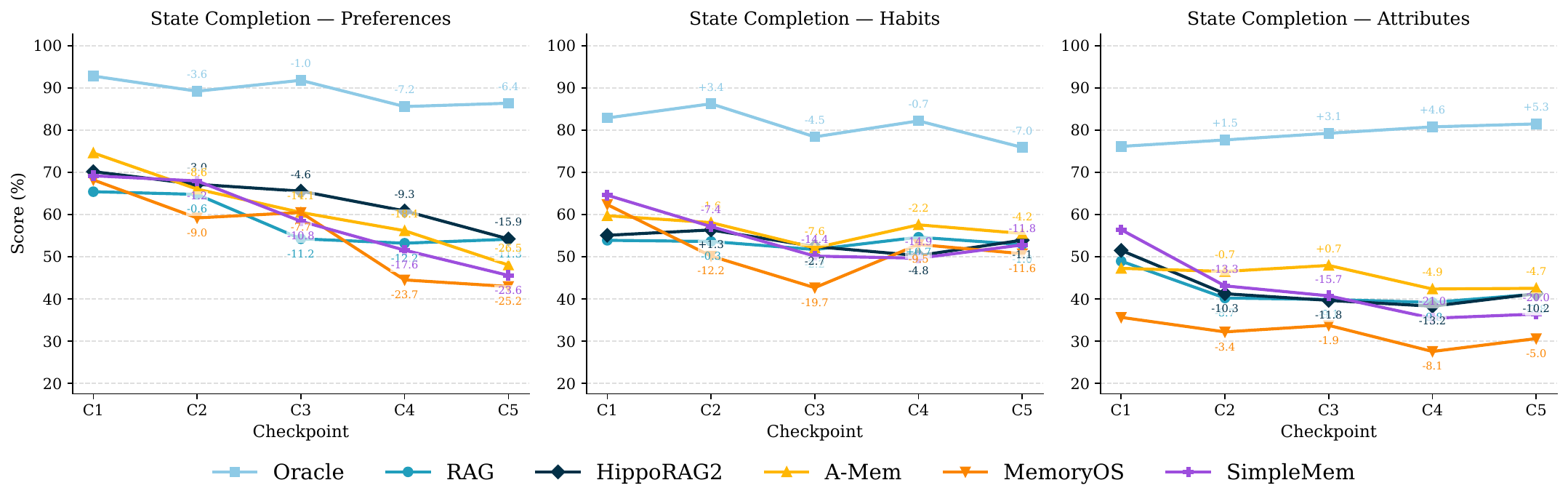}
\caption{State Completion score across the five checkpoints C1–C5, decomposed by state family: (a) Preferences, (b) Habits, (c) Attributes, for all memory systems. Markers are annotated with their change relative to C1.}
\label{fig:state_completion_by_category}
\end{figure*}

%% file: NeurIPS_2026/figures/finding3.tex
\begin{figure*}[t]
\centering
\includegraphics[width=1.0\textwidth]{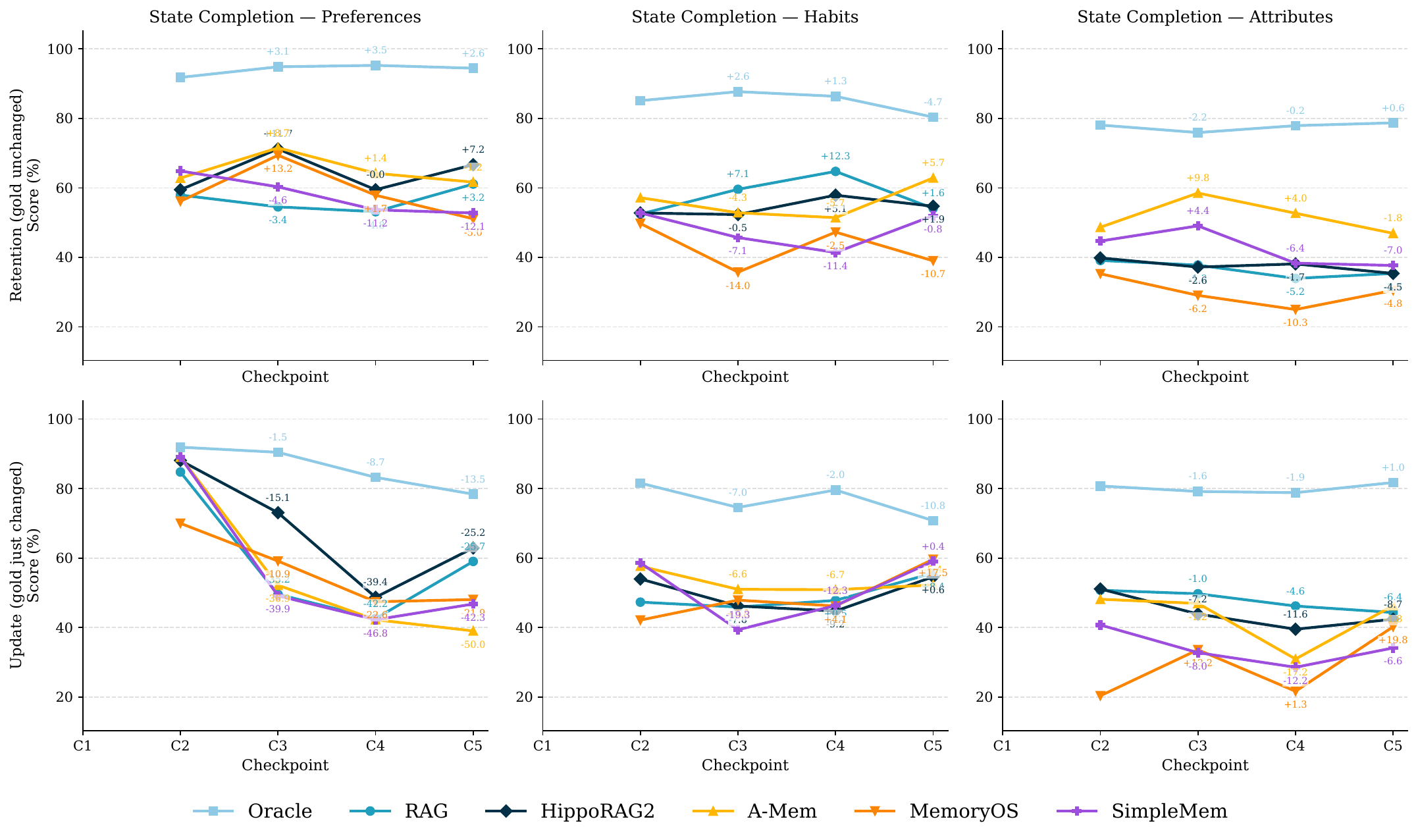}
\caption{State Completion split into long-range retention (top row) and update (bottom row), by family. For \emph{retention}, at each checkpoint $t$ we score the system on facts present at C1 and again at $t$ with the
same gold value; the lookback from C1 grows from one checkpoint at C2 to four at C5, so the top row reads as a forgetting curve. For \emph{update}, at each checkpoint we score the system on facts whose gold value just changed from their most recent previous presence.  Neither regime is defined at C1. Marker labels show the change in score from C2.}
\label{fig:finding3}
\end{figure*}

%% file: NeurIPS_2026/tables/ps_habit_collapse.tex
\begin{table}[t]
\centering\small
\caption{Habits collapse on Personalized Service (PS). For each memory system we report the habit score under State Completion (SC) and PS---the two tasks read the same memory---with the PS scores of the other two families for context.}
\label{tab:ps_habit_collapse}
\begin{tabular}{lcccc}
\toprule
 & \multicolumn{2}{c}{Habit} & \multicolumn{2}{c}{PS, other families} \\
\cmidrule(lr){2-3}\cmidrule(lr){4-5}
System & SC & PS & Pref. & Attr. \\
\midrule
Vanilla RAG & 53.5 & 5.3 & 65.0 & 43.9 \\
HippoRAG2   & 53.6 & 6.8 & 66.4 & 47.9 \\
A-Mem       & 56.9 & 7.4 & 64.4 & 49.6 \\
MemoryOS    & 52.3 & 9.6 & 62.5 & 39.1 \\
SimpleMem   & 55.0 & 7.6 & 64.7 & 46.2 \\
\bottomrule
\end{tabular}
\end{table}

%% file: NeurIPS_2026/sections/error_analysis.tex
\subsection{Error Analysis}
\label{sec:error-analysis}
Aggregate scores tell us how often a system answers correctly, but not where in the memory pipeline a failure originates. To diagnose this, we classify each failed prediction by what the memory system \emph{delivered to the answer model}, rather than by what the answer model itself produced. This isolates memory-system design failures from answer-model behaviour.

We restrict attention to items whose score is below 1.0, and analyse both State Completion (SC) and Personalized Service (PS). For each failure we collect the cited evidence that the memory system surfaced when answering (failures with no cited evidence are excluded).

Each valid failure receives exactly one label. We assign it via a
strict-precedence decision tree: starting from the broadest check
(is the cited evidence relevant at all?) and progressing to more
specific ones (does it establish the gold answer's identity? are the
gold's details matched?). The first failed check determines the
label, so the four memory-failure types are mutually exclusive by
construction, and we add a residual label for cases where every
check passes:

\begin{itemize}[leftmargin=1.4em,itemsep=0pt,parsep=0pt,topsep=0pt]
\item \textsc{Irrelevant Evidence}: the cited evidence contains no
content about what the question is asking.
\item \textsc{Identity Miss}: the evidence is on topic, but the
central identity of the gold answer (the activity for a habit, the
preference direction for a preference, the named entity for an
attribute) is not establishable from any evidence entry.
\item \textsc{Detail Miss}: the identity is establishable, but at
least one of the gold's supporting details (the specific day or time
for a habit, the named options for a preference, the year or version
for an attribute) is absent from the evidence.
\item \textsc{Conflated Evidence}: the correct identity (or its
details) is present, but a competing alternative filling the same role
is also surfaced, leaving the value ambiguous.
\item \textsc{All Clear}: every layer is present and unambiguous;
because the prediction nevertheless failed, the failure is attributable
to the answer model rather than the memory system.
\end{itemize}

We use LLM (\texttt{gpt-5.4-2026-03-05}) to perform error type classification. For each case, the LLM receives three
inputs: the actual query the memory system was invoked with (the
per-key retrieval query for SC; the scenario plus task instruction for
PS), the gold answer, and the cited evidence list. \emph{The system's
prediction is intentionally withheld from the LLM}, so the
classification reflects what the memory system delivered, not how the
answer model used it. We sample 300 failures per (system, task)
combination, stratified across the three state families. The full
classification prompts are in Appendix~\ref{app:error-taxonomy}.
\input{NeurIPS_2026/figures/error_analysis}
\setcounter{finding}{0}
\begin{finding}
Structured memory shifts the failure profile from detail
to identity.
\end{finding}
Within their State Completion failures, the four systems that build
structured representations beyond raw chunks (A-Mem, HippoRAG2,
MemoryOS, SimpleMem) all concentrate at 35--36\%
\textsc{Identity Miss}, with \textsc{Detail Miss} dropping to
28--33\%. RAG shows the opposite balance: \textsc{Detail Miss}
is its largest category at 36\%, against 31\% \textsc{Identity Miss}.
The operations that turn raw evidence into structured memory
(merging, summarizing, consolidating) preserve surrounding detail
better than raw chunks but abstract away the named anchor of a state
more often, moving where failure tends to occur. The results are also very similar across memory systems: across four very different structuring mechanisms, the
\textsc{Identity Miss} rate sits within one point of 35\%, suggesting
identity preservation is a systemic challenge of structured memory
rather than a specific design's flaw.

\begin{finding}
Hierarchical summarization concentrates failures earlier than other memory designs.
\end{finding}
MemoryOS shows higher \textsc{Irrelevant Evidence} rates than the other systems on both tasks (29\% SC against 23--24\% for the rest;
40\% PS against 35--37\%). Combined with its 36--37\%
\textsc{Identity Miss}, roughly two thirds of MemoryOS's failures
involve evidence that does not pin down the user's state. This shape
follows from how MemoryOS stores memory: by compacting many logs into
a multi-level hierarchical summary, retrieval surfaces an abstracted
view rather than the specific log a precise state query needs. The
other four architectures retain memory at finer granularities (raw
chunks, atomic triples, per-state notes, or per-event consolidations),
and a relevant fine-grained entry remains within reach more often,
leading to fewer relevance-level failures.

\begin{finding}
Improving memory is far more actionable than improving the answer model.
\end{finding}
\textsc{All Clear}---cases where the cited evidence is complete and
unambiguous but the prediction is still wrong---stays at 2--7\%
across all ten (system, task) combinations. In more than 93\% of
failures, the bottleneck is what the memory system could deliver, not
what the answer model did with it. Until this ratio shifts, accuracy
on this benchmark will track each memory system's deliverable
quality, and changes to the answer-generation LLM will leave most of
the gap untouched.

%% file: NeurIPS_2026/figures/error_analysis.tex
\begin{figure}[t]
\centering
\includegraphics[width=0.9\columnwidth]{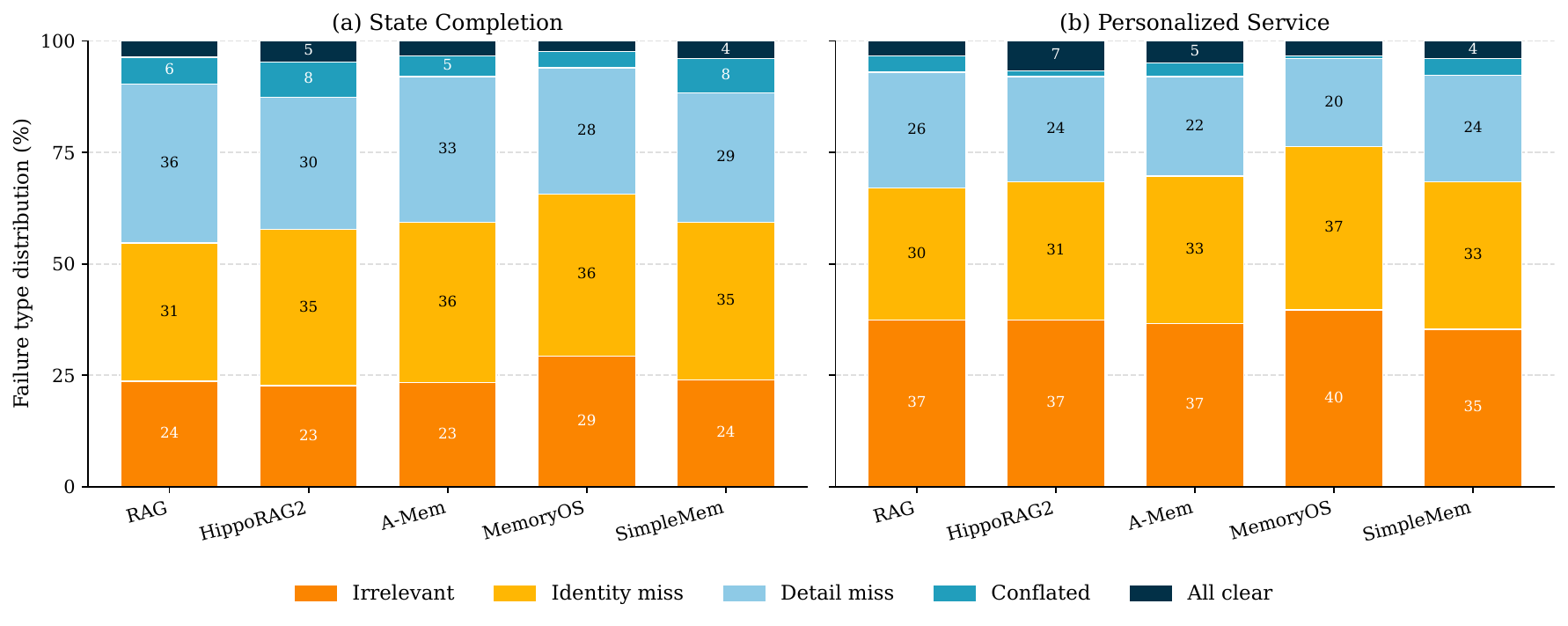}
\caption{Failure-type distribution per system on State Completion (a) and Personalized Service (b). Each bar shows 300 sampled failures (score $< 1$), stratified across the three state families and classified by the LLM judge described in Section~\ref{sec:error-analysis}. Segments $\geq 4\%$ are labeled in white/black for legibility.}
\label{fig:error-analysis}
\end{figure}

%% file: NeurIPS_2026/sections/conclusion.tex
\section{Conclusion}
We presented \muse, a synthetic benchmark for evaluating long-horizon memory in personalized LLM agents under evolving user contexts. \muse~provides 15-month, user-consistent trajectories spanning 16 applications, in which each user's profile---decomposed into attributes, habits, and preferences---evolves under explicit, causally grounded drivers, and a checkpoint protocol that probes memory as history accumulates. Benchmarking five representative memory systems exposes weaknesses that aggregate accuracy hides. As the horizon grows, systems become steadily worse at reconstructing the user's profile---especially implicit preferences that must be inferred from scattered behavior---even though acting on that profile in downstream services does not degrade. No design jointly retains facts that stay true and overwrites facts that change: the same mechanism that preserves a stable fact tends to block replacing a changed one, and the right update semantics differ by category, with in-place revision suiting habits and preferences but explicit overwriting required for named-entity attributes. Crucially, over 93\% of failures trace to what the memory system delivers rather than to the answer model, so the headroom lies in memory itself. We hope \muse~serves as a controlled testbed for memory architectures that treat retention and update as distinct objectives and reliably track how and why user profiles change over time.

%% file: NeurIPS_2026/sections/appendix.tex
\newpage
\section*{Impact Statement}
This work introduces \muse, a benchmark for evaluating long-horizon memory and personalization in real-world settings with evolving user contexts. By modeling temporally grounded user states, causal event chains, and multi-application interactions, \muse~enables systematic analysis of memory failures that are not observable in short-context or static benchmarks. The benchmark is fully synthetic and does not rely on real user data, mitigating privacy concerns while supporting reproducible research. At the same time, \muse~highlights challenges related to long-term user modeling, including temporal misalignment and overgeneralization, which are important to address before deploying memory-augmented systems in real-world personalized applications.

\section*{Limitation}
\muse~is designed for diagnostic memory evaluation in real-world settings, not for claiming full behavioral realism across all dimensions of human activity. As a synthetic benchmark, it may still miss rare behaviors and platform-specific idiosyncrasies that appear in proprietary real user logs.

\appendix
\section*{Appendix}

\section{Extended Related Work}
\label{app:full_related_work}
  \noindent \textbf{Long-Horizon Memory Benchmarks}.
Existing long-horizon memory benchmarks evaluate recall, temporal reasoning, or personalization under simplified settings, but they do not directly capture the setting we study: long-horizon personalization, where systems must recover and update user state over months of interleaved activities.
Prior work first establishes the importance of retaining information across interactions. Benchmarks such as LongMemEval~\citep{wu2024longmemeval}, LoCoMo~\citep{maharana2024evaluating}, and ConvoMem~\citep{pakhomov2025convomem} evaluate whether systems can recall prior exchanges and remain consistent across multiple dialogue sessions. However, our setting extends beyond dialogue: user state must be inferred and updated from heterogeneous application traces rather than from conversational context alone.
A newer line evaluates memory agents more directly. MemBench~\citep{tan2025membench}, MemoryAgentBench~\citep{hu2025evaluating}, and MemGym~\citep{xu2026memgym} broaden memory evaluation beyond single-shot recall, including more interactive or agentic settings. These benchmarks are important because they shift attention from long-context reading to memory as an agent capability. \muse~ studies a different bottleneck: whether a memory system can maintain an evolving user profile whose evidence is implicit, distributed, and grounded in cross-application behavior. In our setting, the target state is not simply remembered from a prior utterance; it must often be inferred from many small actions and updated when external context changes the user's attributes, habits, or preferences.
A related line of work expands the challenge from repeated interactions to long personal histories. TimelineQA~\citep{tan2023timelineqa} and CloneMem~\citep{hu2026clonemem} test whether models can recover temporally relevant evidence and reason about when events occurred. Our work further requires models to explain how such events drive later changes in user state, making causal attribution central rather than incidental.
Another nearby direction studies personalization more directly. PERSONAMEM~\citep{jiang2025know}, PersonaMem-v2~\citep{jiang2025personamem}, KnowMe-Bench~\citep{wu2026knowme} and Mem-PAL~\citep{huang2026mem} evaluate whether systems maintain persona knowledge, preferences, or personalized service behavior over long-term interactions. These works are closest in spirit to \muse~, but they remain primarily dialogue- or service-log-centered and do not jointly model multi-timescale profile evolution, causally grounded state transitions, and state-consistent multi-app behavior. In \muse~, user attributes, habits, and preferences are not fixed facts to be recalled; they are evolving states shaped by ongoing behavior across applications. This makes retrieval, causal attribution, and state tracking inseparable parts of the same problem.


\noindent \textbf{Memory Systems for Agents}.
Memory systems for long-horizon agents can be compared along representation, retrieval, update operations, and write/update control.
Retrieval-first methods such as Vanilla RAG~\citep{lewis2021retrievalaugmentedgenerationknowledgeintensivenlp}, HippoRAG2~\citep{gutiérrez2025ragmemorynonparametriccontinual}, and Zep~\citep{rasmussen2025zep} improve evidence access via vector or graph retrieval, but typically rely on heuristic update rules and remain weak on temporally consistent state updates.
Structured external-memory systems, including MemGPT/Letta~\citep{packer2023memgpt}, Mem0~\citep{chhikara2025mem0}, A-Mem~\citep{xu2025mem}, O-MEM~\citep{wang2025mem}, MemoryOS~\citep{kang2025memoryosaiagent}, and Nemori~\citep{nan2025nemori}, introduce explicit memory objects with operations such as summarization, consolidation, overwrite, and eviction, improving controllability but still leaving open how reliably systems track evolving user state under sparse evidence.
Recent systems also explore more adaptive organization for personalized interaction. RMM~\citep{tan2025prospect} and PRIME~\citep{zhang2025prime} are representative of this direction: they organize long-term interaction history into more reflective or cognitively motivated memory structures for personalization. These systems are related to \muse~'s emphasis on long-term user modeling, but they are primarily evaluated in dialogue or personalization settings where the underlying user state is less explicitly tied to cross-app causal transitions. \muse~ provides a complementary testbed for asking whether such memory mechanisms can handle profile changes that are not directly stated, are triggered by external context, and must be inferred from distributed behavioral traces.
Neural or learned-policy memory methods such as LongMem~\citep{wang2023augmenting}, MemoryLLM~\citep{wang2024memoryllm}, M+~\citep{wang2025m+}, Memory-R1~\citep{yan2025memory}, MemAgent~\citep{yu2025memagent}, and AgeMem~\citep{yu2026agentic} learn what and when to write or update, but are often evaluated without explicit supervision on cross-event causal state transitions in multi-app trajectories.
Our work is complementary to these families: rather than proposing another memory algorithm, \muse~provides a controlled proxy for long-horizon memory challenges in real-world applications with unified tests for retrieval, causal attribution, dynamic state tracking, retention of stable facts, and overwriting of changed facts.

\section{Base User Profile Schema and Construction}
\label{app:base_profile}

This appendix describes the \textbf{base user profile} used in \muse~to anchor user-specific generation.
The base profile captures \emph{stable or slowly-changing} factors that shape behavior across all life domains (e.g., work, finances, health),
and serves as the conditioning context for (i) domain-specific dynamic state trajectories, (ii) world background sampling, and (iii) downstream event/log synthesis.

\subsection{Schema}
\label{app:base_profile_schema}

We define the base profile $b$ as a structured record with four groups of fields:
\textbf{(1) demographics \& locale, (2) work \& socioeconomic status, (3) household \& constraints, (4) cognitive/behavioral traits}.
Table~\ref{tab:base_profile_fields} enumerates all fields and their allowed value spaces.
The schema is intentionally \emph{typed} (categorical, bounded integer, and short string) to support validation, controlled sampling, and auditable conditioning.

\subsection{Construction Procedure}
\label{app:base_profile_construction}

We construct $b$ in three steps: schema instantiation, constraint-aware sampling, and consistency validation.

\paragraph{Step 1: Schema instantiation.}
Starting from a seed persona (e.g., PersonaHub~\cite{ge2025scalingsyntheticdatacreation}), we map persona descriptions into the base schema by extracting or inferring required fields
(e.g., age, occupation, education level, and digital literacy).
When the seed persona does not specify a required field, we synthesize a plausible value while preserving coherence with the persona narrative
(e.g., ``Graduate Student'' implies \textsc{Student} employment status and typically correlates with \textsc{Bachelor/Master} education).

\paragraph{Step 2: Constraint-aware sampling.}
For fields not explicitly determined by the persona, we sample values subject to hard constraints and soft plausibility rules.
Hard constraints prevent invalid combinations (e.g., age bounds; weekly work hours in $[0,70]$; household size in $[1,8]$).
Soft rules encourage plausible correlations, such as:
(i) employment status constraining weekly work hours,
(ii) education level correlating with typical occupation/industry,
and (iii) income band and financial buffer aligning with employment type and housing status.
We implement these rules as rejection sampling or rule-based repairs when violations are detected.

\paragraph{Step 3: Consistency validation.}
We validate each generated base profile with a set of deterministic checks
(type checks, range checks, and cross-field consistency checks).

\subsection{Role in the Overall Pipeline}
\label{app:base_profile_role}

The base profile $b$ anchors all downstream synthesis stages:
(1) it conditions the sampling of world background $w_k^t$ (exogenous drivers) to make changes user-specific;
(2) it constrains feasible dynamic state items in each life domain;
and (3) it provides stable context during intent-conditioned chain generation and log grounding
(e.g., language choices, digital literacy, planning orientation, and risk tolerance affect action patterns and decision processes).
In this way, $b$ reduces templated generation and helps maintain coherence across domains and over long horizons.
\begin{table*}[t]
\centering
\small
\setlength{\tabcolsep}{4pt}
\renewcommand{\arraystretch}{1.08}

\begin{tabularx}{\textwidth}{
  >{\raggedright\arraybackslash}p{2.6cm}
  >{\raggedright\arraybackslash}p{4.2cm}
  >{\raggedright\arraybackslash}X
}
\toprule
\textbf{Field Group} & \textbf{Field} & \textbf{Value space (examples)} \\
\midrule
Demographics \& locale
& \texttt{age} & Integer in $[18,85]$ \\
& \texttt{gender} & \textsc{Male} $|$ \textsc{Female} $|$ \textsc{Non\_binary} $|$ \textsc{Other} \\
& \texttt{location\_type} & \textsc{Urban\_Metropolis} $|$ \textsc{Urban} $|$ \textsc{Suburban} $|$ \textsc{Town} $|$ \textsc{Rural} \\
& \texttt{location\_region} & Short string, e.g., ``San Francisco Bay Area'', ``Beijing'', ``London'' \\
& \texttt{languages} & List of ISO language codes, e.g., \texttt{en}, \texttt{zh}, \texttt{es} \\
\midrule
Work \& socioeconomic
& \texttt{occupation} & Specific job title, e.g., ``Software Engineer'', ``Retail Cashier'' \\
& \texttt{employment\_status} & \textsc{Full\_time} $|$ \textsc{Part\_time} $|$ \textsc{Self\_employed} $|$ \textsc{Unemployed} $|$ \textsc{Student} $|$ \textsc{Retired} $|$ \textsc{Caregiver} \\
& \texttt{industry} & Specific industry, e.g., ``Technology'', ``Healthcare'' \\
& \texttt{weekly\_work\_hours} & Integer in $[0,70]$ \\
& \texttt{education\_level} & \textsc{Middle\_School} $|$ \textsc{High\_School} $|$ \textsc{Vocational} $|$ \textsc{Associate} $|$ \textsc{Bachelor} $|$ \textsc{Master} $|$ \textsc{Doctorate} \\
& \texttt{income\_band} & \textsc{Low} $|$ \textsc{Lower\_Mid} $|$ \textsc{Mid} $|$ \textsc{Upper\_Mid} $|$ \textsc{High} \\
& \texttt{financial\_buffer\_months} & \textsc{0} $|$ \textsc{<1} $|$ \textsc{1-3} $|$ \textsc{3-6} $|$ \textsc{6-12} $|$ \textsc{12+} \\
& \texttt{housing\_status} & \textsc{Renting} $|$ \textsc{Home\_Owner} $|$ \textsc{Living\_with\_Family} $|$ \textsc{Dormitory} $|$ \textsc{Other} \\
\midrule
Household \& constraints
& \texttt{household\_size} & Integer in $[1,8]$ \\
& \texttt{household\_composition} & Short string, e.g., ``Lives alone'', ``Married with 2 children'' \\
& \texttt{caregiving\_load} & \textsc{None} $|$ \textsc{Light} $|$ \textsc{Moderate} $|$ \textsc{High} \\
& \texttt{health\_constraint\_level} & \textsc{None} $|$ \textsc{Mild} $|$ \textsc{Moderate} $|$ \textsc{Severe} \\
\midrule
Cognitive/behavioral traits
& \texttt{digital\_literacy} & \textsc{Low} $|$ \textsc{Basic} $|$ \textsc{Advanced} $|$ \textsc{Native} \\
& \texttt{planning\_orientation} & \textsc{Present\_biased} $|$ \textsc{Balanced} $|$ \textsc{Future\_oriented} \\
& \texttt{risk\_tolerance} & \textsc{Risk\_Averse} $|$ \textsc{Neutral} $|$ \textsc{Risk\_Seeking} \\
\bottomrule
\end{tabularx}

\caption{Base user profile fields and value spaces. This base profile is stable and conditions all downstream domain-level dynamic state generation.}
\label{tab:base_profile_fields}
\end{table*}

\section{Data Schema and Representation}
\label{app:representation_of_dynamic_profile}
\paragraph{Attributes.}
Each attribute $a_i$ takes one of two value types:
\begin{itemize}
    \item \textit{Single string}: e.g., \texttt{gym\_membership: "Planet Fitness"}
    \item \textit{List of strings}: e.g., \texttt{dietary\_restrictions: ["gluten-free", "dairy-free"]}
\end{itemize}

\paragraph{Habits.}
Each habit $h_j$ is a named routine (the behavior itself, e.g., \texttt{batch\_cooking}) whose value records $(\textit{schedule}, \textit{timing}, \textit{location})$:
\begin{itemize}
    \item \textit{schedule}: Frequency (daily, weekly, biweekly, monthly)
    \item \textit{timing}: Start and end times
    \item \textit{location}: Where the habit occurs (e.g., "local park", "home kitchen")
\end{itemize}
Each habit also carries an auxiliary \textit{priority} field (e.g., high/medium/low) that is used only during cross-domain conflict resolution (Appendix~\ref{app:validation_cross_domain}); it is not part of any evaluated answer.
Example: \texttt{batch\_cooking} $\rightarrow$ (\textit{weekly}, \textit{Sunday evening 18:00--19:00}, \textit{home})

\paragraph{Preferences.}
Each preference $p_m$ is a pair $(\textit{statement}, \textit{signal})$:
\begin{itemize}
    \item \textit{statement}: A declarative preference (e.g., "prefers independent films over blockbusters")
    \item \textit{signal}: An observable behavioral indicator (e.g., "80\%+ of movie selections are from A24 or similar studios")
\end{itemize}
\section{Life Domain Taxonomy}
\label{app:life_domains}

This appendix describes the life-domain taxonomy adopted in \muse~and how it relates to prior well-being and quality-of-life frameworks.
Life domains provide the \textbf{organizational backbone} for our dynamic user modeling: for each user, we maintain a domain-specific dynamic state trajectory $\{s_k^t\}$, and generate intent-conditioned event chains and grounded interaction logs \emph{per domain and per time window}.
This design supports (i) modular state evolution, (ii) clearer causal attribution of changes, and (iii) cross-domain consistency checks when merging trajectories.

\subsection{Mapping to Prior Frameworks}
\label{app:domain_mapping}
\input{NeurIPS_2026/tables/life_domain}

\input{NeurIPS_2026/tables/life_domain_definition}

We derive our taxonomy by aligning commonly used life-domain frameworks from subjective well-being (SWB) literature and the OECD Better Life Index with the functional areas of everyday digital behavior.
Table~\ref{tab:domain_mapping} reports the mapping from our adopted domains to corresponding categories in the OECD Better Life Index and to the domain-satisfaction perspective within the subjective well-being framework discussed by Diener et al.~\citep{diener2018advances}.
While many domains have direct counterparts (e.g., \emph{Health} and \emph{Jobs/Education}), we also include a dedicated domain for \emph{Leisure \& Media Consumption} to reflect contemporary multi-app behaviors that are not explicitly separated in some classical frameworks.

\subsection{Adopted Domains and Definitions}
\label{app:domain_definitions}

We use six life domains, each representing a major functional area of everyday life in which user attributes, behaviors, and preferences are naturally organized.
Table~\ref{tab:life_domain_taxonomy} provides operational definitions for each domain.
In our pipeline, each domain $k$ maintains a dynamic state $s_k^t=(A_k^t,H_k^t,P_k^t)$ (attributes, habits, preferences) that evolves across time windows.
Domain boundaries also constrain event-chain generation: each non-empty domain-level delta $\Delta_k^t$ must be realized by at least one intent-conditioned event chain within the same window, while a subset of unchanged states are surfaced stochastically to keep stable context repeatedly observable.




\section{World Background Generation}
\label{app:world background}
We constructed a comprehensive world background by incorporating physical environments (e.g., seasonal changes), social rhythms (e.g., festivals and holidays), key events (e.g., political or sports milestones), and prevailing trends (e.g., public sentiment and technological zeitgeist). This setup captures the texture of daily life through recurrent details like seasonal weather fluctuations and the NBA season, while anchoring the timeline with specific historical milestones, such as the release of a groundbreaking technology, major election outcomes, or unforeseen natural disasters.

Our world background is designed to be (i) \textbf{user-conditioned}: it is grounded in each persona's geographic and cultural context and is generated \emph{per life domain}, so the same calendar period yields different drivers for different users; (ii) \textbf{exogenous}: it narrates only the world, never the user's own actions, feelings, or choices, so it serves as an external cause of change rather than revealing the user's profile directly; and (iii) \textbf{coverage-complete across time}: one entry spans all five contemporaneous windows ($w_0$--$w_4$) with consistent physical, social, event, and trend layers, giving every state transition a datable, plausible trigger. The full generation prompt is provided in Appendix~\ref{app:data_prompt}.

To illustrate the format and content of a single \emph{world background} entry, Figure~\ref{fig:world-background-example} shows the verbatim Social~\& Community domain context generated for one persona. The full entry decomposes into five contemporaneous time windows ($w_0$--$w_4$) spanning Q4~2023 through Q4~2024; for space we display only the first and last windows and omit the three intermediate windows. 

\begin{tcolorbox}[
  enhanced,
  breakable,
  listing only,
  colback=green!6,
  colframe=green!45!black,
  title=\textbf{World background example --- Social \& Community},
  fonttitle=\footnotesize,
  listing options={
    basicstyle=\scriptsize\ttfamily,
    breaklines=true,
    breakatwhitespace=true,
    columns=fullflexible,
    keepspaces=true,
    showstringspaces=false,
    extendedchars=true,
    inputencoding=utf8,
  },
  arc=2pt,
  boxrule=0.6pt,
  left=4pt, right=4pt, top=4pt, bottom=4pt,
]
[Window w0]  2023-10-01 -- 2023-12-31

Autumn tightens over the Pittsburgh metro as the hills turn copper and the rivers run slate-gray under shorter days. Suburban streets swing between crisp, dry afternoons and sudden cold rain, with early frosts arriving before Thanksgiving and the first real snow threats showing up in December. [...]

National and global events bleed into community conversation in visible ways. The Israel-Hamas war ignites after the October 7, 2023 attacks, prompting vigils, fundraisers, and heightened tension across faith communities and civic spaces.  [...] The broader zeitgeist in late 2023 carries a mix of fatigue and cautious optimism---costs still feel high, schedules feel packed, and people increasingly coordinate social plans through group chats, school apps, and neighborhood platforms rather than spontaneous drop-ins.

----------------------------------------------------------------
\\
\noindent[Window w1] 2024-01-01 -- 2024-03-31\par
\vspace{0.5em}
\noindent[Window w2] 2024-04-01 -- 2024-07-01 \par 
\vspace{0.5em}
\noindent[Window w3] 2024-07-02 -- 2024-09-30\par

----------------------------------------------------------------

[Window w4]  2024-10-01 -- 2024-12-31

Fall returns with sharp, clear mornings and earlier sunsets that pull social life back indoors. The Pittsburgh landscape turns dramatic again---river valleys fill with fog, trees shift to deep reds, and the first hard cold snaps bring back the familiar commute squeeze at bridges and tunnels. [...]

Holiday season in Pittsburgh has a clear kickoff: Light Up Night on November 23, 2024 launches weeks of downtown programming, lighting displays, and family-oriented outings that compete with cold winds off the rivers. [...] Across the window, the prevailing trend is "structured togetherness": more planned gatherings, more sign-ups and ticketed events, more coordination through apps and shared calendars, and a steady undercurrent of conversation about safety, costs, and how communities keep showing up for one another even when the year feels politically and emotionally charged.
\end{tcolorbox}
\captionof{figure}{LLM-generated world background for \texttt{user\_001}'s Social \& Community domain. The full entry contains five contemporaneous time windows ($w_0$--$w_4$); we display only $w_0$ and $w_4$ and abbreviate the three intermediate windows.} 
\label{fig:world-background-example}

\section{Details of Multi-Timescale User Profile Construction Validation}
\label{app:validation-multitimescale-details}

\paragraph{Overview.}
Stage~1 of \muse~generates a domain-level dynamic profile for each life domain, consisting of an \texttt{initial\_state} and a sequence of windowed deltas (cf.\ Appendix~\ref{app:representation_of_dynamic_profile}).
In pilot generation, we found that most profile errors fall into a small set of recurring patterns; we therefore codify a \textbf{7-type error taxonomy} and validate profiles with a programmatic detector followed by LLM-based patch repair.
Concretely, detectors emit a list of \emph{localized issues} (with IDs and JSON paths), and the repair model outputs a list of JSON patches (actions: \texttt{replace}, \texttt{append}, \texttt{add\_key}, \texttt{remove}) that are applied to the profile.
All repairs are required to maintain \textbf{schema validity} and \textbf{narrative consistency}: when edits change a state or delta, the corresponding \texttt{summary} and/or \texttt{reason} fields are updated to match the revised trajectory.

\subsection{In-Domain Validation (Rules 1--5)}
\label{app:validation_in_domain}
We first validate each domain profile independently. The five in-domain error types are encoded as Rules~1--5:

\paragraph{Rule 1: Short-term changes must have follow-ups (missing reversion).}
Temporary, exogenous drivers (e.g., seasons, holidays, short-lived events) should not cause permanent state flips unless explicitly justified.
\textbf{Detection:} we flag changes whose \texttt{reason} indicates a short-term factor (e.g., ``winter'', ``holiday'', ``Olympics'') but no later window contains an appropriate rollback (e.g., \texttt{drop}/\texttt{adjust} for habits; \texttt{modify}/\texttt{remove} for attributes; \texttt{refine}/\texttt{shift} for preferences).
\textbf{Repair:} we add a follow-up delta in a plausible later window and update that window's \texttt{summary} to mention the rollback; if the change is meant to persist, we rewrite the \texttt{reason} to make the permanence explicit.

\paragraph{Rule 2: Modification operations require prior existence (invalid transitions).}
Edits must be well-typed and only operate on items that exist at the time of the edit.
\textbf{Detection:} for each delta in window $t$, we verify that the target item exists either in \texttt{initial\_state} or was introduced in earlier windows. We also enforce operation semantics: singular attributes only allow \texttt{modify}; collection attributes only allow \texttt{add}/\texttt{remove}; habits only allow \texttt{adjust}/\texttt{drop} if the habit exists; preferences only allow \texttt{shift}/\texttt{refine} if the preference exists.
\textbf{Repair:} we either (i) add the missing item with a baseline value into \texttt{initial\_state} (plus a summary update), or (ii) rewrite the invalid delta to a valid operation type and adjust its \texttt{reason} accordingly.

\paragraph{Rule 3: Required fields and schema completeness.}
Profiles must satisfy the structured schema (including nested habit and preference objects).
\textbf{Detection:} we check required keys at each level: each window has \texttt{window\_description} and \texttt{summary}; \texttt{initial\_state} has \texttt{summary}; \texttt{user\_attributes\_state} contains both \texttt{singular} and \texttt{collections} (possibly empty); each habit object contains a complete \texttt{schedule} (including all required fields for its \texttt{frequency\_type}) and a complete \texttt{timing} (\texttt{start\_time}, \texttt{end\_time}); each preference contains a \texttt{statement} and 2--4 concrete \texttt{signals}; each change contains a non-empty \texttt{reason}; dropped habits use JSON \texttt{null} (not a string like ``none'').
\textbf{Repair:} we add missing keys and fill plausible values conditioned on the user profile and world background, and we update summaries/reasons to stay consistent with the completed structure.

\paragraph{Rule 4: Essential items must be initialized if evolved (implicit baseline).}
Some items are expected to exist as plausible baselines; later ``evolution'' implies a prior existence.
\textbf{Detection:} if a later window evolves an item (e.g., \texttt{modify}/\texttt{adjust}/\texttt{shift}) but the item is absent from \texttt{initial\_state}, we flag it when the item is deemed essential for the user and domain (e.g., a modern adult having at least one phone/device; basic financial accounts/payment methods).
\textbf{Repair:} we add the item to \texttt{initial\_state} with a coherent baseline value (aligned with income, digital literacy, and locale), update \texttt{initial\_state.summary}, and apply cascade fixes if the later deltas need to be rewritten (e.g., removing an ``add first phone'' event once a baseline phone exists).

\paragraph{Rule 5: Temporal feasibility and intra-domain time conflicts.}
Within a domain and window, habits must form a feasible schedule.
\textbf{Detection:} we check (i) overlaps between habit \texttt{timing} intervals, and (ii) travel feasibility: if two habits occur at different locations, we require a minimum gap (e.g., 30 minutes) between the end of one and the start of the next.
We additionally verify time windows are chronologically ordered and non-overlapping.
\textbf{Repair:} we minimally shift a habit's \texttt{timing} and/or \texttt{schedule} to resolve conflicts and update associated \texttt{reason}/\texttt{summary} fields. If a habit must be removed, we scan subsequent windows and delete or rewrite downstream deltas that reference the removed habit (cascade handling).

\subsection{Cross-Domain Validation}
\label{app:validation_cross_domain}
After in-domain fixes, we validate \emph{global} consistency across domains. This adds two cross-domain error types:

\paragraph{Error type 6: Cross-domain attribute key alignment and attribute conflicts.}
Different domains may refer to the same underlying attribute using different key names (e.g., \texttt{main\_car} vs.\ \texttt{primary\_vehicle}), and aligned attributes may still disagree in value.
\textbf{Key alignment:} we first produce canonical mappings by aligning only strict synonyms with the same attribute type (singular vs.\ collection), and we rewrite keys to canonical names.
\textbf{Conflict detection:} after alignment, for any shared singular attribute that appears in multiple domains within the same window, we flag mismatched values; for collection attributes, we flag duplicate items that appear across domains in the same window.
\textbf{Repair:} for singular conflicts, we pick the most coherent value (using domain authority and the base profile as tie-breakers) and align domains to that value in the specific window; for collection duplicates, we de-duplicate by keeping one authoritative instance and removing duplicates from the other domain(s) (without merging whole collections across domains).

\paragraph{Error type 7: Cross-domain temporal conflicts (global schedule feasibility).}
Even if each domain is internally feasible, habits across domains can still collide (e.g., a work meeting overlapping with a health routine).
\textbf{Detection:} we construct a per-window schedule across \emph{all} domain habit states and deltas and detect overlaps using the same location-aware feasibility rules as Rule~5.
\textbf{Repair:} we resolve conflicts by modifying only a designated \emph{focus habit} in the current window (typically the habit involved in the most conflicts), preferring schedule/timing adjustments; if dropping is unavoidable, we apply cascade handling by removing or rewriting all downstream deltas and summaries that reference the dropped habit.

\section{Log Schema}
\label{app:app_schema}
\subsection{App APIs Used}
We model app usage as discrete API calls to a fixed set of consumer apps. Each app-log entry records an \texttt{app\_name} and an \texttt{api\_name} drawn from the following API surfaces:
\begin{itemize}
    \item \textbf{Amazon}: \texttt{SearchProducts}, \texttt{ShowProduct}, \texttt{AddToCart}, \texttt{ShowCart}, \texttt{ShowWishlist}, \texttt{AddToWishlist}, \texttt{Checkout}, \texttt{WriteReview}.
    \item \textbf{Spotify}: \texttt{SearchSongs}, \texttt{PlaySong}, \texttt{AddToPlaylist}, \texttt{FollowArtist}.
    \item \textbf{Fitbit}: \texttt{LogWorkout}, \texttt{RecordActivity}, \texttt{SyncDevice}, \texttt{SetGoals}.
    \item \textbf{Chase}: \texttt{GetBalance}, \texttt{GetTransactions}, \texttt{SearchTransactions}, \texttt{TransferMoney}, \texttt{PayBill}.
    \item \textbf{Robinhood}: \texttt{GetPortfolio}, \texttt{GetWatchlist}, \texttt{SearchStocks}, \texttt{GetStockQuote}, \texttt{BuyStock}, \texttt{SellStock}.
    \item \textbf{WhatsApp}: \texttt{GetMessages}, \texttt{SendMessage}, \texttt{SendMedia}.
    \item \textbf{Gmail}: \texttt{GetInbox}, \texttt{ReadEmail}, \texttt{SendEmail}, \texttt{ReplyEmail}.
    \item \textbf{LinkedIn}: \texttt{UpdateProfile}, \texttt{AddExperience}, \texttt{AddSkill}, \texttt{PostUpdate}, \texttt{GetFeed}, \texttt{LikePost}, \texttt{CommentOnPost}, \texttt{SearchJobs}, \texttt{ApplyJob}, \texttt{SendConnectionRequest}.
    \item \textbf{Notion}: \texttt{GetPages}, \texttt{CreatePage}, \texttt{UpdatePage}, \texttt{SearchContent}, \texttt{CreateDatabaseEntry}.
    \item \textbf{Netflix}: \texttt{SearchContent}, \texttt{ShowTitle}, \texttt{PlayContent}, \texttt{AddToMyList}, \texttt{RateContent}.
    \item \textbf{Goodreads}: \texttt{SearchBooks}, \texttt{ShowBook}, \texttt{AddToShelf}, \texttt{RateBook}, \texttt{WriteReview}.
    \item \textbf{Instagram}: \texttt{CreatePost}, \texttt{PostStory}, \texttt{LikePost}, \texttt{CommentOnPost}, \texttt{SendDirectMessage}, \texttt{FollowUser}, \texttt{UnfollowUser}, \texttt{GetFollowing}.
    \item \textbf{Google}: \texttt{Search}, \texttt{ClickResult}.
    \item \textbf{LLM Assistant}: \texttt{CreateConversation}, \texttt{ContinueConversation}.
    \item \textbf{Google Maps}: \texttt{ShareLocation}, \texttt{CheckIn}, \texttt{GetDirections}, \texttt{SearchPlaces}.
    \item \textbf{UberEats}: \texttt{SearchRestaurants}, \texttt{GetMenu}, \texttt{PlaceOrder}, \texttt{GetOrderHistory}.
\end{itemize}

\subsection{Input-Output Schema for Each API}
For every API $\langle \texttt{app\_name}, \texttt{api\_name} \rangle$, we define typed request and response schemas and expose them as JSON Schema constraints during log generation.
    Concretely, Stage~3 prompts an LLM to produce a JSON object with keys \texttt{input} and \texttt{output}, validates them against the corresponding schemas, and stores them in the log as \texttt{request} and \texttt{response}.
    Each app-log entry has the following top-level structure:
    \begin{itemize}
        \item identifiers: \texttt{app\_log\_id}, \texttt{event\_id}, \texttt{atomic\_event\_id}
        \item API call: \texttt{timestamp}, \texttt{app\_name}, \texttt{api\_name}, \texttt{request}, \texttt{response}
    \item traceability: \texttt{metadata} (domain/window/chain and token-usage accounting) and \texttt{golden\_evidence} (which state items this call supports)
    \end{itemize}

\section{Data Synthesis, Validation, and Task Construction Details}
\label{app:data_synthesis_task_construction}

This appendix section describes the construction settings used to instantiate \muse~and the validation pipeline that turns synthesized trajectories into evaluation-ready checkpoint tasks.
The goal of this pipeline is not only to produce realistic app logs, but also to ensure that every benchmark item is answerable from checkpoint-bounded evidence and that the task prompt itself does not reveal the personalized state being evaluated.

\subsection{Synthesis Model and Structured Generation}
\label{app:construction_model_cost}
Our current \muse~instantiation uses \texttt{gemini-3-flash-preview} as the construction model for data synthesis and evaluation-task authoring.
All construction prompts are issued with structured JSON output constraints, and generated objects are parsed and checked against the expected schema before they are consumed by later stages.
The implementation is provider-agnostic: the same construction pipeline can be run through Gemini-style \texttt{generate\_content} endpoints or OpenAI-compatible \texttt{/chat/completions} endpoints, as long as the backend supports reliable JSON generation.

The synthesis pipeline uses the same model family across the main construction stages: base-profile expansion, world-background generation, domain-state evolution, intent-conditioned event-chain generation, and app-log grounding.
At each stage, the model receives only the structured context needed for that stage, and the output is carried forward as typed JSON rather than as free-form prose.
This design makes each generated log traceable to a user profile state, an event-chain intent, and an app/API schema.

\subsection{Application-Level State During Log Grounding}
Each app maintains a per-user application state that is updated deterministically after every generated API call and then used as conditioning context for subsequent calls.
We maintain (i) a bounded \texttt{api\_call\_history} and \texttt{last\_api\_call} across apps, and (ii) app-specific state sufficient for continuity, such as Amazon cart and order history, Spotify playlists and play history, Gmail inbox/sent messages, WhatsApp message history, Google search history, Maps saved places and directions, and UberEats orders.
To support long chains and resumability, we checkpoint the latest per-app states and recent chain logs, and feed the current app state back into the next generation call.

\subsection{Checkpoint State Validation}
After trajectory synthesis, we build temporal checkpoints from the prefix of app logs available at each checkpoint.
For every candidate state, the raw checkpoint artifact contains \texttt{expected\_snapshot\_state}, which records the candidate current state, and \texttt{state\_observability}, which records the checkpoint-bounded evidence logs associated with that state.
Because a latent state is not necessarily a fair evaluation target, we run an independent state-validation stage before task construction.

State validation is field-level.
First, deterministic filters remove fields that should not become answer targets, including preference support fields such as \texttt{signal}/\texttt{signals} and authoring-only scheduling metadata such as \texttt{priority}, \texttt{schedule\_date}, and \texttt{schedule\_dates}.
The remaining candidate fields are checked against evidence logs available at that checkpoint.
The validator decides whether each candidate field is supported by the local evidence, and only accepted fields are retained in \texttt{validated\_snapshot\_state}.
Thus, if a structured state is only partially supported, the benchmark keeps the supported fields rather than accepting or rejecting the whole state as a single unit.
The resulting \texttt{validated\_snapshot\_state} is the sole source of gold state information for downstream task construction.

\input{NeurIPS_2026/tables/state_filtering_summary}
\input{NeurIPS_2026/tables/state_validation_audit}

\subsection{Evaluation Task Construction}
We construct two prebuilt task packs from the validated checkpoint states.
For \textsc{State Completion}, the builder creates one item per retained state key.
Each item contains a fixed question, a retrieval query, and an answer template derived from the validated current-state projection.
When a state record is transition-like, the task target is deterministically projected to the current value at the checkpoint before the item is authored.
No task item is authored from unvalidated fields.

For \textsc{Personalized Service}, the builder turns the same validated states into service-facing tasks.
The mapping is fixed by state family: habits become user-communication tasks, preferences become search/filter completion tasks, and attributes become action-configuration tasks.
Each item contains a leakage-safe scenario, a fixed task instruction, and either a reference natural-language assistant message or a structured reference output.
The scenario is allowed to describe the current service moment, but it must not reveal the user-specific state that the memory system is supposed to recover.

\paragraph{Generate--validate--rewrite loop.}
Personalized-service task construction uses a generate--validate--rewrite pipeline.
For each validated state key, the generator first proposes a candidate service item.
The validator then checks the candidate against the five service-validity criteria below.
If a candidate fails, a rewrite prompt receives the failed criteria and revises only the mutable task fields, such as \texttt{scenario}, \texttt{reference\_answer}, \texttt{output\_template}, or \texttt{reference\_output}.
We allow up to two rewrite attempts; an item is retained only if it passes semantic validation.
After semantic acceptance, evaluation fields or scoring points are materialized programmatically from the accepted reference, rather than being authored as part of the generation prompt.

\paragraph{Service-validity criteria.}
We use the same five acceptance criteria for all personalized-service families:
\begin{itemize}[leftmargin=*, itemsep=1pt, topsep=2pt]
    \item \textbf{Answerability}: the scenario and task instruction define a single clear assistant task whose correct output is determined by the validated state and checkpoint-bounded evidence.
    \item \textbf{Service realism}: the task describes a natural assistant-mediated service moment, such as sending a useful reminder, setting search filters, or filling setup/form fields, rather than a backend placeholder or a contrived state-exposure query.
    \item \textbf{Full-field dependency}: the required output depends on the retained validated fields; omitting an important field should make the response materially incomplete or incorrect.
    \item \textbf{Low leakage}: the scenario and task text do not restate, paraphrase, or strongly imply the personalized facts that should be recovered from memory.
    \item \textbf{Output groundedness}: every personalized part of the reference answer or structured reference output is supported by the validated state, without adding unsupported user-specific facts.
\end{itemize}

At evaluation time, all systems consume these prebuilt packs directly.
Retrieval queries and visible task text are fixed by the task pack, which prevents a baseline from gaining advantage by rewriting the evaluation query or by accessing information beyond the checkpoint.

\section{Detailed Experiment Setup}
\label{app:exp_setup}

\paragraph{Baselines.}
We compare five representative long-term memory designs. Vanilla RAG stores the checkpoint-bounded app-log stream as raw chunks and retrieves directly from this store at answer time. A-Mem maintains linked memory notes that are revised as new user logs arrive. HippoRAG2 builds a graph-structured memory and retrieves through associative graph neighborhoods. MemoryOS organizes user information through a hierarchical memory structure with cross-level consolidation. SimpleMem recursively consolidates related entries into higher-level summaries and retrieves through a hybrid of semantic, keyword, and structured channels. We also include an Oracle diagnostic setting, where the model receives the gold supporting logs rather than relying on a constructed memory representation.

\paragraph{Input app-log organization.}
All baselines receive the same user trajectory as a chronological stream of app interaction records. Each record is a structured JSON object with a stable \texttt{app\_log\_id}, timestamp, \texttt{app\_name}, \texttt{api\_name}, and nested \texttt{request} and \texttt{response} payloads. The \texttt{app\_name} and \texttt{api\_name} identify the application surface and action type, while the request/response fields preserve the app-specific content. For example, a financial transfer, a fitness activity, an email action, or an assistant conversation is represented in the same outer schema but with different nested payload fields. Before being passed to a baseline, logs are sorted by timestamp and log id; at checkpoint $c$, the baseline can only access the prefix of this stream observed up to that checkpoint.

\paragraph{Shared temporal protocol.}
We use the same checkpoint protocol and the same prebuilt task packs for all systems. The paper-facing tasks are \textsc{State Completion}, which asks a system to reconstruct the user's current state, and \textsc{Personalized Service}, which asks a system to use that state in a realistic assistant-service scenario. For every item, the retrieval query is taken directly from the prebuilt task pack, and the answering prompt uses the corresponding pack-authored question. This keeps the task surface fixed across baselines and prevents systems from gaining advantage by rewriting evaluation queries or accessing post-checkpoint information.

\paragraph{Memory construction.}
For stateful memory systems, we build checkpoint memory snapshots from the raw app-log stream and then reuse these snapshots for answer generation. We keep each baseline's original memory-construction procedure and default internal setup, while standardizing the shared model components: \texttt{gpt-5-mini} is used as the base LLM for memory construction, and \texttt{text-embedding-3-large} is used wherever the baseline performs embedding-based retrieval, linking, or memory access. Thus, A-Mem uses its note creation and revision procedure, HippoRAG2 uses its graph construction pipeline, and MemoryOS uses its hierarchical consolidation mechanism. Vanilla RAG does not perform a separate LLM-based memory-writing stage; its memory store is the checkpoint-bounded raw log corpus.

\paragraph{Question answering and retrieval budget.}
For all baselines, we use \texttt{gpt-5-mini} as the answer-generation LLM and \texttt{text-embedding-3-large} for retrieving. Because different memory systems return different retrieval units, we do not force the same numerical \texttt{top\_k} across systems. Instead, we choose retrieval settings that keep the resulting retrieved-memory context sizes in a similar range (roughly 12--21k tokens), rather than forcing an identical \texttt{top\_k}. In the main paper runs, RAG and HippoRAG2 use \texttt{top\_k=20}, A-Mem uses \texttt{top\_k=5} with \texttt{linked\_neighbor\_top\_k=5}, MemoryOS uses \texttt{top\_k=10}, and SimpleMem uses \texttt{top\_k=10} across its semantic, keyword, and structured retrieval channels. Measured by retrieved-memory context size, the average \textsc{State Completion} context is 12.6k tokens for RAG, 12.8k for HippoRAG2, 15.6k for A-Mem, 12.7k for MemoryOS, and 20.8k for SimpleMem; for \textsc{Personalized Service}, the corresponding averages are 14.6k, 14.1k, 17.0k, 12.9k, and 20.7k tokens.

\paragraph{LLM-as-Judge Setup and Reliability.}
We use \texttt{gpt-5.4}~\citep{openai2025gpt5_1_chat} as the judge model for scoring of \textsc{State Completion} and \textsc{Personalized Service}. Since both tasks require semantic matching beyond exact string overlap, the judge compares each prediction against the corresponding gold state or service reference and produces structured judgments. To verify judge reliability, we manually audited a calibration slice from user 001 across four systems: A-Mem, HippoRAG2, Oracle, and RAG, with 189 judged items per system. We marked a judge result as unreasonable when the human auditor found its score or rationale inconsistent with the reference, prediction, or task requirements. Across 756 audited judgments, only 28 were marked unreasonable, yielding an overall agreement rate of 96.30\%. Per-system agreement was 96.30\% for A-Mem, 94.71\% for HippoRAG2, 96.30\% for Oracle, and 97.88\% for RAG, suggesting that the judge is sufficiently reliable for aggregate comparison.

\section{Prompt Template}
\label{app:prompt_template}

\subsection{LLM-as-a-Judge Prompt}
\label{app:judge}

\begin{tcolorbox}[
  promptbox,
  title=State Completion Task,
  breakable,
  fontupper=\footnotesize,
  fonttitle=\bfseries\footnotesize,
  left=1.2mm,
  right=1.2mm,
  top=1mm,
  bottom=1mm,
  boxsep=1mm
]
[Task Instruction]\\
You are evaluating whether a predicted personal profile entry matches a reference profile entry. Use a Core $+$ Detail field evaluation method: judge each requested field separately, using the full prediction as context. For every field, first identify the field's core meaning and supporting details, then score core correctness and detail quality separately.
\vspace{0.6em}
[Definitions]\\
-- \texttt{state\_key}: a label for the profile entry being evaluated; use it to understand the entry type and topic before deciding what is core.\\
-- \texttt{golden}: the reference profile entry to evaluate against.\\
-- \texttt{predicted}: the model's predicted profile entry.\\
-- \texttt{fields\_to\_judge}: the exact field paths to judge; return one judgment for each field path.\\
-- \texttt{field\_path}: the requested field identifier; copy it exactly into the output.\\
-- \texttt{core\_correct}: boolean judgment for whether \texttt{predicted} captures the requested field's core meaning.\\
-- \texttt{detail\_quality}: integer detail judgment: 2 complete/accurate, 1 partially complete or slightly imprecise, 0 mostly missing/wrong/contradictory.\\
-- \texttt{semantic equivalent}: the same practical meaning despite harmless wording or formatting differences, such as weekday names versus weekday indexes where 0=Monday and 6=Sunday.\\
\{category\_guidance\}
\vspace{0.6em}
[Evaluation Method]\\
For each requested field, use a two-part Core $+$ Detail judgment:\\
1. Identify the requested field's core meaning from \texttt{state\_key}, \texttt{field\_path}, and \texttt{golden}. Core is the central value, relationship, direction, routine component, or attribute that must be present for the prediction to be practically the same profile information.\\
2. Decide \texttt{core\_correct}. Set \texttt{true} when the predicted entry captures that core meaning, even if details are incomplete; set \texttt{false} when the prediction omits the field, contradicts it, gives a different core value, or is too vague to identify the same field meaning.\\
3. Decide \texttt{detail\_quality}. Detail means supporting precision beyond the core, such as exact wording, exact time, exact encoding, qualifiers, constraints, tier/version, branch/address, examples, or scope. Use 2 when key details are complete and accurate; 1 when important details are missing, vague, or slightly imprecise; 0 when details are mostly missing, wrong, contradictory, or unsupported. If \texttt{core\_correct=false}, \texttt{detail\_quality} should normally be 0 unless the prediction contains accurate but non-core details.
\vspace{0.6em}
[Constraints]\\
1. Judge only the requested \texttt{fields\_to\_judge}.\\
2. Return every requested \texttt{field\_path} exactly once.\\
3. Judge each requested \texttt{field\_path} independently, while using the full predicted profile entry as context.\\
4. Use \texttt{detail\_quality} only for detail completeness and precision; do not use it to override \texttt{core\_correct}.\\
5. Do not require exact wording, JSON field names, key order, or identical formatting when the meaning is semantically equivalent.\\
6. In every field judgment, write \texttt{analysis} before \texttt{core\_correct} and \texttt{detail\_quality}.
\vspace{0.6em}
[Example]\\
{[Example Input]}\\ 
\{example\_input\}\\
{[Example Output]}\\ 
\{example\_output\}
\vspace{0.6em}

[Input/Output Format]\\
Input: \texttt{state\_key}, \texttt{golden}, \texttt{predicted}, \texttt{fields\_to\_judge}.\\
Output JSON only:
\begin{lstlisting}[style=jsonprompt]
{
  "field_judgments": [
    {
      "field_path": "<field_path>",
      "analysis": "<brief analysis before the labels>",
      "core_correct": "<bool>",
      "detail_quality": "<0|1|2 integer>"
    }
  ]
}
\end{lstlisting}

[Input]\\
\{actual\_input\}
\end{tcolorbox}

\begin{tcolorbox}[
  promptbox,
  title=Personalized Service Task,
  breakable,
  fontupper=\footnotesize,
  fonttitle=\bfseries\footnotesize,
  left=1.2mm,
  right=1.2mm,
  top=1mm,
  bottom=1mm,
  boxsep=1mm
]
[Task Instruction]\\
You are evaluating whether a predicted assistant response matches the reference response for the same personalized service situation. Use a Core $+$ Detail field evaluation method for service fields: judge each requested assistant-message or structured-output field separately, using the full service context. For every field, first identify the field's core service value and supporting service details, then score core correctness and detail quality separately.
\vspace{0.6em}
[Definitions]\\
-- \texttt{scenario}: the concrete service moment or user situation.\\
-- \texttt{task\_instruction}: what the assistant is supposed to complete.\\
-- \texttt{reference}: the reference assistant message or structured service output.\\
-- \texttt{predicted}: the model's predicted assistant message or structured service output.\\
-- \texttt{fields\_to\_judge}: the exact field paths to judge; return one judgment for each field path.\\
-- \texttt{field\_path}: the requested field identifier; copy it exactly into the output.\\
-- \texttt{core\_correct}: boolean judgment for whether \texttt{predicted} captures the requested field's core service value.\\
-- \texttt{detail\_quality}: integer detail judgment: 2 complete/accurate, 1 partially complete or slightly imprecise, 0 mostly missing/wrong/contradictory.\\
-- \emph{semantic equivalent}: the same practical service meaning despite harmless wording, formatting, key naming, or ordering differences.\\
\{guidance\}
\vspace{0.6em}
[Evaluation Method]\\
For each requested field, use a two-part Core $+$ Detail judgment:\\
1. Identify the requested field's core service value from \texttt{scenario}, \texttt{task\_instruction}, \texttt{field\_path}, \texttt{fields\_to\_judge}, and \texttt{reference}. Core is the central service meaning that must be present for the predicted response to be practically useful for the same personalized service moment. Core belongs to the service output field being judged, not to a raw source record.\\
2. Decide \texttt{core\_correct}. Set \texttt{true} when the predicted response captures that core service value, even if details are incomplete; set \texttt{false} when the prediction omits the field, contradicts it, gives a different core value, targets a different service moment, or is too vague to identify the same field meaning.\\
3. Decide \texttt{detail\_quality}. Detail means service-useful precision beyond the core, such as exact time, date, place, cadence, exact encoding, qualifiers, exclusions, constraints, tier/version, branch/address, examples, or scope. Use 2 when key details are complete and accurate; 1 when important details are missing, vague, or slightly imprecise; 0 when details are mostly missing, wrong, contradictory, or unsupported. If \texttt{core\_correct=false}, \texttt{detail\_quality} should normally be 0 unless the prediction contains accurate but non-core details.
\vspace{0.6em}
[Constraints]\\
1. Judge only the requested \texttt{fields\_to\_judge}.\\
2. Return every requested \texttt{field\_path} exactly once.\\
3. Judge each requested \texttt{field\_path} independently, while using the full predicted response and service context.\\
4. Use \texttt{detail\_quality} only for detail completeness and precision; do not use it to override \texttt{core\_correct}.\\
5. Do not require exact wording, JSON field names, key order, or identical formatting when the meaning is semantically equivalent.\\
6. Do not penalize the prediction for not restating source records when the service output is correct, but do penalize missing details needed for the service response to be specific and actionable.\\
7. In every field judgment, write \texttt{analysis} before \texttt{core\_correct} and \texttt{detail\_quality}.
\vspace{0.6em}
[Example]\\
{[Example Input]}\\ 
\{example\_input\}\\
{[Example Output]}\\ 
\{example\_output\}
\vspace{0.6em}
[Input/Output Format]\\
Input: \texttt{scenario}, \texttt{task\_instruction}, \texttt{reference}, \texttt{predicted}, \texttt{fields\_to\_judge}.\\
Output JSON only:\\
\begin{lstlisting}[style=jsonprompt]
{
  "field_judgments": [
    {
      "field_path": "<field_path>",
      "analysis": "<brief analysis before the labels>",
      "core_correct": "<bool>",
      "detail_quality": "<0|1|2 integer>"
    }
  ]
}
\end{lstlisting}
[Input]\\
\{actual\_input\}
\end{tcolorbox}

\subsection{Answer Prompt}
\label{app:answer_prompt}
\noindent
\begin{tcolorbox}[
  promptbox,
  title=State Completion Task,
  breakable,
  fontupper=\footnotesize,
  fonttitle=\bfseries\footnotesize,
  left=1.2mm,
  right=1.2mm,
  top=1mm,
  bottom=1mm,
  boxsep=1mm
]
[Instructions]\\
-- Answer the question below using only the system memory about the user's trajectory provided in [Memory]\ldots[/Memory].\\
-- \textbf{Make each answer value as detailed and accurate as the memory supports.} Preserve specific names, times, dates, places, labels, and constraints instead of giving vague summaries.\\
-- Follow the template exactly, and fill every requested field with the most precise supported value.\\
\{schedule\_instruction\_if\_any\}-- \texttt{evidence} for each key must be a list of objects with \texttt{app\_log\_id} (use the exact app log id when it can be identified; otherwise ``'') and \texttt{evidence\_content} (one short supporting snippet or close paraphrase, kept local and concise).\\
-- If evidence is unknown, use \texttt{[]}.\\
-- Return JSON only. No extra keys.
\vspace{0.6em}
[Output format]\\
\begin{lstlisting}[style=jsonprompt]
{
  "user_state": {
    "<key>": "<detailed value or nested object>",
    ...
  },
  "evidence": {
    "<key>": [
      {
        "app_log_id": "<app_log_id>",
        "evidence_content": "<snippet>"
      }
    ],
    ...
  }
}
\end{lstlisting}
[Question] \\
\{task\_query\} \\
{[/Question]}
\vspace{0.6em}
\{memory\_section\}
\vspace{0.6em}
Concrete JSON skeleton to fill: \\
\{fill\_template\_block\}
\end{tcolorbox}

\begin{tcolorbox}[
  promptbox,
  title=Personalized Service Task,
  breakable,
  fontupper=\footnotesize,
  fonttitle=\bfseries\footnotesize,
  left=1.2mm,
  right=1.2mm,
  top=1mm,
  bottom=1mm,
  boxsep=1mm
]
[Instructions]\\
-- Use [Assistant Task] and the system-maintained memory of the user's trajectory in [Memory]\ldots[/Memory] to complete the assistant task.\\
-- Put the completed task result in \texttt{answer}.\\
-- Put the memory evidence supporting that result in \texttt{evidence}.\\
-- \texttt{evidence} must be a list of objects with \texttt{app\_log\_id} (use the exact app log id when it can be identified; otherwise ``'') and \texttt{evidence\_content} (one short supporting snippet or close paraphrase, kept local and concise).\\
-- If evidence is unknown, use \texttt{[]}.\\
-- Return JSON only; do not write any text outside the JSON object.
\vspace{0.6em}
[Output format]\\[0.3em]

\modeheader{Text mode}
\begin{lstlisting}[style=jsonprompt]
{
  "answer": "<specific and complete assistant message>",
  "evidence": [
    {
      "app_log_id": "<app_log_id>",
      "evidence_content": "<snippet>"
    }
  ]
}
\end{lstlisting}

\vspace{0.4em}
\modeheader{Structured mode}
\begin{lstlisting}[style=jsonprompt]
{
  "answer": {
    <output_template>
  },
  "evidence": [
    {
      "app_log_id": "<app_log_id>",
      "evidence_content": "<snippet>"
    }
  ]
}
\end{lstlisting}
[Assistant Task]\\
\{task\_body\} \\
{[/Assistant Task]}
\vspace{0.6em}
\{memory\_section\}
\end{tcolorbox}

\subsection{Trajectory Synthesis Prompt}
\label{app:data_prompt}

\begin{tcolorbox}[
  promptbox,
  title=Basic Profile Generation,
  breakable,
  fontupper=\footnotesize,
  fonttitle=\bfseries\footnotesize,
  left=1.2mm,
  right=1.2mm,
  top=1mm,
  bottom=1mm,
  boxsep=1mm
]
[Task Instruction]\\
You are creating a concrete user profile for behavioral simulation. Generate ONE specific, realistic profile based on the description below.
\vspace{0.6em}
[Input]\\
\{user\_description\}
\vspace{0.6em}
[Instructions]\\
1. Extract explicit facts from the description.\\
2. Infer missing details to create a complete, coherent profile.\\
3. Ensure all fields are internally consistent (e.g., student income matches student occupation).\\
4. Choose specific values: no vague or placeholder text.
\vspace{0.6em}
[Consistency Rules]\\
-- \textbf{Students}: income Low/Lower\_Mid, work\_hours 0--25, education Associate/Bachelor.\\
-- \textbf{Full-time workers}: work\_hours 35--50, income Mid or higher.\\
-- \textbf{Parents with young children}: caregiving\_load Moderate/High, household\_size 2+.\\
-- \textbf{Age \& education}: Bachelor typically 22+, Master 24+, Doctorate 27+.\\
-- \textbf{Income \& buffer}: High income $\to$ 6+ months buffer; Low income $\to$ 0--3 months.
\vspace{0.6em}
[Output Format (JSON only)]
\begin{lstlisting}[style=jsonprompt]
{
  "age": <integer 18-85>,
  "gender": "<Male|Female|Non_binary|Other>",
  "location_type": "<Urban_Metropolis|Urban|Suburban|Town|Rural>",
  "location_region": "<string>",
  "languages": ["<ISO codes>"],
  "occupation": "<specific job title>",
  "employment_status": "<Full_time|Part_time|Self_employed|Unemployed|Student|Retired|Caregiver>",
  "industry": "<specific industry>",
  "weekly_work_hours": <integer 0-70>,
  "education_level": "<Middle_School|High_School|Vocational|Associate|Bachelor|Master|Doctorate>",
  "income_band": "<Low|Lower_Mid|Mid|Upper_Mid|High>",
  "financial_buffer_months": "<0|<1|1-3|3-6|6-12|12+>",
  "housing_status": "<Renting|Home_Owner|Living_with_Family|Dormitory|Other>",
  "household_size": <integer 1-8>,
  "household_composition": "<string>",
  "caregiving_load": "<None|Light|Moderate|High>",
  "health_constraint_level": "<None|Mild|Moderate|Severe>",
  "digital_literacy": "<Low|Basic|Advanced|Native>",
  "planning_orientation": "<Present_biased|Balanced|Future_oriented>",
  "risk_tolerance": "<Risk_Averse|Neutral|Risk_Seeking>"
}
\end{lstlisting}
\end{tcolorbox}

\begin{tcolorbox}[
  promptbox,
  title=World Background Generation,
  breakable,
  fontupper=\footnotesize,
  fonttitle=\bfseries\footnotesize,
  left=1.2mm,
  right=1.2mm,
  top=1mm,
  bottom=1mm,
  boxsep=1mm
]
[Task Instruction]\\
You are an expert world-builder for behavioral simulation. Create a world background that captures the texture of daily life during the simulation period, narrated in third-person and focused on the world itself. The background should be grounded in the user's geographic and cultural context and relevant to the given life domain, but it should not describe the user's actions, feelings, or choices.
\vspace{0.6em}
[Input Information]\\
-- \texttt{user\_basic\_profile}: \{user\_basic\_profile\_json\}.\\
-- \texttt{life\_domain}: \{domain\_name\}: \{domain\_scope\_definition\}.\\
-- \texttt{predefined\_time\_windows}: w0 (2023-10-01 to 2023-12-31), w1 (2024-01-01 to 2024-03-31), w2 (2024-04-01 to 2024-07-01), w3 (2024-07-02 to 2024-09-30), w4 (2024-10-01 to 2024-12-31).
\vspace{0.6em}
[World Background Components]\\
1. \textbf{Physical Environment}: seasonal changes, weather patterns, local geography.\\
2. \textbf{Social Rhythms}: festivals, holidays, academic calendars, local customs.\\
3. \textbf{Key Events}: major world events (political, sports, cultural); prioritize events relevant to the user's region and culture.\\
4. \textbf{Prevailing Trends}: public sentiment, technological trends, economic conditions, domain-specific trends.
\vspace{0.6em}
[Critical Requirements]\\
1. \textbf{Narrator Perspective}: third-person describing the world; do not describe the user or use second-person.\\
2. \textbf{Domain Relevance}: focus on aspects of the world that directly impact the specified life domain.\\
3. \textbf{Temporal Coherence}: each window has a distinct character; maintain continuity between windows.\\
4. \textbf{Concrete Details}: specific dates for major events; reference real events, trends, and cultural phenomena.
\vspace{0.6em}
[Output Format (JSON)]
\begin{lstlisting}[style=jsonprompt]
{
  "world_backgrounds": [
    {
      "window_id": "w0",
      "time_range": ["2023-10-01", "2023-12-31"],
      "background": "<2-4 paragraph third-person description, relevant to the domain>"
    },
    ...
  ]
}
\end{lstlisting}
\end{tcolorbox}

\begin{tcolorbox}[
  promptbox,
  title=Life Context Baseline,
  breakable,
  fontupper=\footnotesize,
  fonttitle=\bfseries\footnotesize,
  left=1.2mm,
  right=1.2mm,
  top=1mm,
  bottom=1mm,
  boxsep=1mm
]
[Task Instruction]\\
You are a constraint extraction agent. Generate a minimal life context that defines external constraints on when/where/with-whom events can occur.
\vspace{0.6em}
[Framework]\\
-- \textbf{Capability}: physical reach (travel time) and time budgets (when user is available).\\
-- \textbf{Authority}: location access rules (opening hours, memberships).\\
-- \textbf{Coupling}: coordination requirements (lead time, participants).
\vspace{0.6em}
[Input]\\
-- \texttt{basic\_profile}: \{user\_basic\_profile\_json\}.\\
-- \texttt{baseline\_profile}: \{dynamic\_profiles\_initial\_state\_json\}.
\vspace{0.6em}
[Capability]\\
Generate four time-budget blocks (defaults: \texttt{sleep} 23:00--07:00 soft, \texttt{work\_core} 09:00--18:00 weekday hard, \texttt{free\_evening} 18:00--23:00 weekday, \texttt{free\_weekend} 09:00--23:00). Mobility edges only for home$\leftrightarrow$work and home$\leftrightarrow$locations in 3+ weekly habits; use generic location types.
\begin{lstlisting}[style=jsonprompt]
{"name": "sleep|work_core|free_evening|free_weekend",
 "days": "all|weekday|weekend",
 "start": "HH:MM", "end": "HH:MM",
 "rigidity": "hard|soft"}

{"mode_defaults": {"local": "drive|walk|transit"},
 "edges": [{"from": "...", "to": "...", "mode": "...",
            "travel_time": {"kind": "quantiles", "min": X, "p50": Y, "p90": Z, "max": W},
            "time_sensitivity": "rush_hour|weather|stable"}]}
\end{lstlisting}
\vspace{0.6em}
[Authority]\\
Generate 5--8 places. Mandatory: \texttt{home} (private), \texttt{work} (if employed). Use generic types only.
\begin{lstlisting}[style=jsonprompt]
{"id": "...", "type": "residence|workplace|fitness|retail|social|healthcare",
 "access": {"kind": "private|restricted|public",
            "open_hours": [["weekday", "09:00", "18:00"]],
            "membership_required": true}}
\end{lstlisting}
\vspace{0.6em}
[Coupling]\\
Include only habits with explicit timing and other people involved. Standard coordination patterns: co-resident partner (min\_h=0, penalty=high); small group 4--8 (min\_h=48, preferred\_h=168); professional (min\_h=24, preferred\_h=72).
\begin{lstlisting}[style=jsonprompt]
{"id": "...", "activity_tag": "date_night|board_game|gym_class",
 "rule": {"recurrence": "...", "day": "...", "time": "...", "duration_min": N},
 "participants": [...], "location_hint": "...", "preemptible": false}
\end{lstlisting}
\vspace{0.6em}
[Output Format]
\begin{lstlisting}[style=jsonprompt]
{"life_context": {
   "timezone": "America/Los_Angeles",
   "capability": {"time_budgets": [...], "mobility": {...}},
   "authority": {"places": [...]},
   "coupling": {"commitments": [...], "coordination_rules": [...]}}}
\end{lstlisting}
\vspace{0.6em}
[Rules]\\
1. Use generic types, not specific names.\\
2. Extract only what's in habits or blocks scheduling.\\
3. Minimize entries: 4 time\_budgets, 3--5 edges, 5--8 places.\\
4. If unsure, omit (better sparse than speculative).
\end{tcolorbox}

\begin{tcolorbox}[
  promptbox,
  title=Life Context Delta,
  breakable,
  fontupper=\footnotesize,
  fonttitle=\bfseries\footnotesize,
  left=1.2mm,
  right=1.2mm,
  top=1mm,
  bottom=1mm,
  boxsep=1mm
]
[Task Instruction]\\
You are a constraint delta agent. Analyze a time window to identify TEMPORARY changes to the user's life context constraints.
\vspace{0.6em}
[Input]\\
-- \texttt{basic\_profile}: \{user\_basic\_profile\_json\}.\\
-- \texttt{baseline}: \{life\_context\_baseline\_json\}.\\
-- \texttt{time\_window}: \{window\_description\}, \{window\_summary\}, \{window\_world\_background\}.
\vspace{0.6em}
[Task]\\
Generate constraint deltas (add/modify/suspend) that apply ONLY during this window. Focus on:\\
1. \textbf{Travel/Relocation}: does the user leave the primary region? (capability.mobility)\\
2. \textbf{Schedule Disruptions}: do work hours change? Holidays? (capability.time\_budgets)\\
3. \textbf{Access Changes}: new locations needed? Facilities closed? (authority.places)\\
4. \textbf{Coordination Shifts}: are recurring commitments suspended? New group activities? (coupling)
\vspace{0.6em}
[Delta Schemas]\\
Each delta has \texttt{change\_type}, \texttt{effective\_dates} (specific range or null for entire window), and \texttt{reason}.
\begin{lstlisting}[style=jsonprompt]
// Time budget
{"change_type": "suspend|modify|add",
 "target": "work_core|sleep|free_evening|free_weekend",
 "effective_dates": [...], "new_state": {...}, "reason": "..."}

// Mobility
{"change_type": "add_temporary_location|modify_edge",
 "details": {"location_id": "...", "from_baseline": "home",
             "effective_dates": [...]}, "reason": "..."}

// Place access
{"change_type": "add|suspend|modify", "place_id": "...",
 "effective_dates": [...], "modification": {...}, "reason": "..."}

// Commitment
{"change_type": "suspend|reschedule|add", "commitment_id": "...",
 "effective_dates": [...], "modification": {"suspended": true, "reason": "..."}}
\end{lstlisting}
\vspace{0.6em}
[Trigger Signals]\\
-- \textbf{In window\_description}: ``traveling to'', ``vacation in'', ``visiting'', ``holiday'', ``break'', ``time off'', ``office shutdown'', ``partner away'', ``hosting visitors''.\\
-- \textbf{In window\_summary}: habit changes (``drops outdoor\_running''), attribute changes (``adds travel gear''), relationship changes.\\
-- \textbf{In window\_world\_background}: major events (Olympics, holidays), natural events (heatwave, storm).
\vspace{0.6em}
[Default to NO DELTA when]\\
-- Window only describes internal changes (skills, preferences).\\
-- Events are purely informational (reading about Olympics is not attending).\\
-- Changes are already covered by existing baseline seasonal modifiers.
\vspace{0.6em}
[Output Format]
\begin{lstlisting}[style=jsonprompt]
{
  "window_id": "w1|w2|w3|w4",
  "window_dates": ["YYYY-MM-DD", "YYYY-MM-DD"],
  "has_constraint_changes": true,
  "deltas": {"capability": [...], "authority": [...], "coupling": [...]},
  "rationale": "..."
}
\end{lstlisting}
\vspace{0.6em}
[Rules]\\
1. Only generate deltas with explicit evidence in window text.\\
2. Specify \texttt{effective\_dates} when possible (rather than ``entire window'').\\
3. Prefer \texttt{suspend} over \texttt{delete}: constraints usually return after the window.\\
4. Don't infer lifestyle from events (e.g., ``Bitcoin Halving'' does not change schedule).\\
5. Don't duplicate baseline seasonal modifiers.
\end{tcolorbox}

\begin{tcolorbox}[
  promptbox,
  title=Events Chain Generation,
  breakable,
  fontupper=\footnotesize,
  fonttitle=\bfseries\footnotesize,
  left=1.2mm,
  right=1.2mm,
  top=1mm,
  bottom=1mm,
  boxsep=1mm
]
[Task Instruction]\\
You are an expert at generating realistic event chains that demonstrate user behaviors based on their dynamic profile state. Given a user's state for a specific time window, generate a sequence of realistic events that would naturally occur based on their attributes, habits, and preferences. Each event must specify which app and API it uses, along with the user intent.
\vspace{0.6em}
[Core Principle: Lossless State-to-Event Conversion]\\
1. \textbf{Semantic Completeness}: events must fully capture the meaning of each state item (both sides of a preference, etc.).\\
2. \textbf{Behavioral Fidelity}: events should simulate ACTUAL user behavior with realistic details (what exactly the user types/clicks/searches).\\
3. \textbf{Intent as Generation Context}: each event's \texttt{user\_intent} provides motivation AND content direction so downstream generation can produce realistic, specific app inputs.\\
4. \textbf{Process + Outcome}: for any state change, show both the causation events and the demonstration events.
\vspace{0.6em}
[Input]\\
-- \texttt{world\_background}: \{world\_background\}.\\
-- \texttt{user\_basic\_profile}: \{user\_basic\_profile\}.\\
-- \texttt{previous\_window\_summary}: \{user\_previous\_window\_summary\}.\\
-- \texttt{domain\_previous\_window\_summary}: \{user\_domain\_previous\_window\_summary\}.\\
-- \texttt{this\_window\_description}: \{user\_this\_window\_description\}.\\
-- \texttt{domain\_window\_state}: \{domain\_window\_state\}.\\
-- \texttt{app\_catalog}: \{app\_catalog\_json\}.
\vspace{0.6em}
[App/API Compliance]\\
Every \texttt{app\_name} must exist in the catalog; every \texttt{api\_name} must belong to its app per the catalog. If the user behavior requires functionality not available in the catalog, fall back to \texttt{LLM Assistant} with \texttt{ContinueConversation} or \texttt{CreateConversation}.
\vspace{0.6em}
[Lossless Conversion Mapping]\\
-- \textbf{Attribute value} $\to$ concrete usage demonstrating possession/capability.\\
-- \textbf{Attribute acquisition} $\to$ research, decision-making, purchase/setup.\\
-- \textbf{Habit schedule} $\to$ events at specified times on specified days.\\
-- \textbf{Habit evolution} $\to$ early executions differ from later ones.\\
-- \textbf{Preference direction} $\to$ choices that favor X over alternatives.\\
-- \textbf{Preference strength} $\to$ consistency and frequency of preference-aligned choices.\\
-- \textbf{Change reason} $\to$ events that make the trigger/catalyst observable.
\vspace{0.6em}
[Required Observable Fields]\\
Each state item carries \texttt{required\_observable\_fields} listing fields that MUST be evidenced. Every field listed there must be evidenced by at least one event in \texttt{evidenced\_fields}.
\vspace{0.6em}
[User Intent Composition]\\
Each \texttt{user\_intent} should include: (1) what the user wants to do, (2) why, (3) specific constraints or criteria. Generic intents that lose state richness are not acceptable.
\vspace{0.6em}
[Output Format (JSON)]
\begin{lstlisting}[style=jsonprompt]
{
  "window_id": "...",
  "events": [
    {
      "event_id": "e_<window>_<n>",
      "timestamp": "YYYY-MM-DD HH:MM:SS",
      "app_name": "<from catalog>",
      "api_name": "<from catalog>",
      "user_intent": "<detailed motivation + content direction>",
      "related_state_items": [{"name": "...", "evidenced_fields": ["..."]}],
      "chain_role": "initial_research | catalyst | execution | follow_up | ..."
    }
  ]
}
\end{lstlisting}
\end{tcolorbox}

\begin{tcolorbox}[
  promptbox,
  title=App Log Generation,
  breakable,
  fontupper=\footnotesize,
  fonttitle=\bfseries\footnotesize,
  left=1.2mm,
  right=1.2mm,
  top=1mm,
  bottom=1mm,
  boxsep=1mm
]
[Task Instruction]\\
Generate a realistic API call log (input and output) for a specific user interaction with an app. You are simulating what a real user would input and what the app would return based on the user's profile, context, and current app state.
\vspace{0.6em}
[Input Context]\\
-- \texttt{api\_schema}: input/output fields and types: \{api\_schema\}.\\
-- \texttt{user\_profile}: demographics, domain context, preferences: \{user\_profile\}.\\
-- \texttt{app\_state}: current app state before this call (history, saved items, conversations): \{app\_state\}.\\
-- \texttt{event\_payload}: time spec, user intent, related state items, chain context: \{event\_payload\}.\\
-- \texttt{previous\_chain\_logs}: prior logs in the same chain (for consistency in IDs, conversations, narrative): \{previous\_chain\_logs\}.
\vspace{0.6em}
[Generation Rules]\\
1. \textbf{Follow Schema Exactly}: include ALL required fields; respect types and nesting structure.\\
2. \textbf{Realistic User Input}: short, natural, casual phrasing OK; reflect user's expertise; serve the stated \texttt{user\_intent}.\\
3. \textbf{Realistic App Output}: results match what the app would return given input and state; rank by relevance using profile; realistic names/IDs; consistent with previous chain logs.\\
4. \textbf{Use State Discriminately}: \texttt{user\_intent} and \texttt{related\_state\_items} are ground truth for what to surface; do not fabricate unrelated history.\\
5. \textbf{Time}: pick a specific timestamp consistent with the chain; keep timestamps strictly increasing.\\
6. \textbf{Personalization Without Leakage}: reflect preferences/habits indirectly through ranking and selection; do not restate user attributes verbatim in user-facing fields.
\vspace{0.6em}
[Output Format]
\begin{lstlisting}[style=jsonprompt]
{
  "input":  { ... full input matching the API schema ... },
  "output": { ... full output matching the API schema ... },
  "timestamp": "YYYY-MM-DD HH:MM:SS"
}
\end{lstlisting}

[Critical Constraints]\\
1. JSON only; no markdown or commentary outside the JSON.\\
2. Schema fidelity: every required field present and correctly typed.\\
3. No hallucinated apps/APIs; this prompt is scoped to the single API in \texttt{api\_schema}.\\
4. Stay consistent with \texttt{previous\_chain\_logs} and \texttt{app\_state}.\\
5. Reflect the \texttt{user\_intent} end-to-end: input asks for it, output delivers it.
\end{tcolorbox}

\subsection{Evaluation Task Construction Prompt}
\label{app:qa_prompt}

\begin{tcolorbox}[
  promptbox,
  title=Personalized Service Task -- Habit Reminder,
  breakable,
  fontupper=\footnotesize,
  fonttitle=\bfseries\footnotesize,
  left=1.2mm,
  right=1.2mm,
  top=1mm,
  bottom=1mm,
  boxsep=1mm
]
[Task]\\
Generate exactly one habit-conditioned communication task for a user-facing assistant. Each item contains \texttt{scenario}, \texttt{task\_instruction} (copy fixed string exactly), and \texttt{reference\_answer}. The task should require the assistant to use the user's habit state, not only the scenario. The \texttt{reference\_answer} must be exactly one proactive natural-language assistant-to-user message for this moment.
\vspace{0.6em}
[Design Principle]\\
Keep \texttt{scenario} leakage-safe: it may set up the current moment and local situation, but it must not restate, paraphrase, or strongly imply the user-state facts that should instead be recovered from \texttt{state\_value}.
\vspace{0.6em}
[Definitions]\\
-- \texttt{terminal field}: one leaf field in \texttt{state\_value} whose value is scalar or array-valued.\\
-- \texttt{leaf path}: dot path to a terminal field (e.g., \texttt{schedule.days\_of\_week}, \texttt{timing.start\_time}).\\
-- \emph{world-background scenario}: short third-person description of what is true now in the world.\\
-- \emph{current-world anchor}: weekday, calendar date, and current clock time in \texttt{scenario}.
\vspace{0.6em}
[Hard Constraints]\\
1. Generate exactly one item.\\
2. Output JSON only with one top-level key \texttt{item} containing \texttt{scenario}, \texttt{task\_instruction}, \texttt{reference\_answer}.\\
3. Copy \texttt{task\_instruction} exactly as the fixed string \{fixed\_task\_instruction\}.\\
4. \texttt{scenario} must be short, concrete, third-person world background; no first/second-person (``I'', ``we'', ``you'', ``your'', ``you've'').\\
5. \texttt{scenario} must anchor the current moment clearly (weekly/weekday routines: weekday + clock time; monthly/date-like routines: calendar anchor + clock time). For schedule day fields, clock time alone is insufficient.\\
6. \texttt{scenario} may include only: the current moment; whether something has or has not happened yet; whether something has or has not been prepared; at most one additional plausible situational fact.\\
7. \texttt{scenario} must not restate or paraphrase the routine action, frequency, stored start time, end time, location, or any other personalized habit fact already in \texttt{state\_value}.\\
8. \texttt{reference\_answer} is exactly one natural assistant-to-user message; complete enough that a fully correct answer would use every terminal field in \texttt{state\_value}.\\
9. Before finalizing, silently confirm: answerability, service\_realism, full\_field\_dependency, low\_leakage, output\_groundedness.
\vspace{0.6em}
[Good Example]
\begin{lstlisting}[style=jsonprompt]
state_key: "habits_state:client_technical_briefing"
state_value: {"schedule": {"frequency_type": "weekly", "days_of_week": [0]},
              "timing": {"start_time": "10:00"},
              "location": "regional corporate headquarters"}

{"item": {
  "scenario": "It is Monday at 09:20. Nothing has been started yet this morning.",
  "task_instruction": <fixed>,
  "reference_answer": "Your weekly client technical briefing is at 10:00 today at the regional corporate headquarters. Since Monday is the scheduled day, it is almost time to get ready."
}}
\end{lstlisting}
\vspace{0.6em}
[Input]\\
\texttt{state\_key}: \{state\_key\}; \texttt{state\_value}: \{state\_value\}.
\vspace{0.6em}
[Output Format (JSON only)]
\begin{lstlisting}[style=jsonprompt]
{"item": {
   "scenario": "...",
   "task_instruction": <fixed>,
   "reference_answer": "..."
}}
\end{lstlisting}
\end{tcolorbox}

\begin{tcolorbox}[
  promptbox,
  title=Personalized Service Task -- Auto Filtering,
  breakable,
  fontupper=\footnotesize,
  fonttitle=\bfseries\footnotesize,
  left=1.2mm,
  right=1.2mm,
  top=1mm,
  bottom=1mm,
  boxsep=1mm
]
[Task]\\
Generate exactly one preference-conditioned search-filter task for an assistant helping the user browse, search, compare, or plan options. Each item contains \texttt{scenario}, \texttt{task\_instruction} (copy fixed string exactly), \texttt{output\_template}, \texttt{reference\_output}, and \texttt{reference\_anchors}. The \texttt{reference\_output} must fill a structured search/filter object for the assistant to apply before showing matching options; it must not be a final recommendation, ranked list, or free-form explanation.
\vspace{0.6em}
[Key Design Goal]\\
This is a search-filter task, not a copy-the-statement task. The generated item should require the answering assistant to translate the user's preference statement into semantically meaningful filters.
\vspace{0.6em}
[Definitions]\\
-- \emph{preference statement}: the value in \texttt{state\_value.statement}.\\
-- \emph{world-background scenario}: short third-person description of what is happening now in the product or assistant context.\\
-- \emph{fill leaf}: one \texttt{<fill>} slot in \texttt{output\_template} and the corresponding filled value in \texttt{reference\_output}.\\
-- \emph{reference\_anchors}: audit notes tying each fill leaf to the state/reference basis used; for traceability, not scoring.\\
-- \emph{core fill}: a fill leaf whose value captures the field-local core preference.\\
-- \emph{detail fill}: optional second fill leaf adding grounded precision/qualification/exclusion; must be useful, not filler.
\vspace{0.6em}
[Hard Constraints]\\
1. Generate exactly one item.\\
2. Output JSON only with one top-level key \texttt{item} containing \texttt{scenario}, \texttt{task\_instruction}, \texttt{output\_template}, \texttt{reference\_output}, \texttt{reference\_anchors}.\\
3. Copy \texttt{task\_instruction} exactly as the fixed string \{fixed\_task\_instruction\}.\\
4. \texttt{scenario} short, natural, world-background; no first/second-person; plausible user product moment, not a backend log line. Avoid robotic phrasing (``a filtering step is about to run'', etc.).\\
5. \texttt{scenario} may include only: immediate user goal or option space; the assistant setting search/filter fields; at most one additional situational fact.\\
6. \texttt{scenario} must not restate or paraphrase the user's actual preference content.\\
7. \texttt{output\_template} and \texttt{reference\_output} have the same nested shape; every leaf in \texttt{output\_template} is the string \texttt{<fill>}.\\
8. \texttt{output\_template} contains one or two fill leaves total. At least one is a core fill; a second may be a detail fill when grounded and service-useful.\\
9. \texttt{reference\_anchors}: one object per fill leaf with \texttt{target\_path}, \texttt{role} (\texttt{core}$|$\texttt{detail}), \texttt{state\_reference}, \texttt{anchor\_note}.\\
10. Do not use a fixed universal key like \texttt{preference\_statement}; synthesize request-facing keys that decompose the preference into meaningful filtering dimensions (preferred types, desired attributes, required features, avoided options, priorities).\\
11. Every filled value in \texttt{reference\_output} must be supported by \texttt{state\_value}.\\
12. Before finalizing, silently confirm: answerability, service\_realism, full\_field\_dependency, low\_leakage, output\_groundedness.
\vspace{0.6em}
[Good Example]
\begin{lstlisting}[style=jsonprompt]
state_key: "preferences_state:learning_modality"
state_value: {"statement": "Prefers in-depth, self-paced technical white papers and webinars over large live conferences"}

{"item": {
  "scenario": "The user is browsing professional-development resources in a learning portal. The assistant is setting search filters before showing matching options.",
  "task_instruction": <fixed>,
  "output_template": {"content_search_filters": {"resource_formats": "<fill>", "avoid_setting": "<fill>"}},
  "reference_output": {"content_search_filters": {"resource_formats": "in-depth, self-paced technical white papers or webinars", "avoid_setting": "large live conferences"}},
  "reference_anchors": [
    {"target_path": "content_search_filters.resource_formats", "role": "core",
     "state_reference": "statement: in-depth, self-paced technical white papers and webinars over large live conferences",
     "anchor_note": "captures the field-local core learning-resource preference"},
    {"target_path": "content_search_filters.avoid_setting", "role": "detail",
     "state_reference": "statement: over large live conferences",
     "anchor_note": "records the grounded exclusion needed for filtering"}
  ]
}}
\end{lstlisting}
\vspace{0.6em}
[Input]\\
\texttt{state\_key}: \{state\_key\}; \texttt{state\_value}: \{state\_value\}.
\vspace{0.6em}
[Output Format (JSON only)]
\begin{lstlisting}[style=jsonprompt]
{"item": {
   "scenario": "...",
   "task_instruction": <fixed>,
   "output_template": {"<synth_request_key>": {"<synth_filter_key>": "<fill>"}},
   "reference_output": {"<same shape>": "..."},
   "reference_anchors": [{"target_path": "...", "role": "core|detail",
                          "state_reference": "...", "anchor_note": "..."}]
}}
\end{lstlisting}
\end{tcolorbox}

\begin{tcolorbox}[
  promptbox,
  title=Personalized Service Task -- Auto Configuration,
  breakable,
  fontupper=\footnotesize,
  fonttitle=\bfseries\footnotesize,
  left=1.2mm,
  right=1.2mm,
  top=1mm,
  bottom=1mm,
  boxsep=1mm
]
[Task]\\
Generate exactly one attribute-conditioned action-configuration task for an assistant helping the user set up, connect, complete, or submit something. Each item contains \texttt{scenario}, \texttt{task\_instruction} (copy fixed string exactly), \texttt{output\_template}, \texttt{reference\_output}, and \texttt{reference\_anchors}. The \texttt{reference\_output} must fill a structured action-configuration object for a user-facing tool, setup flow, form, or executable service; it must not be a user-facing message, retrieval request, or free-form explanation.
\vspace{0.6em}
[Key Design Goal]\\
This is a setup/form-configuration task, not a copy-the-attribute task. The generated item should require the answering assistant to translate the user's known attributes into the specific fields needed to complete a user-facing action.
\vspace{0.6em}
[Definitions]\\
-- \emph{attribute value}: the information contained in \texttt{state\_value}.\\
-- \emph{grounded decomposition}: a raw attribute string may be split into multiple configuration leaves only when each resulting leaf is directly supported by \texttt{state\_value} and serves a distinct execution role.\\
-- \emph{deterministic auto-fill}: every filled value should be determined by \texttt{state\_value} and the neutral setup/form context, not by an extra user choice.\\
-- \emph{fill leaf}, \emph{reference\_anchors}, \emph{core fill}, \emph{detail fill}: as in the search-filter task above.
\vspace{0.6em}
[Hard Constraints]\\
1. Generate exactly one item.\\
2. Output JSON only with one top-level key \texttt{item} containing \texttt{scenario}, \texttt{task\_instruction}, \texttt{output\_template}, \texttt{reference\_output}, \texttt{reference\_anchors}.\\
3. Copy \texttt{task\_instruction} exactly as the fixed string \{fixed\_task\_instruction\}.\\
4. \texttt{scenario} short, natural, world-background; no first/second-person; plausible user product moment. Prefer natural situations: completing checkout, finishing a setup flow, preparing a profile/form before submission, connecting a device/account, the assistant auto-filling setup/form fields.\\
5. \texttt{scenario} may include only: immediate user goal/action; the assistant filling setup/form/configuration fields; at most one additional situational fact. It must not restate or paraphrase the user's actual attribute values.\\
6. \texttt{output\_template} and \texttt{reference\_output} have the same nested shape; every leaf in \texttt{output\_template} is the string \texttt{<fill>}; one or two fill leaves total.\\
7. At least one fill leaf is a core fill; a second may be a detail fill when grounded and service-useful.\\
8. \texttt{reference\_anchors}: one object per fill leaf with \texttt{target\_path}, \texttt{role}, \texttt{state\_reference}, \texttt{anchor\_note}.\\
9. Prefer configuration-facing schemas that decompose compound attribute strings into execution-relevant fields when supported. Do not invent facts not directly in \texttt{state\_value}.\\
10. Avoid scenarios that require an extra user choice not in \texttt{state\_value} (subset, quantity, recipient, priority, destination, commitment).\\
11. Every filled value in \texttt{reference\_output} must be supported by \texttt{state\_value}; for list-valued state preserve source order when configuration represents per-item entries.\\
12. Before finalizing, silently confirm: answerability, service\_realism, full\_field\_dependency, low\_leakage, output\_groundedness.
\vspace{0.6em}
[Good Example]
\begin{lstlisting}[style=jsonprompt]
state_key: "user_attributes_state:primary_job_role"
state_value: "Senior Coatings Consultant at PPG Industries (specializing in heavy-duty infrastructure and marine protection)"

{"item": {
  "scenario": "The user is completing registration for a technical industry symposium. The assistant is filling the professional credential fields before submission.",
  "task_instruction": <fixed>,
  "output_template": {"symposium_registration": {"professional_profile": {"role_title": "<fill>", "specialization": "<fill>"}}},
  "reference_output": {"symposium_registration": {"professional_profile": {"role_title": "Senior Coatings Consultant at PPG Industries", "specialization": "heavy-duty infrastructure and marine protection"}}},
  "reference_anchors": [
    {"target_path": "symposium_registration.professional_profile.role_title", "role": "core",
     "state_reference": "Senior Coatings Consultant at PPG Industries (specializing in heavy-duty infrastructure and marine protection)",
     "anchor_note": "captures the field-local core professional identity"},
    {"target_path": "symposium_registration.professional_profile.specialization", "role": "detail",
     "state_reference": "specializing in heavy-duty infrastructure and marine protection",
     "anchor_note": "adds the grounded specialization needed by the credential fields"}
  ]
}}
\end{lstlisting}
[Input]\\
\texttt{state\_key}: \{state\_key\}; \texttt{state\_value}: \{state\_value\}.
\vspace{0.6em}
[Output Format (JSON only)]
\begin{lstlisting}[style=jsonprompt]
{"item": {
   "scenario": "...",
   "task_instruction": <fixed>,
   "output_template": {"<synth_config_key>": {"<synth_exec_field>": "<fill>"}},
   "reference_output": {"<same shape>": "..."},
   "reference_anchors": [{"target_path": "...", "role": "core|detail",
                          "state_reference": "...", "anchor_note": "..."}]
}}
\end{lstlisting}
\end{tcolorbox}

\section{Error Taxonomy: Judge Prompt}
\label{app:error-taxonomy}

The LLM judge described in Section~\ref{sec:error-analysis}
classifies each failure with the prompt below. The prompt is fixed
across systems; only the per-case block (QUESTION, GOLD\_ANSWER,
CITED\_EVIDENCE) changes between cases. Three worked examples drawn
from the same state key make the decision boundaries concrete.

\begin{tcolorbox}[
  promptbox,
  title=State Completion Task,
  breakable,
  fontupper=\footnotesize,
  fonttitle=\bfseries\footnotesize,
  left=1.2mm,
  right=1.2mm,
  top=1mm,
  bottom=1mm,
  boxsep=1mm
]
\begin{lstlisting}[basicstyle=\small\ttfamily,breaklines=true]
You are diagnosing a memory system's failure. You will be given:

  - QUESTION: the actual natural-language query the memory system received when answering. It identifies which aspect of the user's state the system was asked to recall, and may include a target template indicating the fields the answer is expected to fill.
  - GOLD_ANSWER: the target value, written as a structured object that names the user's actual state. For habits the gold has multiple fields (schedule, timing, location). For preferences the gold is a statement of preference direction with named options. For attributes the gold names a specific entity with its modifiers.
  - CITED_EVIDENCE: the list of memory entries the memory system surfaced for this question. Each entry has an id and a content string. This is everything the answer model had access to.

You are NOT shown the answer model's final prediction. Each case fed to you is already known to be a failure (the prediction scored less than full credit). Your job is to classify WHY the memory system's deliverable made the failure possible, by methodically comparing the cited evidence against the gold answer, and walking the decision tree below. Stop at the first label that triggers. Each case gets exactly one label.

IDENTITY vs DETAIL

This taxonomy classifies failures at the case level (one label per case). To make multi-field gold answers (especially habits) classifiable with a single label, we factor the gold into two layers:

- IDENTITY: the central identity of the state item the answer is built around. It is the primary "what is this state item about" of the gold answer.
    * Habits - the activity identity plus the recurrence pattern at a coarse level (e.g., "a recurring family dinner routine", "a weekly deep-work block"). The specific scheduled day, exact start/end time, and specific named location are NOT part of identity; they belong to the detail layer.
    * Preferences - the preference direction plus the main object or constraint (e.g., "prefers self-paced study over live events"). Specific named options and contrast examples belong to detail.
    * Attributes - the specific named entity (e.g., "Honda Odyssey"). Entity-level modifiers (year, model, recent additions) belong to detail.

- DETAIL: the supporting precision around the identity. For habits this is the specific scheduled day(s), the specific timing window, and the specific named location. For preferences this is the named options and contrast examples. For attributes this is the entity's year, version, recent modifications, and so on.

Important rule: if the IDENTITY is establishable from the cited evidence but at least one required field-level DETAIL is absent or imprecise, the case is Detail_Miss, not Identity_Miss.

(Note: this case-level "Identity / Detail" framework is intentionally coarser than the per-field "core / detail" scoring rubric, so that each whole case can receive one diagnostic label. The rubric's per-field core may treat 'days of week' as a field-level core; here we treat the specific weekday as a detail, because the case-level identity is the activity itself.)

LABELS

The four memory-failure labels (these are the categories we report):

  - Irrelevant_Evidence
  - Identity_Miss
  - Detail_Miss
  - Conflated_Evidence

Plus one residual label for the rare case where the evidence is complete and unambiguous yet the prediction is still wrong:

  - All_Clear

DECISION TREE

Compare the cited evidence against the gold answer at each step. Stop at the first triggered label.

  1) Relevance check.
     Does the cited evidence contain any content about what the question is asking? Use the gold answer to determine what 'on-topic' means.
       - NO  -> label = Irrelevant_Evidence. Stop.
       - YES -> continue.

  2) Identity check.
     Does at least one evidence entry establish the gold answer's IDENTITY (the activity identity for habits, the preference direction for preferences, the named entity for attributes)?
       - The identity is not establishable from any evidence entry -> label = Identity_Miss. Stop.
       - The identity IS establishable in some entry, but other evidence entries assert a different identity filling the same role (e.g., a different activity, the opposite preference, a different named entity), without indication of which is current -> label = Conflated_Evidence. Stop.
       - The identity is establishable and uncontested -> continue.

  3) Detail check.
     Does the cited evidence include the gold answer's field-level DETAILS (specific day, specific time, specific location, named options, year/version, etc.)?
       - Any required detail is absent (evidence carries the identity but does not mention the gold's specific details) -> label = Detail_Miss. Stop.
       - The details are present BUT contradicted by competing alternative details for the same identity in other evidence entries -> label = Conflated_Evidence. Stop.
       - The details are present and uncontested -> continue.

  4) All checks passed: relevance, identity, and details are all clearly present and unambiguous in the cited evidence. Since we know the prediction still failed, the failure is attributable to the answer model rather than the memory system. -> label = All_Clear.

OUTPUT FORMAT (JSON)

  {
    "label": "<one of: Irrelevant_Evidence, Identity_Miss, Detail_Miss, Conflated_Evidence, All_Clear>",
    "step_failed": <1 | 2 | 3 | 4>,
    "justification": "<one or two sentences citing evidence id(s) that drove your decision; for multi-field gold (especially habits), briefly note which fields' details are present and which are absent in the evidence>"
  }

NOTES

  - Paraphrased identity references (e.g., "the user's minivan" when gold identity is "Honda Odyssey") count as establishing the identity ONLY if the reference is unambiguous given the surrounding evidence; otherwise treat as Identity_Miss.
  - "Competing identity" means a different specific identity filling the SAME role, not unrelated entities. "Honda Odyssey + Toyota Sienna" both as the user's family vehicle is competing. "Honda Odyssey + Marconi Union (music track)" is not.

EXAMPLES

Example 1 (Irrelevant_Evidence)
  QUESTION: "Infer the user's current state for primary family vehicle (user_attributes_state:primary_family_vehicle) using this template: {\"model\": \"<fill>\", \"detail\": \"<fill>\"}."
  GOLD_ANSWER: "Honda Odyssey 2022 (Recently detailed and fitted with new all-season tires for safe holiday travel)"
  CITED_EVIDENCE:
    - id=log_034, content="User spent the afternoon woodworking with walnut lumber in the basement workshop."
    - id=log_037, content="User asked technical cooking question about Wagner roasting technique."
  Answer:
  {
    "label": "Irrelevant_Evidence",
    "step_failed": 1,
    "justification": "Comparing the evidence against the gold, log_034 and log_037 are about hobbies (workshop, cooking) and contain no content about the user's vehicle."
  }

Example 2 (Identity_Miss)
  QUESTION: "Infer the user's current state for primary family vehicle (user_attributes_state:primary_family_vehicle) using this template: {\"model\": \"<fill>\", \"detail\": \"<fill>\"}."
  GOLD_ANSWER: "Honda Odyssey 2022 (Recently detailed and fitted with new all-season tires for safe holiday travel)"
  CITED_EVIDENCE:
    - id=log_021, content="User mentioned needing to take the kids to soccer practice in the family car this weekend."
    - id=log_088, content="User paid for routine car maintenance at the local dealer for the family vehicle."
  Answer:
  {
    "label": "Identity_Miss",
    "step_failed": 2,
    "justification": "Comparing the evidence against the gold identity 'Honda Odyssey': evidence is on topic (family vehicle) but never establishes the entity; only generic references like 'family car' and 'family vehicle' appear."
  }

Example 3 (Detail_Miss, single-entity attribute)
  QUESTION: "Infer the user's current state for primary family vehicle (user_attributes_state:primary_family_vehicle) using this template: {\"model\": \"<fill>\", \"detail\": \"<fill>\"}."
  GOLD_ANSWER: "Honda Odyssey 2022 (Recently detailed and fitted with new all-season tires for safe holiday travel)"
  CITED_EVIDENCE:
    - id=log_010, content="User's primary family vehicle is the Honda Odyssey 2022, used for carpooling and errands."
    - id=log_055, content="User scheduled a wash at the local detailer for the Odyssey."
  Answer:
  {
    "label": "Detail_Miss",
    "step_failed": 3,
    "justification": "Comparing the evidence against the gold details: identity (Honda Odyssey 2022) is established in log_010; the year-detail is present, but the gold's specific 'all-season tires for safe holiday travel' detail is not mentioned in any evidence entry."
  }

Example 4 (Detail_Miss, multi-field habit)
  QUESTION: "Infer the user's current state for sunday family dinner (habits_state:sunday_family_dinner) using this template: {\"schedule\": {\"days_of_week\": [...], \"frequency_type\": \"<fill>\"}, \"timing\": {\"start_time\": \"<fill>\", \"end_time\": \"<fill>\"}, \"location\": \"<fill>\"}."
  GOLD_ANSWER: {"schedule": {"frequency_type": "weekly", "days_of_week": [6]}, "timing": {"start_time": "14:30", "end_time": "16:30"}, "location": "family home dining room"}
  CITED_EVIDENCE:
    - id=log_00347, content="Sitting Time: 14:30 - 16:30. Adhering to the family dinner protocol."
    - id=log_00322, content="Main: Shish Tawook inspired Roasted Chicken. Prepared in the home kitchen."
    - id=log_00282, content="Sitting Time: 14:30 - 16:30. Fixed window to ensure the kids can finish school prep."
    - id=log_00466, content="Sitting Time: 14:30 - 16:30. 100% family attendance recorded."
  Answer:
  {
    "label": "Detail_Miss",
    "step_failed": 3,
    "justification": "Comparing the evidence against the gold: the identity (a recurring family dinner routine with the kids) is establishable from log_00347/00282/00466. The timing detail (14:30-16:30) is also present and consistent across entries. However, the gold's specific scheduled day (Sunday) is not stated in any evidence entry, and the location detail 'dining room' is only generically referenced as the home kitchen. Fields present: timing, family attendance. Fields absent: scheduled day, exact location."
  }
\end{lstlisting}
\end{tcolorbox}

\begin{tcolorbox}[
  promptbox,
  title=Personalized Service Task,
  breakable,
  fontupper=\footnotesize,
  fonttitle=\bfseries\footnotesize,
  left=1.2mm,
  right=1.2mm,
  top=1mm,
  bottom=1mm,
  boxsep=1mm
]
\begin{lstlisting}[basicstyle=\small\ttfamily,breaklines=true]
You are diagnosing a memory system's failure. You will be given:

  - QUESTION: a personalized service scenario that includes a real-world moment in the user's day plus a task instruction telling the model what to produce (for example, drafting a reminder, completing a form, or recommending an action). The question also names the state context the system is supposed to ground in.
  - GOLD_ANSWER: the target response, written so that it reflects the user's actual state. It carries an identity for the state being applied and supporting details that make the response user-specific.
  - CITED_EVIDENCE: the list of memory entries the memory system surfaced for this question. Each entry has an id and a content string. This is everything the answer model had access to.

You are NOT shown the answer model's final response. Each case fed to you is already known to be a failure (the response scored less than full credit). Your job is to classify WHY the memory system's deliverable made the failure possible, by methodically comparing the cited evidence against the gold answer, and walking the decision tree below. Stop at the first label that triggers. Each case gets exactly one label.

IDENTITY vs DETAIL

This taxonomy classifies failures at the case level (one label per case). To make multi-field gold answers (especially habits) classifiable with a single label, we factor the gold into two layers:

- IDENTITY: the central identity of the state item the response must ground in. It is the primary "what is this state about" of the gold answer.
    * Habits - the activity identity plus the recurrence pattern at a coarse level (e.g., "a recurring family dinner routine", "a weekly deep-work block"). The specific scheduled day, exact start/end time, and specific named location are NOT part of identity; they belong to the detail layer.
    * Preferences - the preference direction plus the main object or constraint (e.g., "prefers self-paced study over live events"). Specific named options and contrast examples belong to detail.
    * Attributes - the specific named entity (e.g., "Honda Odyssey"). Entity-level modifiers (year, model, recent additions) belong to detail.

- DETAIL: the supporting precision around the identity. For habits this is the specific scheduled day(s), the specific timing window, and the specific named location. For preferences this is the named options and contrast examples. For attributes this is the entity's year, version, recent modifications, and so on.

Important rule: if the IDENTITY is establishable from the cited evidence but at least one required field-level DETAIL is absent or imprecise, the case is Detail_Miss, not Identity_Miss.

(Note: this case-level "Identity / Detail" framework is intentionally coarser than the per-field "core / detail" scoring rubric, so that each whole case can receive one diagnostic label.)

LABELS

The four memory-failure labels (these are the categories we report):

  - Irrelevant_Evidence
  - Identity_Miss
  - Detail_Miss
  - Conflated_Evidence

Plus one residual label for the rare case where the evidence is complete and unambiguous yet the response is still wrong:

  - All_Clear

DECISION TREE

Compare the cited evidence against the gold answer at each step. Stop at the first triggered label.

  1) Relevance check.
     Does the cited evidence contain any content about what the scenario is asking the system to ground in? Use the gold answer to determine what 'on-topic' means.
       - NO  -> label = Irrelevant_Evidence. Stop.
       - YES -> continue.

  2) Identity check.
     Does at least one evidence entry establish the gold answer's IDENTITY (the activity identity for habits, the preference direction for preferences, the named entity for attributes)?
       - The identity is not establishable from any evidence entry -> label = Identity_Miss. Stop.
       - The identity IS establishable, but other evidence entries assert a different identity filling the same role (e.g., a different routine, the opposite preference, a different named entity), without indication of which is current -> label = Conflated_Evidence. Stop.
       - The identity is establishable and uncontested -> continue.

  3) Detail check.
     Does the cited evidence include the gold answer's field-level DETAILS (specific day, specific time, specific location, named options, year/version, etc.)?
       - Any required detail is absent -> label = Detail_Miss. Stop.
       - Details are present BUT contradicted by competing alternative details for the same identity in other evidence entries -> label = Conflated_Evidence. Stop.
       - The details are present and uncontested -> continue.

  4) All checks passed: relevance, identity, and details are all clearly present and unambiguous in the cited evidence. Since we know the response still failed, the failure is attributable to the answer model rather than the memory system. -> label = All_Clear.

OUTPUT FORMAT (JSON)

  {
    "label": "<one of: Irrelevant_Evidence, Identity_Miss, Detail_Miss, Conflated_Evidence, All_Clear>",
    "step_failed": <1 | 2 | 3 | 4>,
    "justification": "<one or two sentences citing evidence id(s) that drove your decision; for multi-field gold (especially habits), briefly note which fields' details are present and which are absent in the evidence>"
  }

NOTES

  - Paraphrased identity references (e.g., "the user's weekend ritual" when gold identity is "Sunday budget review") count as establishing the identity ONLY if the reference is unambiguous given the surrounding evidence; otherwise treat as Identity_Miss.
  - "Competing identity" means a different specific identity filling the SAME role, not unrelated entities. Two different budget-review routines for the same scenario is competing; a budget-review routine plus an unrelated cooking-class log is not.

EXAMPLES (all three examples use the same scenario for direct comparison)

Example 1 (Irrelevant_Evidence)
  QUESTION:
    Scenario: It is Sunday at 09:15. The morning coffee has just been poured.
    Task: Draft a specific reminder message for the user about the upcoming routine in this scenario.
    State context being applied: habits_state:budget_review
  GOLD_ANSWER: "Today's Sunday 09:00-10:30 budget review at the desk: open Mint, reconcile Chase transactions, update savings goals."
  CITED_EVIDENCE:
    - id=log_034, content="User spent the afternoon woodworking with walnut lumber in the basement workshop."
    - id=log_037, content="User asked technical cooking question about Wagner roasting technique."
  Answer:
  {
    "label": "Irrelevant_Evidence",
    "step_failed": 1,
    "justification": "Comparing the evidence against the gold, log_034 and log_037 are about hobbies (workshop, cooking) and contain no content about the user's budget review routine."
  }

Example 2 (Identity_Miss)
  QUESTION:
    Scenario: It is Sunday at 09:15. The morning coffee has just been poured.
    Task: Draft a specific reminder message for the user about the upcoming routine in this scenario.
    State context being applied: habits_state:budget_review
  GOLD_ANSWER: "Today's Sunday 09:00-10:30 budget review at the desk: open Mint, reconcile Chase transactions, update savings goals."
  CITED_EVIDENCE:
    - id=log_142, content="User mentioned needing to do some financial planning this weekend."
    - id=log_198, content="User browsed personal-finance tips on the way home from work."
  Answer:
  {
    "label": "Identity_Miss",
    "step_failed": 2,
    "justification": "Comparing the evidence against the gold identity 'a recurring budget-review routine': evidence is on topic (financial planning) but never establishes the recurring routine; both entries refer only to general financial activity, not the user's specific budget-review habit."
  }

Example 3 (Detail_Miss, multi-field habit)
  QUESTION:
    Scenario: It is Sunday at 09:15. The morning coffee has just been poured.
    Task: Draft a specific reminder message for the user about the upcoming routine in this scenario.
    State context being applied: habits_state:budget_review
  GOLD_ANSWER: "Today's Sunday 09:00-10:30 budget review at the desk: open Mint, reconcile Chase transactions, update savings goals."
  CITED_EVIDENCE:
    - id=log_211, content="The user has a recurring weekend budget review they call their 'Sunday ritual.'"
    - id=log_233, content="User mentioned they like doing the budget review before kids wake up."
  Answer:
  {
    "label": "Detail_Miss",
    "step_failed": 3,
    "justification": "Comparing the evidence against the gold: the identity (recurring Sunday budget review) is establishable from log_211. Fields present: schedule (recurring weekend / Sunday), partial timing context (before kids wake up). Fields absent: the specific 09:00-10:30 timing window, the desk location, and the Mint/Chase/savings-goal action details required by the gold."
  }
\end{lstlisting}
\end{tcolorbox}

%% file: NeurIPS_2026/tables/life_domain.tex
\begin{table*}[t]
\centering
\caption{Mapping from prior life-domain frameworks to the taxonomy adopted in \muse.}
\label{tab:domain_mapping}
\resizebox{\textwidth}{!}{
\begin{tabular}{p{4cm} p{5cm} p{5cm}}
\toprule
\textbf{Our Life Domain} & \textbf{Diener et al. (SWB literature)} & \textbf{OECD Better Life Index} \\
\midrule

Work \& Education &
Work, Career &
Jobs, Education, Work-life balance \\
\addlinespace[0.9em]

Family \& Close Relationships &
Family, Marriage / Partner relationship &
Social connections (family-related aspects) \\
\addlinespace[0.9em]

Social \& Community &
Social relationships, Community involvement &
Social connections, Civic engagement, Safety \\
\addlinespace[0.9em]

Health \& Self-care &
Health &
Health, Environment, Work-life balance (health-related aspects) \\
\addlinespace[0.9em]

Finances \& Material Living &
Financial situation, Income &
Income, Housing \\
\addlinespace[0.9em]

Leisure \& Media Consumption &
Leisure, Free time &
-- (no direct counterpart; extended to reflect modern digital behaviors) \\

\bottomrule
\end{tabular}
}
\end{table*}

%% file: NeurIPS_2026/tables/life_domain_definition.tex
\begin{table*}[t]
\centering
\caption{Life domain taxonomy used in \muse. Each life domain corresponds to a major functional area of everyday life in which user attributes, behaviors, and preferences are organized.}
\label{tab:life_domain_taxonomy}
\small
\setlength{\tabcolsep}{6pt}
\renewcommand{\arraystretch}{1.08}

\begin{tabularx}{\textwidth}{
  @{}
  >{\raggedright\arraybackslash}p{3.8cm}
  >{\raggedright\arraybackslash}X
  @{}
}
\toprule
\textbf{Life Domain} & \textbf{Definition} \\
\midrule

Work \& Education &
This domain covers users' professional roles, career development, and learning activities, including employment status, occupational goals, skill acquisition, and educational pursuits. It reflects how users engage in productive activities and long-term capability building through work- and study-related behaviors. \\
\addlinespace[0.9em]

Family \& Close Relationships &
This domain describes users' family structure and intimate relationships, such as partnerships, parenting roles, and household responsibilities. It captures close interpersonal bonds that shape daily routines, obligations, and life decisions. \\
\addlinespace[0.9em]

Social \& Community &
This domain represents users' social networks and community involvement beyond the core family, including friendships, group participation, and civic activities. It reflects how users interact with broader social contexts and maintain social connectedness. \\
\addlinespace[0.9em]

Health \& Self-care &
This domain encompasses users' physical and mental well-being, including health conditions, lifestyle habits, and self-care practices such as exercise, diet, sleep, and healthcare-seeking behavior. It describes how users manage and optimize their health over time. \\
\addlinespace[0.9em]

Finances \& Material Living &
This domain covers users' financial situations and material conditions, including income level, spending patterns, housing, and access to material resources. It reflects how users allocate economic resources and maintain living standards. \\
\addlinespace[0.9em]

Leisure \& Media Consumption &
This domain captures users' recreational activities and content preferences, including entertainment, hobbies, travel, and consumption of digital media such as videos, music, games, and books. It reflects how users spend discretionary time and pursue enjoyment. \\

\bottomrule
\end{tabularx}
\end{table*}

%% file: NeurIPS_2026/tables/state_filtering_summary.tex

%% file: NeurIPS_2026/tables/state_validation_audit.tex



%% file: NeurIPS_2026/checklist.tex
\section*{NeurIPS Paper Checklist}

\begin{enumerate}

\item {\bf Claims}
    \item[] Question: Do the main claims made in the abstract and introduction accurately reflect the paper's contributions and scope?
    \item[] Answer: \answerYes{}
    \item[] Justification: The claims are supported by Section 5.
    \item[] Guidelines:
    \begin{itemize}
        \item The answer \answerNA{} means that the abstract and introduction do not include the claims made in the paper.
        \item The abstract and/or introduction should clearly state the claims made, including the contributions made in the paper and important assumptions and limitations. A \answerNo{} or \answerNA{} answer to this question will not be perceived well by the reviewers. 
        \item The claims made should match theoretical and experimental results, and reflect how much the results can be expected to generalize to other settings. 
        \item It is fine to include aspirational goals as motivation as long as it is clear that these goals are not attained by the paper. 
    \end{itemize}

\item {\bf Limitations}
    \item[] Question: Does the paper discuss the limitations of the work performed by the authors?
    \item[] Answer: \answerYes{} 
    \item[] Justification: Right before the appendix.
    \item[] Guidelines:
    \begin{itemize}
        \item The answer \answerNA{} means that the paper has no limitation while the answer \answerNo{} means that the paper has limitations, but those are not discussed in the paper. 
        \item The authors are encouraged to create a separate ``Limitations'' section in their paper.
        \item The paper should point out any strong assumptions and how robust the results are to violations of these assumptions (e.g., independence assumptions, noiseless settings, model well-specification, asymptotic approximations only holding locally). The authors should reflect on how these assumptions might be violated in practice and what the implications would be.
        \item The authors should reflect on the scope of the claims made, e.g., if the approach was only tested on a few datasets or with a few runs. In general, empirical results often depend on implicit assumptions, which should be articulated.
        \item The authors should reflect on the factors that influence the performance of the approach. For example, a facial recognition algorithm may perform poorly when image resolution is low or images are taken in low lighting. Or a speech-to-text system might not be used reliably to provide closed captions for online lectures because it fails to handle technical jargon.
        \item The authors should discuss the computational efficiency of the proposed algorithms and how they scale with dataset size.
        \item If applicable, the authors should discuss possible limitations of their approach to address problems of privacy and fairness.
        \item While the authors might fear that complete honesty about limitations might be used by reviewers as grounds for rejection, a worse outcome might be that reviewers discover limitations that aren't acknowledged in the paper. The authors should use their best judgment and recognize that individual actions in favor of transparency play an important role in developing norms that preserve the integrity of the community. Reviewers will be specifically instructed to not penalize honesty concerning limitations.
    \end{itemize}

\item {\bf Theory assumptions and proofs}
    \item[] Question: For each theoretical result, does the paper provide the full set of assumptions and a complete (and correct) proof?
    \item[] Answer: \answerNA{} 
    \item[] Justification: Does not include theoretical results.
    \item[] Guidelines:
    \begin{itemize}
        \item The answer \answerNA{} means that the paper does not include theoretical results. 
        \item All the theorems, formulas, and proofs in the paper should be numbered and cross-referenced.
        \item All assumptions should be clearly stated or referenced in the statement of any theorems.
        \item The proofs can either appear in the main paper or the supplemental material, but if they appear in the supplemental material, the authors are encouraged to provide a short proof sketch to provide intuition. 
        \item Inversely, any informal proof provided in the core of the paper should be complemented by formal proofs provided in appendix or supplemental material.
        \item Theorems and Lemmas that the proof relies upon should be properly referenced. 
    \end{itemize}

    \item {\bf Experimental result reproducibility}
    \item[] Question: Does the paper fully disclose all the information needed to reproduce the main experimental results of the paper to the extent that it affects the main claims and/or conclusions of the paper (regardless of whether the code and data are provided or not)?
    \item[] Answer: \answerYes{} 
    \item[] Justification: We fully disclose the experimental setting in both section 5 and appendix.
    \item[] Guidelines:
    \begin{itemize}
        \item The answer \answerNA{} means that the paper does not include experiments.
        \item If the paper includes experiments, a \answerNo{} answer to this question will not be perceived well by the reviewers: Making the paper reproducible is important, regardless of whether the code and data are provided or not.
        \item If the contribution is a dataset and\slash or model, the authors should describe the steps taken to make their results reproducible or verifiable. 
        \item Depending on the contribution, reproducibility can be accomplished in various ways. For example, if the contribution is a novel architecture, describing the architecture fully might suffice, or if the contribution is a specific model and empirical evaluation, it may be necessary to either make it possible for others to replicate the model with the same dataset, or provide access to the model. In general. releasing code and data is often one good way to accomplish this, but reproducibility can also be provided via detailed instructions for how to replicate the results, access to a hosted model (e.g., in the case of a large language model), releasing of a model checkpoint, or other means that are appropriate to the research performed.
        \item While NeurIPS does not require releasing code, the conference does require all submissions to provide some reasonable avenue for reproducibility, which may depend on the nature of the contribution. For example
        \begin{enumerate}
            \item If the contribution is primarily a new algorithm, the paper should make it clear how to reproduce that algorithm.
            \item If the contribution is primarily a new model architecture, the paper should describe the architecture clearly and fully.
            \item If the contribution is a new model (e.g., a large language model), then there should either be a way to access this model for reproducing the results or a way to reproduce the model (e.g., with an open-source dataset or instructions for how to construct the dataset).
            \item We recognize that reproducibility may be tricky in some cases, in which case authors are welcome to describe the particular way they provide for reproducibility. In the case of closed-source models, it may be that access to the model is limited in some way (e.g., to registered users), but it should be possible for other researchers to have some path to reproducing or verifying the results.
        \end{enumerate}
    \end{itemize}

\item {\bf Open access to data and code}
    \item[] Question: Does the paper provide open access to the data and code, with sufficient instructions to faithfully reproduce the main experimental results, as described in supplemental material?
    \item[] Answer: \answerYes{} 
    \item[] Justification: We open-sourced both the data and code with detailed instruction.
    \item[] Guidelines:
    \begin{itemize}
        \item The answer \answerNA{} means that paper does not include experiments requiring code.
        \item Please see the NeurIPS code and data submission guidelines (\url{https://neurips.cc/public/guides/CodeSubmissionPolicy}) for more details.
        \item While we encourage the release of code and data, we understand that this might not be possible, so \answerNo{} is an acceptable answer. Papers cannot be rejected simply for not including code, unless this is central to the contribution (e.g., for a new open-source benchmark).
        \item The instructions should contain the exact command and environment needed to run to reproduce the results. See the NeurIPS code and data submission guidelines (\url{https://neurips.cc/public/guides/CodeSubmissionPolicy}) for more details.
        \item The authors should provide instructions on data access and preparation, including how to access the raw data, preprocessed data, intermediate data, and generated data, etc.
        \item The authors should provide scripts to reproduce all experimental results for the new proposed method and baselines. If only a subset of experiments are reproducible, they should state which ones are omitted from the script and why.
        \item At submission time, to preserve anonymity, the authors should release anonymized versions (if applicable).
        \item Providing as much information as possible in supplemental material (appended to the paper) is recommended, but including URLs to data and code is permitted.
    \end{itemize}

\item {\bf Experimental setting/details}
    \item[] Question: Does the paper specify all the training and test details (e.g., data splits, hyperparameters, how they were chosen, type of optimizer) necessary to understand the results?
    \item[] Answer: \answerYes{} 
    \item[] Justification: In section 5 and appendix.
    \item[] Guidelines:
    \begin{itemize}
        \item The answer \answerNA{} means that the paper does not include experiments.
        \item The experimental setting should be presented in the core of the paper to a level of detail that is necessary to appreciate the results and make sense of them.
        \item The full details can be provided either with the code, in appendix, or as supplemental material.
    \end{itemize}

\item {\bf Experiment statistical significance}
    \item[] Question: Does the paper report error bars suitably and correctly defined or other appropriate information about the statistical significance of the experiments?
    \item[] Answer: \answerNo{} 
    \item[] Justification: Due to limitation of computation cost.
    \item[] Guidelines:
    \begin{itemize}
        \item The answer \answerNA{} means that the paper does not include experiments.
        \item The authors should answer \answerYes{} if the results are accompanied by error bars, confidence intervals, or statistical significance tests, at least for the experiments that support the main claims of the paper.
        \item The factors of variability that the error bars are capturing should be clearly stated (for example, train/test split, initialization, random drawing of some parameter, or overall run with given experimental conditions).
        \item The method for calculating the error bars should be explained (closed form formula, call to a library function, bootstrap, etc.)
        \item The assumptions made should be given (e.g., Normally distributed errors).
        \item It should be clear whether the error bar is the standard deviation or the standard error of the mean.
        \item It is OK to report 1-sigma error bars, but one should state it. The authors should preferably report a 2-sigma error bar than state that they have a 96\% CI, if the hypothesis of Normality of errors is not verified.
        \item For asymmetric distributions, the authors should be careful not to show in tables or figures symmetric error bars that would yield results that are out of range (e.g., negative error rates).
        \item If error bars are reported in tables or plots, the authors should explain in the text how they were calculated and reference the corresponding figures or tables in the text.
    \end{itemize}

\item {\bf Experiments compute resources}
    \item[] Question: For each experiment, does the paper provide sufficient information on the computer resources (type of compute workers, memory, time of execution) needed to reproduce the experiments?
    \item[] Answer: \answerYes{} 
    \item[] Justification: In the appendix.
    \item[] Guidelines:
    \begin{itemize}
        \item The answer \answerNA{} means that the paper does not include experiments.
        \item The paper should indicate the type of compute workers CPU or GPU, internal cluster, or cloud provider, including relevant memory and storage.
        \item The paper should provide the amount of compute required for each of the individual experimental runs as well as estimate the total compute. 
        \item The paper should disclose whether the full research project required more compute than the experiments reported in the paper (e.g., preliminary or failed experiments that didn't make it into the paper). 
    \end{itemize}
    
\item {\bf Code of ethics}
    \item[] Question: Does the research conducted in the paper conform, in every respect, with the NeurIPS Code of Ethics \url{https://neurips.cc/public/EthicsGuidelines}?
    \item[] Answer: \answerYes{} 
    \item[] Justification: The research conform with NeurIPS Code of Ethics
    \item[] Guidelines:
    \begin{itemize}
        \item The answer \answerNA{} means that the authors have not reviewed the NeurIPS Code of Ethics.
        \item If the authors answer \answerNo, they should explain the special circumstances that require a deviation from the Code of Ethics.
        \item The authors should make sure to preserve anonymity (e.g., if there is a special consideration due to laws or regulations in their jurisdiction).
    \end{itemize}

\item {\bf Broader impacts}
    \item[] Question: Does the paper discuss both potential positive societal impacts and negative societal impacts of the work performed?
    \item[] Answer: \answerYes{} 
    \item[] Justification: Before the limitation section.
    \item[] Guidelines:
    \begin{itemize}
        \item The answer \answerNA{} means that there is no societal impact of the work performed.
        \item If the authors answer \answerNA{} or \answerNo, they should explain why their work has no societal impact or why the paper does not address societal impact.
        \item Examples of negative societal impacts include potential malicious or unintended uses (e.g., disinformation, generating fake profiles, surveillance), fairness considerations (e.g., deployment of technologies that could make decisions that unfairly impact specific groups), privacy considerations, and security considerations.
        \item The conference expects that many papers will be foundational research and not tied to particular applications, let alone deployments. However, if there is a direct path to any negative applications, the authors should point it out. For example, it is legitimate to point out that an improvement in the quality of generative models could be used to generate Deepfakes for disinformation. On the other hand, it is not needed to point out that a generic algorithm for optimizing neural networks could enable people to train models that generate Deepfakes faster.
        \item The authors should consider possible harms that could arise when the technology is being used as intended and functioning correctly, harms that could arise when the technology is being used as intended but gives incorrect results, and harms following from (intentional or unintentional) misuse of the technology.
        \item If there are negative societal impacts, the authors could also discuss possible mitigation strategies (e.g., gated release of models, providing defenses in addition to attacks, mechanisms for monitoring misuse, mechanisms to monitor how a system learns from feedback over time, improving the efficiency and accessibility of ML).
    \end{itemize}
    
\item {\bf Safeguards}
    \item[] Question: Does the paper describe safeguards that have been put in place for responsible release of data or models that have a high risk for misuse (e.g., pre-trained language models, image generators, or scraped datasets)?
    \item[] Answer: \answerNA{} 
    \item[] Justification: Fully synthetic data, no risk for misuse.
    \item[] Guidelines:
    \begin{itemize}
        \item The answer \answerNA{} means that the paper poses no such risks.
        \item Released models that have a high risk for misuse or dual-use should be released with necessary safeguards to allow for controlled use of the model, for example by requiring that users adhere to usage guidelines or restrictions to access the model or implementing safety filters. 
        \item Datasets that have been scraped from the Internet could pose safety risks. The authors should describe how they avoided releasing unsafe images.
        \item We recognize that providing effective safeguards is challenging, and many papers do not require this, but we encourage authors to take this into account and make a best faith effort.
    \end{itemize}

\item {\bf Licenses for existing assets}
    \item[] Question: Are the creators or original owners of assets (e.g., code, data, models), used in the paper, properly credited and are the license and terms of use explicitly mentioned and properly respected?
    \item[] Answer: \answerYes{} 
    \item[] Justification: Properly cited and stated.
    \item[] Guidelines:
    \begin{itemize}
        \item The answer \answerNA{} means that the paper does not use existing assets.
        \item The authors should cite the original paper that produced the code package or dataset.
        \item The authors should state which version of the asset is used and, if possible, include a URL.
        \item The name of the license (e.g., CC-BY 4.0) should be included for each asset.
        \item For scraped data from a particular source (e.g., website), the copyright and terms of service of that source should be provided.
        \item If assets are released, the license, copyright information, and terms of use in the package should be provided. For popular datasets, \url{paperswithcode.com/datasets} has curated licenses for some datasets. Their licensing guide can help determine the license of a dataset.
        \item For existing datasets that are re-packaged, both the original license and the license of the derived asset (if it has changed) should be provided.
        \item If this information is not available online, the authors are encouraged to reach out to the asset's creators.
    \end{itemize}

\item {\bf New assets}
    \item[] Question: Are new assets introduced in the paper well documented and is the documentation provided alongside the assets?
    \item[] Answer: \answerYes{} 
    \item[] Justification: Proposed dataset is well documented.
    \item[] Guidelines:
    \begin{itemize}
        \item The answer \answerNA{} means that the paper does not release new assets.
        \item Researchers should communicate the details of the dataset\slash code\slash model as part of their submissions via structured templates. This includes details about training, license, limitations, etc. 
        \item The paper should discuss whether and how consent was obtained from people whose asset is used.
        \item At submission time, remember to anonymize your assets (if applicable). You can either create an anonymized URL or include an anonymized zip file.
    \end{itemize}

\item {\bf Crowdsourcing and research with human subjects}
    \item[] Question: For crowdsourcing experiments and research with human subjects, does the paper include the full text of instructions given to participants and screenshots, if applicable, as well as details about compensation (if any)? 
    \item[] Answer: \answerNo{} 
    \item[] Justification: Not related.
    \item[] Guidelines:
    \begin{itemize}
        \item The answer \answerNA{} means that the paper does not involve crowdsourcing nor research with human subjects.
        \item Including this information in the supplemental material is fine, but if the main contribution of the paper involves human subjects, then as much detail as possible should be included in the main paper. 
        \item According to the NeurIPS Code of Ethics, workers involved in data collection, curation, or other labor should be paid at least the minimum wage in the country of the data collector. 
    \end{itemize}

\item {\bf Institutional review board (IRB) approvals or equivalent for research with human subjects}
    \item[] Question: Does the paper describe potential risks incurred by study participants, whether such risks were disclosed to the subjects, and whether Institutional Review Board (IRB) approvals (or an equivalent approval/review based on the requirements of your country or institution) were obtained?
    \item[] Answer: \answerNA{} 
    \item[] Justification: Not related.
    \item[] Guidelines:
    \begin{itemize}
        \item The answer \answerNA{} means that the paper does not involve crowdsourcing nor research with human subjects.
        \item Depending on the country in which research is conducted, IRB approval (or equivalent) may be required for any human subjects research. If you obtained IRB approval, you should clearly state this in the paper. 
        \item We recognize that the procedures for this may vary significantly between institutions and locations, and we expect authors to adhere to the NeurIPS Code of Ethics and the guidelines for their institution. 
        \item For initial submissions, do not include any information that would break anonymity (if applicable), such as the institution conducting the review.
    \end{itemize}

\item {\bf Declaration of LLM usage}
    \item[] Question: Does the paper describe the usage of LLMs if it is an important, original, or non-standard component of the core methods in this research? Note that if the LLM is used only for writing, editing, or formatting purposes and does \emph{not} impact the core methodology, scientific rigor, or originality of the research, declaration is not required.
    \item[] Answer: \answerYes{} 
    \item[] Justification: Clearly explained in section 3 and 4.
    \item[] Guidelines:
    \begin{itemize}
        \item The answer \answerNA{} means that the core method development in this research does not involve LLMs as any important, original, or non-standard components.
        \item Please refer to our LLM policy in the NeurIPS handbook for what should or should not be described.
    \end{itemize}

\end{enumerate}